\documentclass[10pt,twocolumn,letterpaper]{article}

\usepackage{iccv}
\usepackage{times}
\usepackage{epsfig}
\usepackage{graphicx}
\usepackage{amsmath}
\usepackage{amssymb}
\usepackage{booktabs}
\usepackage[accsupp]{axessibility}  % Improves PDF readability for those with disabilities.

\usepackage[breaklinks=true,bookmarks=false]{hyperref}

\usepackage[capitalize]{cleveref}
\crefname{section}{Sec.}{Secs.}
\Crefname{section}{Section}{Sections}
\Crefname{table}{Table}{Tables}
\crefname{table}{Tab.}{Tabs.}
\usepackage{siunitx}
\usepackage[font=scriptsize]{caption}

\iccvfinalcopy % *** Uncomment this line for the final submission

\def\iccvPaperID{DFAD-9} % *** Enter the ICCV Paper ID here

\ificcvfinal\fi

\begin{document}

%%%%%%%%% TITLE
\title{TrainFors: A Large Benchmark Training Dataset for Image Manipulation Detection and Localization}

\author{Soumyaroop Nandi, Prem Natarajan, Wael Abd-Almageed\\
USC Information Sciences Institute, Marina del Rey, CA, USA\\
% Visual Intelligence and Multimedia Analytics Laboratory\\
{\tt\small \{soumyarn,pnataraj,wamageed\}@isi.edu}}

\maketitle
\ificcvfinal\thispagestyle{empty}\fi

%%%%%%%%% ABSTRACT
\begin{abstract}
\vspace{-5px}
   The evaluation datasets and metrics for image manipulation detection and localization (IMDL) research have been standardized. But the training dataset for such a task is still nonstandard. Previous researchers have used unconventional and deviating datasets to train neural networks for detecting image forgeries and localizing pixel maps of manipulated regions. For a fair comparison, the training set, test set, and evaluation metrics should be persistent. Hence, comparing the existing methods may not seem fair as the results depend heavily on the training datasets as well as the model architecture. Moreover, none of the previous works release the synthetic training dataset used for the IMDL task. We propose a standardized benchmark training dataset for image splicing, copy-move forgery, removal forgery, and image enhancement forgery. Furthermore, we identify the problems with the existing IMDL datasets and propose the required modifications. We also train the state-of-the-art IMDL methods on our proposed TrainFors\footnote{\href{https://github.com/vimal-isi-edu/TrainFors}{\textcolor{red}{https://github.com/vimal-isi-edu/TrainFors}}} dataset for a fair evaluation and report the actual performance of these methods under similar conditions.
\end{abstract}

%%%%%%%%% BODY TEXT
\vspace{-15px}
\section{Introduction}\label{intro}
% \label{sec:intro}

Image manipulation and the effects of such acts have become a challenging problem in today's society, owing to the low-cost, publicly accessible image editing tools (\cite{park2020swapping}, \cite{li2020manigan}, \cite{vinker2020deep}, \cite{dhamo2020semantic}) and photo-realistic generative models like GANs(\cite{goodfellow2014generative},  \cite{mirza2014conditional}, \cite{zhu2017unpaired}) and VAEs(\cite{kingma2013auto}, \cite{pu2016variational}) for creating  manipulated images. Deceitful attackers may use such tools to spread misinformation like deep fakes (\cite{sabir2019recurrent}), fake news (\cite{huh2018fighting}), plagiarised academic publications (\cite{sabir2021biofors}), internet rumors (\cite{wu2022robust}), forged satellite images (\cite{horvath2021manipulation}). A news article (\cite{bbcfakenews2022}) highlights the usage of manipulated images and videos in the Russia-Ukraine war to spread misinformation. Defending such manipulated misinformation spread is the need of the hour and image manipulation detection and localization is an effort to counter such societal problems. 

\newlength{\tempdima}
\newcommand{\rowname}[1]% #1 = text
{\rotatebox{90}{\makebox[\tempdima][c]{\textbf{#1}}}}

% \newcounter{subfigure}[figure]
% \renewcommand{\thesubfigure}{\alph{subfigure}}
% \newcommand{\mycaption}[1]% #1 = caption
% {\refstepcounter{subfigure}\textbf{(\thesubfigure) }{\ignorespaces #1}}

\begin{figure}
    \centering
    \settoheight{\tempdima}{\includegraphics[width=.27\linewidth]    {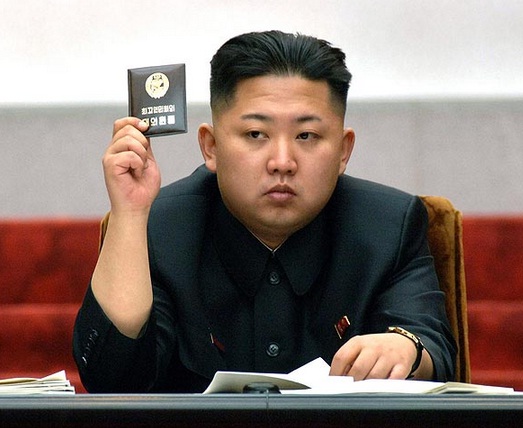}}
    \centering\begin{tabular}{@{}c@{ }c@{ }c@{ }c@{}}
    &\textbf{Pristine} & \textbf{Manipulated} & \textbf{Groundtruth} \\
    \rowname{Splicing}&
    \includegraphics[width=.3\linewidth]{Images/Fig1/splicing_2_orig.jpg}& 
    \includegraphics[width=.3\linewidth]{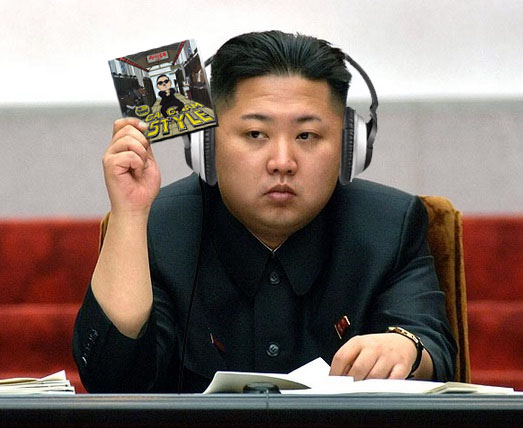}& 
    \includegraphics[width=.3\linewidth]{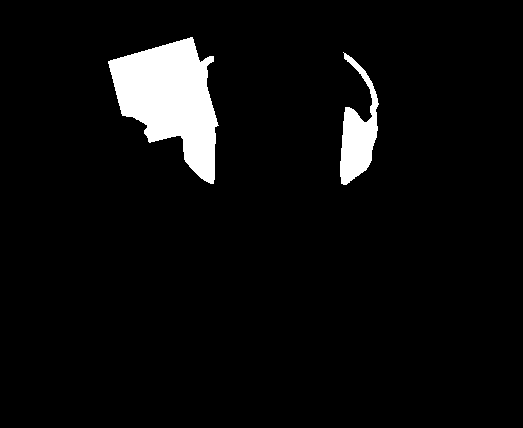}\\[-1.5ex] 
    % &\mycaption{} & \mycaption{} & \mycaption{}\\
    \rowname{Copymove}&
    \includegraphics[width=.3\linewidth]{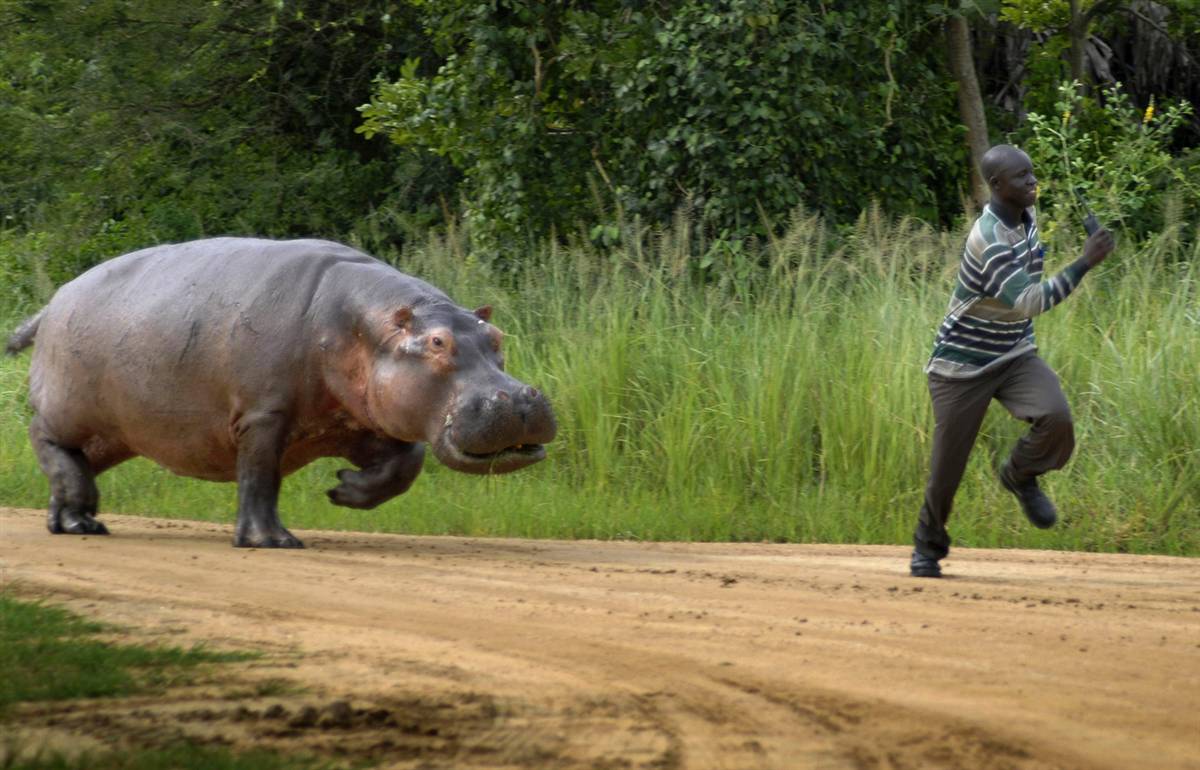}& 
    \includegraphics[width=.3\linewidth]{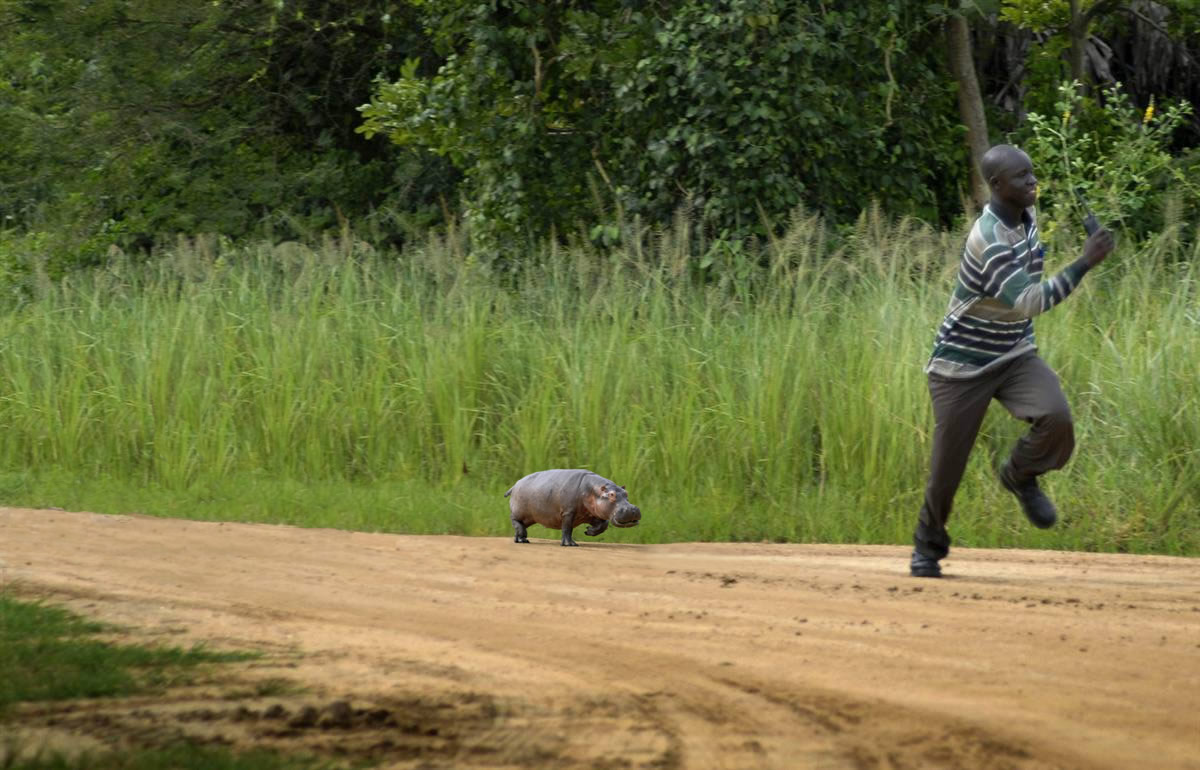}& 
    \includegraphics[width=.3\linewidth]{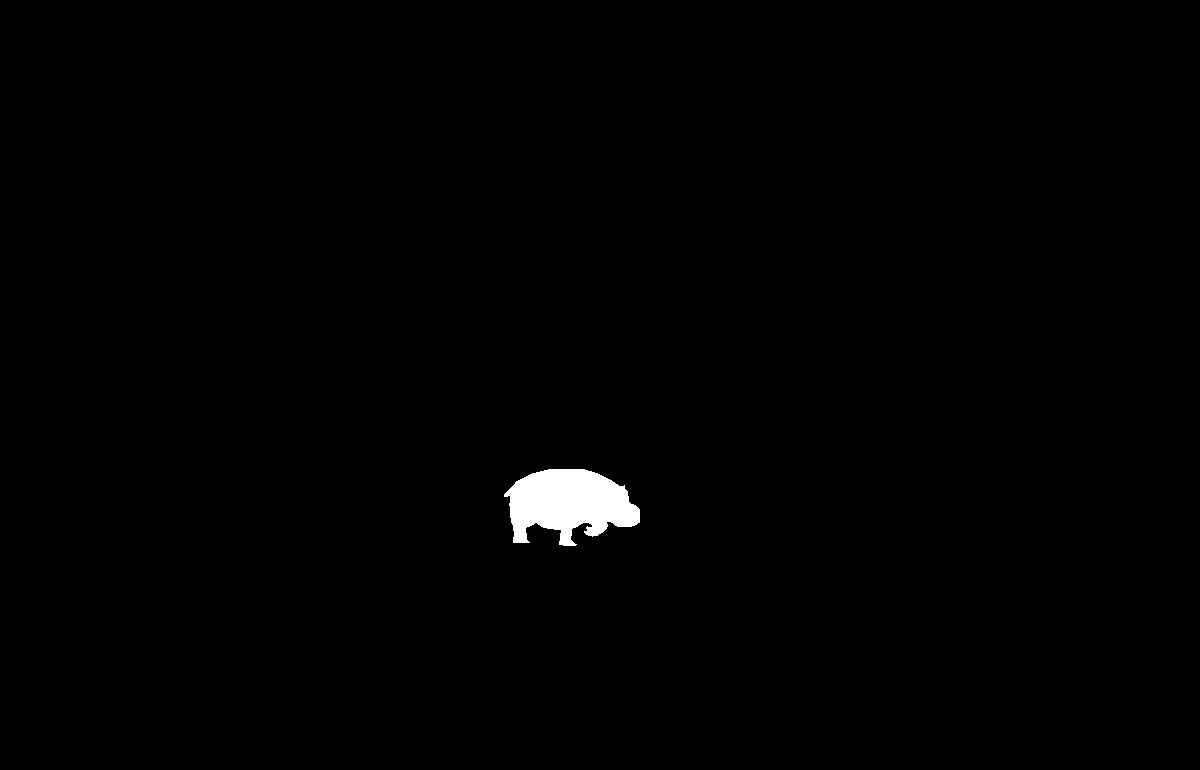}\\[-1ex] 
    % &\mycaption{} & \mycaption{} & \mycaption{}\\
    \rowname{Removal}&
    \includegraphics[width=.3\linewidth]{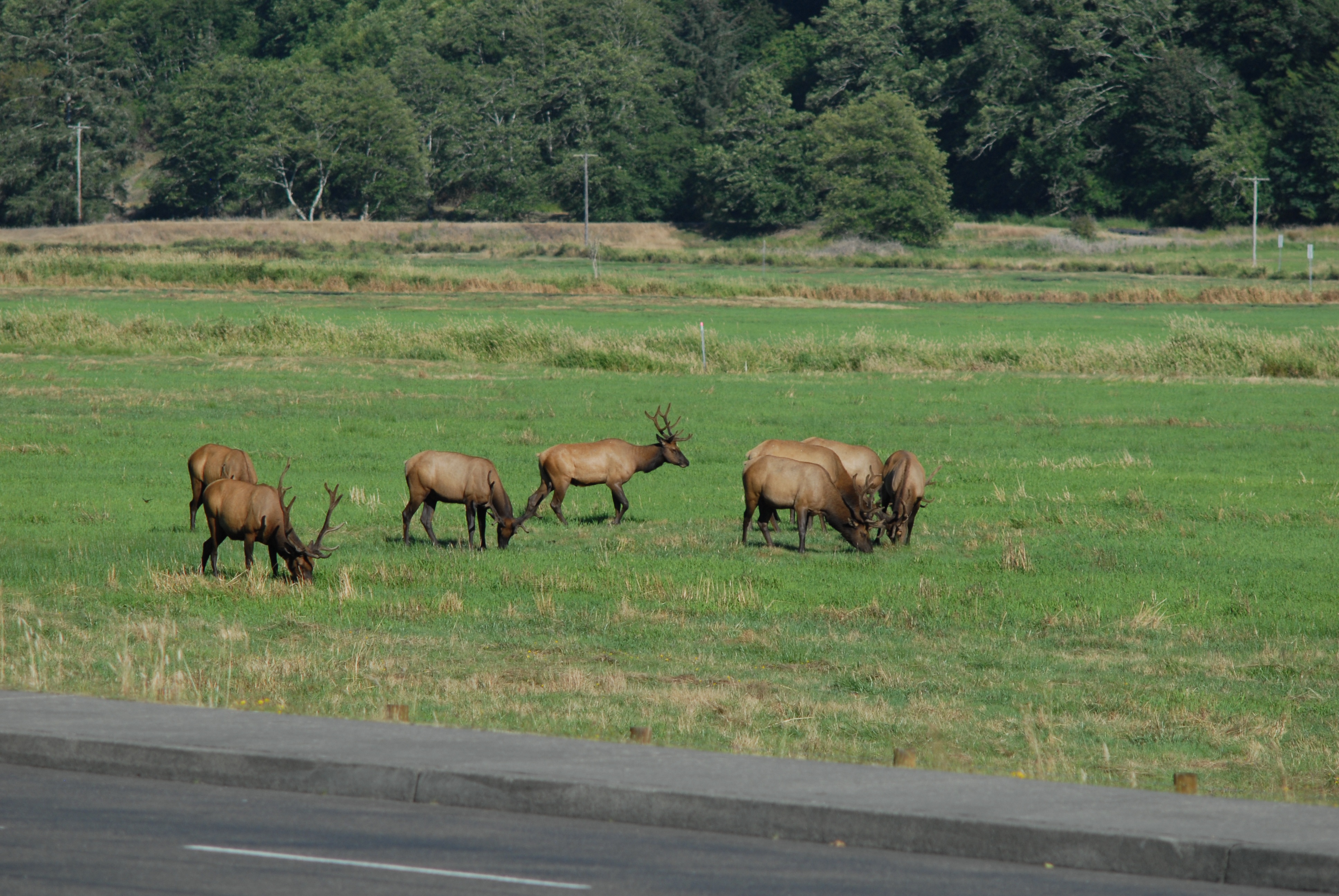}& 
    \includegraphics[width=.3\linewidth]{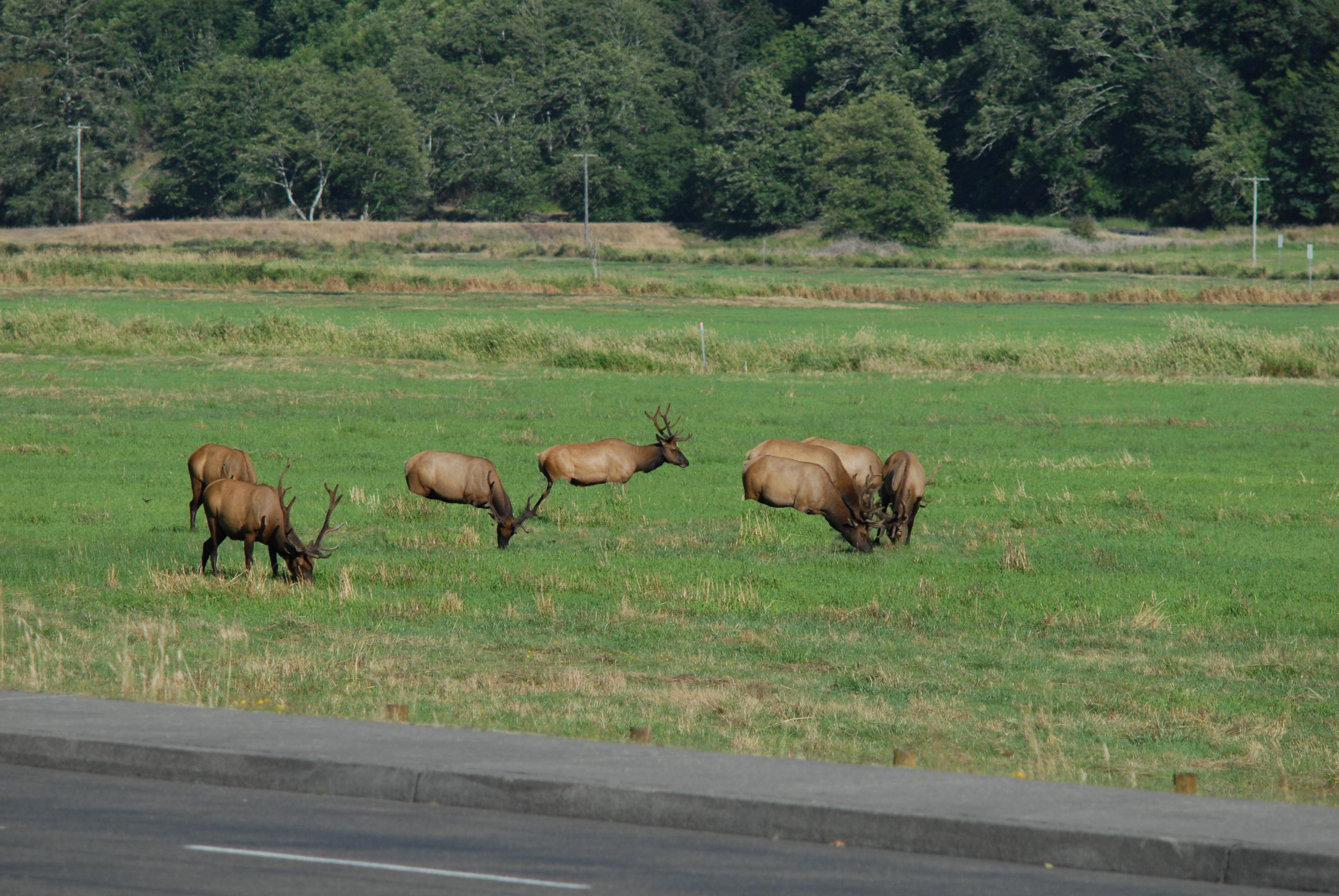}& 
    \includegraphics[width=.3\linewidth]{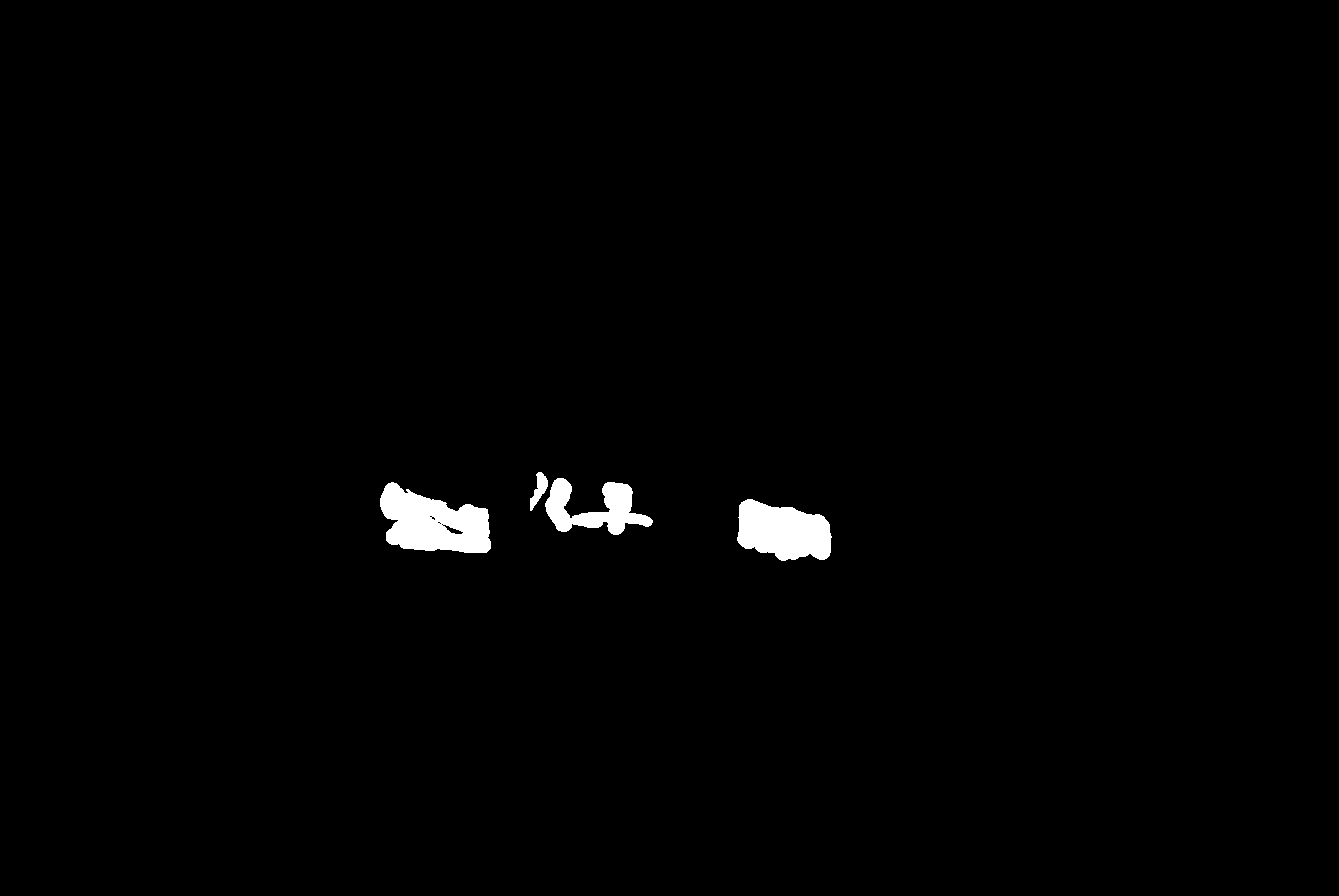}\\[-1ex] 
    \end{tabular}
    \caption{Image Manipulation Localization Examples - From top to bottom, each row shows Image Splicing, Copymove Manipulation, and Removal Manipulation. The binary mask on the third column represents the localized manipulated pixels.}
    \label{fig:IMDL_tasks}
\end{figure}

\begin{figure}
    \centering
    \settoheight{\tempdima}{\includegraphics[width=.05\linewidth]{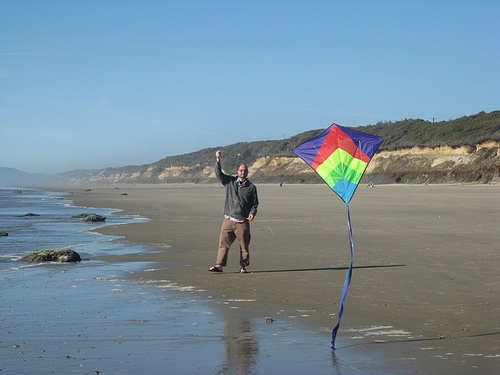}}
    \centering\begin{tabular}{@{}c@{}c@{}c@{}c@{}c@{}c@{}}
    % \textbf{Pristine} & \textbf{JPEG Compress} & \textbf{Resampled} & \textbf{Hist Mani} & \textbf{Quantization} \\
    % \rowname{Splicing}&
    \includegraphics[width=.2\linewidth]{Images/Fig2/orig_img.jpg}
    \includegraphics[width=.2\linewidth]{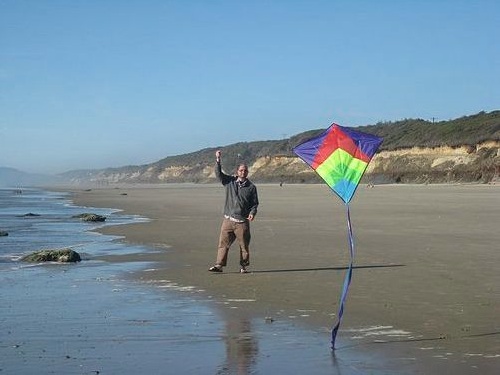}
    \includegraphics[width=.2\linewidth]{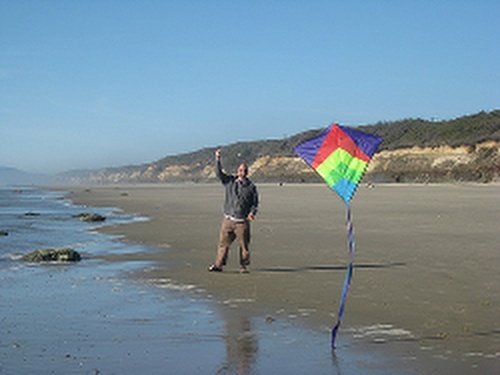}
    \includegraphics[width=.2\linewidth]{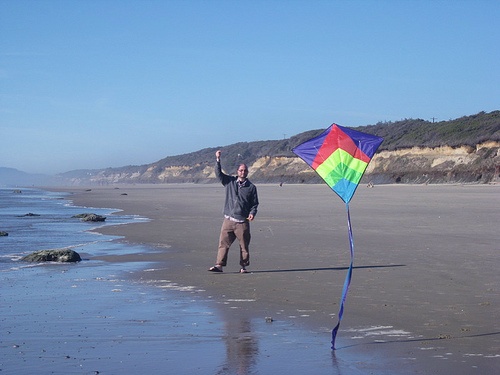}
    \includegraphics[width=.2\linewidth]{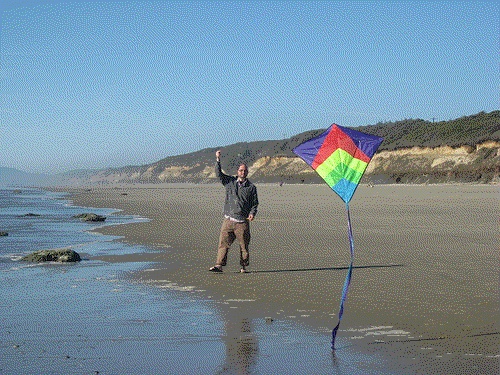}
    \end{tabular}
    \caption{Image Enhancement Manipulations, without modifying the image content, are difficult to detect - From left to right: Pristine, Double JPEG Compressed, Lancoz Resampled, Histogram Manipulation, and Dithering Quantization.}
    \label{fig:IMDL_enhance}
\end{figure}

The problem can be divided into two subtasks - \noindent\textit{{detecting}} manipulated images and \noindent\textit{{localizing}} a pixel map of the manipulated region in the forged images. Generally, an image can be tampered in two ways - using image content or without using any image content. Image content can either be moved from one part of the image to another internally (\textit{copy-move forgery}) or externally import certain parts from an alien image (\textit{splicing forgery}). Some forgers try to remove certain image patches or objects and replace them with surrounding inpainted pixels (\textit{removal forgery}). Besides this, image forgeries can also be executed without modifying the image content, but by \textit{image enhancement} (e.g., compression, resampling, blurring, noise, morphing, quantization, histogram manipulation). see \cref{fig:IMDL_tasks} and \cref{fig:IMDL_enhance}

The image manipulation process often leads to artifacts around the tampered patches. Researchers have used various clues like noise pattern (\cite{yang2016image}, \cite{lyu2014exposing}, \cite{cozzolino2019noiseprint}), camera model parameters (\cite{bondi2017tampering}, \cite{bondi2016first}), edge inconsistencies (\cite{salloum2018image}, \cite{zhang2018boundary}), color inconsistency (\cite{fan2015image}), EXIF inconsistencies (\cite{huh2018fighting}), visual similarities (\cite{wu2017deep}, \cite{wu2018busternet}, \cite{wu2018image}) and JPEG compression artifacts (\cite{han2016efficient}, \cite{uricchio2017localization}, \cite{li2017image}) to detect artifacts in forged images. Most of the previous works have tried to focus on one or two clues to detect and localize manipulations. Also, they localized the manipulated patches or pixels on the tampered images. But in the real-world scenario, we never know if the image is manipulated or pristine. Neither do we know all the types of manipulations applied on an image and post-processing performed to hide such malicious acts. 

Recent DNN-based object detection methods (\cite{zhu2020deformable}, \cite{dosovitskiy2020image}, \cite{he2016deep}, \cite{huang2017densely}) try to detect single or multiple objects in an image. But our task is to determine the tampered pixels in a given image. This task is somewhat more challenging because an attacker removes/modifies certain pixels and replaces those pixels with new objects/backgrounds. An object detector may perform semantic segmentation to distinguish all the objects in an image, but we can never know if any of those objects were tampered with in the first place. This necessitates the development of image manipulation detection and localization (IMDL) models, working in parallel with an object detector.

Previous research works do not use a standard training dataset for developing the IMDL models. They have claimed to improve the AUC and F1 scores with a new neural network architecture, trained on a curated synthetic training dataset that varies for each model. Moreover, neither do they release the synthetic training dataset and the training code to reproduce the claimed results nor do they elaborately describe the synthetic training dataset generation procedure. Only the inference code and pre-trained model weights are released for some of the existing research. This violates the repeatability of scientific experimentation in the IMDL tasks. To overcome this massive flaw in the IMDL experimentation, we propose a new benchmark dataset - TrainFors. We hope that TrainFors will standardize the IMDL algorithm designing and software development. TrainFors contain 1 million images (200K pristine, 800K manipulated), belonging to four standardized image forgery tasks - image splicing, copy-move forgery, image inpainting (removal), and image enhancement. 

The main contributions of our paper are as follows:
\vspace{-5px}
\begin{enumerate}
\item[$\bullet$] A large standardized benchmark training dataset with real-world forgeries for the image manipulation detection and localization (IMDL) task.
\vspace{-5px}
\item[$\bullet$] Extensive analysis of the baseline IMDL models, trained from scratch on our proposed TrainFors dataset. All the baseline models are trained under similar conditions and evaluated with the same evaluation metrics for a fair comparison.
\vspace{-5px}
\item[$\bullet$] The major challenges in curating a training dataset for IMDL tasks are elaborately described and intelligent solutions are proposed to minimize the difference between real-world manipulated images and the forged training images.

\end{enumerate}

\vspace{-5px}
\section{Related Work}
\label{sec:formatting}
\vspace{-5px}

Most of the previous works have focused on one or multiple forgeries like boundary artifacts, noise pattern-based, double JPEG compression-based, color filter array-based to detect and localize the tampered pixels. Some researchers used handcrafted features and others have trained a deep neural network (DNN) to find the image forgeries.

\subsection{\noindent\textbf{Handcrafted feature-based methods}} 
\vspace{-5px}
The most prominent handcrafted feature-based methods used in image forgery detection and localization are \noindent\textbf{ELA} \cite{krawetz2007picture} (finds the compression error difference between forged regions and pristine regions through different JPEG compression qualities), \noindent\textbf{NOI1} \cite{mahdian2009using} (models local noise by using high pass wavelet coefficients) and \noindent\textbf{CFA1} \cite{ferrara2012image} (approximates the camera filter array patterns by using nearby pixels and generates a tampering probability for each pixel). They studied only a single type of forgery (among splicing, copy-move, and inpainting), splicing being the most common.

\subsection{\noindent\textbf{Early DNN-based models and training datasets}} \label{early_DNN}
\vspace{-5px}
The earlier DNN-based studies predominantly detected a specific type of manipulation, e.g., splicing (\cite{cozzolino2015splicebuster}, \cite{wu2017deep}, \cite{kniaz2019point}), copy-move (\cite{cozzolino2015efficient}, \cite{rao2016deep}, \cite{wu2018busternet}, \cite{wu2018image}, \cite{islam2020doa}), removal (\cite{zhu2018deep}) and enhancement (\cite{bayar2016deep}, \cite{bayar2018constrained}, \cite{choi2017detecting}). But we do not have apriori information about the type of forgery implemented on real-world manipulated images. So general forgery detection and localization algorithms were proposed that could detect any type of unknown manipulation in an image.  \noindent\textbf{J-LSTM} \cite{bappy2017exploiting} and \noindent\textbf{H-LSTM} \cite{bappy2019hybrid} jointly trained an LSTM and CNN to capture the boundary-discriminative features by training a synthetic dataset (unavailable) sampled from MS-COCO \cite{lin2014microsoft}, NIST16 \cite{NimbleCh33:online}, and Dresden database \cite{gloe2010dresden}. Both methods detected predefined size regions (restricted by preset patch size) and are very time-consuming. \noindent\textbf{RGB-N} \cite{zhou2018learning} used a synthetic training dataset (unavailable) sampled from MS-COCO \cite{lin2014microsoft} and adopted a steganalysis-based SRM kernel model \cite{fridrich2012rich} and a two-stream Faster R-CNN \cite{ren2015faster}, but it cannot generate pixel-wise segmentation masks but only bounding boxes around tampered regions. \noindent\textbf{ManTraNet} \cite{wu2019mantra} used a synthetic training dataset (unavailable) sampled from KCMI \cite{kcmi:online} and Dresden database \cite{gloe2010dresden} and classified 385 unknown manipulation types and trained bipartite end-to-end network to detect image-level manipulations with one part using SRM kernel \cite{fridrich2012rich} in feature extraction and training, while the second subnet used Bayar convolution on synthetic forgery datasets followed by pixel-wise anomaly detection, but failed for double JPEG compression artifacts. \noindent\textbf{SPAN} (\cite{hu2020span}) used the synthetic dataset created by ManTraNet (unavailable) and used a pyramid structure of local self-attention blocks to model the relationships of pixels on varying scales. However, the correlation is only considered in the local regions and fails in capturing spatial correlation on global features. \noindent\textbf{GSR-Net} \cite{zhou2020generate} used a synthetic training dataset (unavailable) sampled from MS-COCO \cite{lin2014microsoft} and Casiav2 \cite{dong2013casia} and implemented an edge detection and refinement branch that accepts features from different levels. Region segmentation and edge detection are very different tasks and can affect the performance of each other if trained together. 

\subsection{\noindent\textbf{Recent DNN-based models and training datasets}} \label{recent_DNN}
\vspace{-5px}
\noindent\textbf{PSCCNet} (\cite{liu2021pscc}) used a synthetic training dataset (unable to download) sampled from MS-COCO \cite{lin2014microsoft}, KCMI \cite{kcmi:online}, Dresden \cite{gloe2010dresden}, and Busternet database \cite{wu2018busternet} and extracted hierarchical features along a top-down path and used a bottom-up path to detect manipulated regions in images. \noindent\textbf{MVSS-Net} \cite{chen2021image} used Defacto \cite{mahfoudi2019defacto} and Casiav2 \cite{dong2013casia} datasets for both training and evaluation and tried to address this problem by using an  edge-supervised branch. They used noise distribution to create generalizable features and boundary artifacts surrounding tampered regions. They tried to address an important challenge of the IMDL task of balance between sensitivity and specificity, but fails to reach an equilibrium.  \noindent\textbf{CAT-Net} \cite{kwon2021cat} used a synthetic training dataset (unable to download) sampled from MS-COCO \cite{lin2014microsoft}, Casiav2 \cite{dong2013casia}, IMD2020 \cite{novozamsky2020imd2020} and Fantastic Reality \cite{kniaz2019point} and learns forensic features of compression artifacts on RGB and DCT domains and concatenates the features at a middle stage. They employ DCT histograms to detect double JPEG compression for image splicing only. \noindent\textbf{Transforensics} \cite{hao2021transforensics} used Casiav2 \cite{dong2013casia}, Coverage \cite{wen2016coverage} and IMD2020 \cite{novozamsky2020imd2020} for both training and evaluation. They used a transformer-based model with dense self-attention encoders (FCN) and dense correction modules to capture global context and pairwise interaction between local patches through spatial self-attention. \noindent\textbf{RTAG} \cite{bi2021reality} used Fantastic Reality \cite{kniaz2019point} dataset for training and extended the \noindent\textbf{MAGritte} \cite{kniaz2019point} model to perform style transfer to create forged images, detect manipulated images, and localize manipulated pixels for splicing forgery using two GANs (faker for retouching and authenticator for localizing forgeries). They Adversarially train both generators to find whitening and coloring transforms. \noindent\textbf{Trace} \cite{bammey2022non} created a manipulated database with controlled image cues (noise models, image demosaicing, color correction, JPEG compression) and evaluated the image forensic tools. They generated user-defined endo-mask and exo-mask for each manipulation cue and the database has 5000 manipulated images, which is not enough to train a large DNN. This database is not very impactful when trying to detect real-life manipulated images.  \noindent\textbf{Objectformer} (\cite{wang2022objectformer}) used a synthetic training dataset sampled from MS-COCO \cite{lin2014microsoft} and Paris-Street-View \cite{pathak2016context} and detected the manipulation artifacts by extracting and training multimodal patch embeddings with high-frequency features and RGB features combined and then used object prototypes to model object-level consistencies and find patch-level inconsistencies. They did not release their code or training dataset and hence the claimed results cannot be verified.

We found three major problems with the prior research work done in the IMDL task:
\vspace{-5px}
\begin{enumerate}
\item[$\bullet$] The synthetic training datasets for most of the methods are unavailable. Only the pre-trained weights and inference code is released for a few of them. No detail is provided for creating a standardized training dataset. 
\vspace{-5px}
\item[$\bullet$] All the previous methods used varying datasets of different sizes and tasks to create a synthetic training dataset. Some methods used the same dataset for training and evaluation. This may induce bias in evaluation.
\vspace{-5px}
\item[$\bullet$] All the previous methods used varying backbone deep neural networks for pretraining and finetuning. We propose to verify the performance of all the models, when pretrained on a single backbone network.

\end{enumerate}

% \noindent\textbf{CNN/Transformer based model:}

\vspace{-5px}
\section{TrainFors: Benchmark IMDL training set}
\vspace{-5px}

The main contribution of this paper is the development of a large-scale image manipulation dataset - \noindent\textit{TrainFors}. The IMDL community uses benchmark evaluation datasets for manipulation detection and localization. But large-scale training dataset with real-world manipulations for the IMDL task is non-existent. We used open-source images to curate TrainFors. We need both manipulated and pristine images for the image manipulation detection task. The manipulated images can have four types of manipulations as discussed in section \ref{intro}. Hence, we created five sets of images - pristine and the four manipulated types in the TrainFors dataset, see \cref{tab.trainfor_desc}. Most of the previous works used one or more of these five image sets for training. TrainFors is an effort to accumulate all the manipulation types and pristine images under the same umbrella with the goal to diminish the difference between the curated training set tampered images and real-world manipulated images.

\begin{table}
    \vspace{-5px}
    \centering
    \small
    \setlength{\tabcolsep}{1pt}

    \scalebox{0.65}{
    \begin{tabular}{l cccccc}
    \toprule
   
    \bf{Source Dataset} & \bf{\#Pristine} & \bf{\#Manipulated} & \bf{\#Splicing} & \bf{\#Copy-Move} & \bf{\#Removal} & \bf{\#Enhancement} \\
    
    \cmidrule(r){1-1} \cmidrule(lr){2-2} \cmidrule(lr){3-3} \cmidrule(l){4-4} \cmidrule(l){5-5} \cmidrule(l){6-6} \cmidrule(l){7-7}

    \textit{Existing Training} &  &  &  &  &  &   \\
    \cmidrule(r){1-1}
    Trace \cite{bammey2022non} & 1K & 10K & 5K & 5K & 0 & 0 \\
    Defacto\cite{mahfoudi2019defacto} & 0 & 149K & 105K & 19K & 25K & 0 \\
    Casiav2\cite{dong2013casia} & 0 & 5123 & 1828 & 3295 & 0 & 0 \\
    Dresden\cite{gloe2010dresden} & 0 &  35K & 35K & 0 & 0 & 0 \\
    \midrule    

    \textit{Newly Generated Training} &  &  &  &  &  &   \\
    \cmidrule(r){1-1}
    MS-COCO \cite{lin2014microsoft} & 172K & 541K & 25.2K & 172.8K & 171K & 172K \\
    Socrates \cite{galdi2019socrates} & 3.2K & 6.4K & 3.2K & 0 & 0 & 3.2K \\
    Vision \cite{shullani2017vision} & 12.3K & 24.6K & 12.3K & 0 & 0 & 12.3K \\
    FODB \cite{hadwiger2021forchheim} & 7.6K & 15.2K & 7.6 & 0 & 0 & 7.6K \\
    KCMI \cite{kcmi:online} & 916 & 1832 & 916 & 0 & 0 & 916 \\
    Paris-Street-View \cite{pathak2016context} & 3K & 12K & 4K & 0 & 4K & 4K \\

    \midrule
    \bf{Total: TrainFors} & 200K & 800K & 200K & 200K & 200K & 200K \\
    \midrule
    
    \textit{Evaluation} &  &  &  &  &  &  \\
    \cmidrule(r){1-1}
    Columbia \cite{ng2009columbia} & 183 & 180 & 180 & 0 & 0 & 0 \\
    Coverage \cite{wen2016coverage} & 100 & 100 & 0 & 100 & 0 & 0 \\
    CASIAv1 \cite{dong2013casia} & 800 & 920 & 461 & 459 & 0 & 0 \\
    NIST16 \cite{NimbleCh33:online} & 0 & 611 & 225 & 282 & 104 & 0 \\
    IMD2020 \cite{novozamsky2020imd2020} & 414 & 2010 & 1810 & 100 & 100 & 0 \\
    \bottomrule
    
    \end{tabular}}
    \vspace{-5px}
    \caption{Training and Evaluation Image Distribution - Number of images for each type of manipulation}
    \label{tab.trainfor_desc}
\end{table}

% \vspace{-5px}
\subsection{Image Collection}
\vspace{-5px}

There exist a few image manipulation datasets that are released. But they are not sufficient and most of the models create their own synthetic training datasets to train the large DNNs. TrainFors is a large database that contains images from existing manipulated datasets and also some new manipulated images were generated to create an exclusive superset of IMDL training database. We used manipulated images from Trace \cite{bammey2022non}, Defacto \cite{mahfoudi2019defacto}, Casiav2 \cite{dong2013casia} and Dresden \cite{gloe2010dresden} datasets and created another set of pristine and manipulated images using MS-COCO \cite{lin2014microsoft}, Socrates \cite{galdi2019socrates}, Vision \cite{shullani2017vision}, FODB \cite{hadwiger2021forchheim}, KCMI \cite{kcmi:online}, and Paris-Street-View \cite{pathak2016context} datasets.  A detailed description of the source images used to generate the TrainFors database is presented in \cref{tab.trainfor_desc}.

% \vspace{-5px}
\subsection{Dataset Description}
\vspace{-5px}

Socrates \cite{galdi2019socrates}, Vision \cite{shullani2017vision}, FODB \cite{hadwiger2021forchheim}, and KCMI \cite{kcmi:online} are Camera Identifying datasets and are not manipulated directly, neither do they possess ground-truth binary masks of manipulated pixels. To create meaningful tampered images from these images, we have externally added manipulated pixels and generated the ground-truth binary masks. They were used to generate splicing and image enhancement manipulations. The external manipulated pixels were added from MS-COCO \cite{lin2014microsoft} objects because MS-COCO provides a ground-truth mask of two million objects. Semantically meaningful objects were created from MS-COCO annotations and were used to generate splicing dataset. Paris-Street-View \cite{pathak2016context} dataset was directly used for creating removal manipulated images following the inpainting protocol mentioned in \cite{pathak2016context}. Copy-move images can only be generated if the ground-truth masks of objects are available and hence only MS-COCO was used to generate them.

% \vspace{-5px}
\subsection{Robust Forgery Pipelines with TrainFors}
\vspace{-5px}

MS-COCO \cite{lin2014microsoft} has 12 supercategories and 80 categories of object-type images. We have used different combinations of these categories to generate manipulated pixels for splicing, copy-move, and removal images. We tried to make sure that the manipulated images look like real-world images and should be difficult to detect such manipulations by the naked eye. For each of the manipulation types, different pipelines are used to create the manipulated images.

% \subsubsection{Splicing}
\noindent\textbf{Splicing:} We generated spliced images using four types of combinations of MS-COCO categories. In the first combination, two images from the same supercategory were chosen. For example bird and cat categories under the same supercategory animal. In the second combination, objects from two different supercategories were chosen, for example, bird and stop sign from animal and outdoor supercategories respectively. In the third combination, we created spliced images from the same supercategory and category. For example, two different images of the  person category were spliced. We generated 100 instances of each combination to generate the spliced images. 

In image splicing, we have a pair of images - a donor image and a target image. To make the spliced images more realistic, we refined the segmented objects from the donor images using \textit{MGMatting} \cite{yu2021mask}, before inserting them into the target images. The manipulated objects were placed along the X and Y axes to make them look more realistic. We varied the size, and rotation of the donor pixels and randomly added JPEG compression (quality factor 50-99) before inserting them in target images in some cases. For the first and second types of combinations, we tried to choose the donor object to be small, flyable, and most commonly found to ensure a convincing representation. For example, a sports ball can be convincingly placed flying in any outdoor or indoor images.

In the fourth combination, we used the donor objects from MS-COCO and spliced them into target images from the camera identification datasets \cite{galdi2019socrates},  \cite{shullani2017vision},  \cite{hadwiger2021forchheim},  \cite{kcmi:online}, and  \cite{pathak2016context}  as presented in \cref{tab.trainfor_desc}, using a similar setup as the previous combinations.

% \subsubsection{Copy-move}
\noindent\textbf{Copy-move:} The copy-move manipulated image generation process is similar to image splicing, except that there is only a single image, and objects are duplicated in the same image. The camera identification datasets \cite{galdi2019socrates},  \cite{shullani2017vision},  \cite{hadwiger2021forchheim},  \cite{kcmi:online}, and  \cite{pathak2016context} could not be used for generating copy-move forgeries. We used two kinds of combinations of MS-COCO categories to create the copy-move images. In the first combination, a single category image was chosen and the object in that image was duplicated and inserted into a different position in the same image. For example, a bird's pixels were duplicated and inserted at a different location in the same image. Similar to splicing, we refined the segmented objects using \textit{MGMatting} \cite{yu2021mask}, before inserting the duplicated pixels along the X and Y axes, depending on the height and width of the objects, after resizing and rotation.

In the second combination, we extracted images with multiple objects from MS-COCO and duplicated one of the objects to insert them in the source image, similar to the first combination. We repeated the process with the other object in the same image to create another variant of copy-move images. For example, an image may contain a backpack and a handbag. We duplicated the backpack in the first set and then duplicated the handbag in the second set to create two instances of copy-move images.

% \subsubsection{Removal}
\noindent\textbf{Removal:} In the removal manipulated images, one or multiple objects are removed from an image and substituted by background pixels by inpainting. We have used MS-COCO and Paris-Street-View datasets to generate the removal images with three combinations. In the first combination, we simply choose images from any category, and after removing the object, we inpaint it with the background pixels. We used an exemplar-based blending method \cite{reshniak2020nonlocal} for inpainting. We tried to make sure that the chosen objects are not very cluttered or complex. For example, if we remove a hat from a person's head, it may look visually unrealistic if we cannot inpaint the hair and head properly. For the second combination, we selected MS-COCO images with multiple objects and repeated the removal procedure of the first combination with two or more objects, creating multiple instances from a single source image. In the third combination, we created removal images from \cite{pathak2016context} by removing regions from the images and inpainting them using \cite{reshniak2020nonlocal}

% \subsubsection{Image Enhancement}
\noindent\textbf{Image Enhancement:} We have performed a series of image enhancements using major image processing methods - adding noise, image morphing, image compression, image resizing, and image blurring. We added Gaussian Noise, Uniform Noise, Poisson Noise, and Impulse Noise. We performed Open Morphing, Erode Morphing, Dilate Morphing and Closed Morphing. We did Area Resize, Cubic Resize, Lanczo Resize, Linear Resize, and Nearest Resize. We created images with JPEG Compression, JPEG Double Compression, and WEBP Compression. We added Box Blur, Gaussian Blur, Median Blur, and Wavelet Blur. Finally, we created images with Quantization, Dithering, Posterization, Histogram Equalization, and Auto Contrast. We used opencv functions to perform the image enhancements and applied \textit{MGMatting} \cite{yu2021mask} blending when required in certain cases to make the enhanced images look natural.

% \subsubsection{Pristine Images}
\noindent\textbf{Pristine Images:} Image manipulation detection training requires negative samples, represented by unmanipulated pristine images. We have used a sample set of original images from [\cite{lin2014microsoft}, \cite{galdi2019socrates},  \cite{shullani2017vision},  \cite{hadwiger2021forchheim},  \cite{kcmi:online}, and  \cite{pathak2016context}] datasets to create the non-manipulated pristine image set of 200K. 

In total, TrainFors have 800K manipulated images (positive samples) and 200K pristine images (negative samples) as depicted in \cref{tab.trainfor_desc}.

% \subsection{Challenges in IMDL Forensics}

\vspace{-5px}
\section{Experimental Evaluation}
\vspace{-5px}

\begin{table*}%[!h]
% \begin{table}
    \vspace{-5px}
    \centering
    \small
    \setlength{\tabcolsep}{11.5pt}

    \scalebox{0.7}{
    \begin{tabular}{l |c|c| c|c| c|c| c|c| c|c}
    \toprule

     & \multicolumn{2}{c}{\bf{Columbia\cite{ng2009columbia}}} & \multicolumn{2}{c}{\bf{Coverage\cite{wen2016coverage}}} &
     \multicolumn{2}{c}{\bf{CASIAv1\cite{dong2013casia}}} & \multicolumn{2}{c}{\bf{NIST16\cite{NimbleCh33:online}}} & \multicolumn{2}{c}{\bf{IMD20\cite{novozamsky2020imd2020}}} \\

    % \midrule
    \cmidrule(lr){2-3} \cmidrule(lr){4-5} \cmidrule(lr){6-7} \cmidrule(lr){8-9} \cmidrule(lr){10-11}
    
    \bf{Method} & \bf{AUC} & \bf{F1} & \bf{AUC} & \bf{F1} & \bf{AUC} & \bf{F1} & \bf{AUC} & \bf{F1} & \bf{AUC} & \bf{F1} \\
    
    \midrule
    \textbf{Author-Specified Backbone} \\
    \cmidrule(lr){1-1}
    MVSS-Net\cite{chen2021image} & 61.1$\pm$1.2 & 52.9$\pm$1.1 & 51.9$\pm$1.3 & 36.8$\pm$1.3 & 54.6$\pm$1.7 & 35.1$\pm$1.8 & 38.7$\pm$1.8 & 19.3$\pm$2.1 & 47.9$\pm$1.7 & 30.2$\pm$1.6 \\
    \midrule
    Cat-Net\cite{kwon2021cat} & 56.3$\pm$2.1 & 44.7$\pm$2.3 & 23.4$\pm$1.9 & 7.6$\pm$0.7 & 28.7$\pm$1.6 & 8.9$\pm$0.6 & 29.6$\pm$1.4 & 11.4$\pm$1.1 & 19.3$\pm$1.6 & 6.4$\pm$0.3 \\
    \midrule
    PSCCNet\cite{liu2021pscc} & 58.7$\pm$1.3 & 49.4$\pm$1.2 & 63.9$\pm$2.6 & 46.8 $\pm$2.3 & 60.9 $\pm$1.1 & 42.9$\pm$0.9 & 48.6 $\pm$0.8 & 29.3 $\pm$1.3 & 41.3$\pm$1.7 & 28.5$\pm$2.2 \\
    \midrule
    ObjectFormer\cite{wang2022objectformer} & 55.8$\pm$1.1 & 46.3$\pm$1.1 & 64.2$\pm$1.8 & 47.2$\pm$1.7 & 61.3$\pm$1.6 & 43.2$\pm$1.1 & 48.7$\pm$1.2 & 29.4$\pm$1.1 & 42.5$\pm$1.2 & 28.7$\pm$1.3 \\   

    \midrule
    \textbf{EfficientNetV2 \cite{tan2021efficientnetv2} Backbone} \\
    \cmidrule(lr){1-1}
    MVSS-Net\cite{chen2021image} & 68.3$\pm$1.2 & 58.4$\pm$1.2 & 56.7$\pm$1.3 & 42.6$\pm$1.3 & 59.3$\pm$1.8 & 41.5$\pm$1.7 & 42.1$\pm$1.9 & 24.4$\pm$2.1 & \bf{53.2$\pm$1.6} & \bf{34.1$\pm$1.6} \\
    \midrule
    Cat-Net\cite{kwon2021cat} & \bf{70.4$\pm$1.4} & \bf{64.2$\pm$1.9} & 25.6$\pm$1.1 & 9.2$\pm$0.9 & 33.5$\pm$1.8 & 10.6$\pm$0.8 & 37.2$\pm$1.5 & 14.8$\pm$1.2 & 24.8$\pm$1.8 & 9.2$\pm$0.5  \\
    \midrule
    PSCCNet\cite{liu2021pscc} & 62.7$\pm$1.3 & 53.4$\pm$1.3 & \bf{67.1$\pm$1.9} & \bf{50.8$\pm$2.1} & \bf{63.8$\pm$1.2} & \bf{46.7$\pm$1.1} & \bf{52.6$\pm$1.2} & \bf{35.7$\pm$1.1} & 44.5$\pm$1.6 & 32.6$\pm$2.3 \\
    \midrule
    ObjectFormer\cite{wang2022objectformer} & 55.8$\pm$1.1 & 46.3$\pm$1.1 & 64.2$\pm$1.8 & 47.3$\pm$1.7 & 61.3$\pm$1.6 & 43.2$\pm$1.1 & 48.7$\pm$1.2 & 29.4$\pm$1.1 & 42.5$\pm$1.2 & 28.7$\pm$1.3 \\    
    
    \bottomrule
    \end{tabular}
    }
    \vspace{-5px}
    \caption{Manipulation Localization AUC (\%) and F1 (\%) scores of \textbf{Pre-trained models}, when trained with author-specified backbone networks and EfficientNetV2 \cite{tan2021efficientnetv2} backbone network respectively: Upper and Lower limits over 6 runs}
    \label{tab.pretrained_localization}
% \end{table}
\end{table*}

\begin{table*}%[!h]
% \begin{table}
    \vspace{-5px}
    \centering
    \small
    \setlength{\tabcolsep}{11.5pt}

    \scalebox{0.7}{
    \begin{tabular}{l |c|c| c|c| c|c| c|c| c|c}
    \toprule

     & \multicolumn{2}{c}{\bf{Columbia\cite{ng2009columbia}}} & \multicolumn{2}{c}{\bf{Coverage\cite{wen2016coverage}}} &
     \multicolumn{2}{c}{\bf{CASIAv1\cite{dong2013casia}}} & \multicolumn{2}{c}{\bf{NIST16\cite{NimbleCh33:online}}} & \multicolumn{2}{c}{\bf{IMD20\cite{novozamsky2020imd2020}}} \\

    % \midrule
    \cmidrule(lr){2-3} \cmidrule(lr){4-5} \cmidrule(lr){6-7} \cmidrule(lr){8-9} \cmidrule(lr){10-11}
    
    \bf{Method} & \bf{AUC} & \bf{F1} & \bf{AUC} & \bf{F1} & \bf{AUC} & \bf{F1} & \bf{AUC} & \bf{F1} & \bf{AUC} & \bf{F1} \\
    
    \midrule
    \textbf{Author-Specified Backbone} \\
    \cmidrule(lr){1-1}
    MVSS-Net\cite{chen2021image} & 71.2$\pm$1.3 & 63.8$\pm$1.3 & 62.9$\pm$1.2 & 45.3$\pm$1.1 & 63.3$\pm$1.8 & 45.2$\pm$1.8 & 47.8$\pm$1.9 & 29.2$\pm$1.7 & 56.6$\pm$1.7 & 39.8$\pm$1.7 \\
    \midrule
    Cat-Net\cite{kwon2021cat} & 67.8$\pm$2.3 & 55.5$\pm$2.1 & 32.3$\pm$1.2 & 12.9$\pm$0.6 & 33.7$\pm$1.4 & 13.6$\pm$1.1 & 41.3$\pm$1.8 & 17.9$\pm$1.5 & 28.4$\pm$1.6 & 10.1$\pm$0.9 \\
    \midrule
    PSCCNet\cite{liu2021pscc} & 69.5$\pm$1.2 & 60.6$\pm$1.3 & 72.3$\pm$2.1 & 57.4$\pm$2.1 & 71.6$\pm$1.6 & 51.2$\pm$1.1 & 59.3$\pm$1.2 & 38.7$\pm$2.6 & 52.7$\pm$1.6 & 38.4$\pm$1.7 \\
    \midrule
    ObjectFormer\cite{wang2022objectformer} & 66.3$\pm$1.2 & 56.7$\pm$1.1 & 73.1$\pm$1.2 & 56.4$\pm$1.3 & 72.7$\pm$1.7 & 52.6$\pm$1.7 & 59.1$\pm$1.2 & 37.6$\pm$1.2 & 51.9$\pm$1.4 & 38.3$\pm$1.1 \\   

    \midrule
    \textbf{EfficientNetV2 \cite{tan2021efficientnetv2} Backbone} \\
    \cmidrule(lr){1-1}
    MVSS-Net\cite{chen2021image} & 78.9$\pm$1.2 & 62.7$\pm$1.2 & 69.7$\pm$1.2 & 53.2$\pm$1.1 & 70.4$\pm$1.6 & 52.9$\pm$1.7 & 59.2$\pm$1.7 & 36.7$\pm$2.3 & \bf{62.3$\pm$1.8} & \bf{46.4$\pm$1.7} \\
    \midrule
    Cat-Net\cite{kwon2021cat} & \bf{79.5$\pm$1.7} & \bf{65.3$\pm$1.1} & 37.6$\pm$2.1 & 14.6$\pm$1.3 & 38.6$\pm$1.6 & 15.2$\pm$1.2 & 50.1$\pm$1.7 & 22.3$\pm$1.5 & 34.6$\pm$1.5 & 12.8$\pm$1.1 \\
    \midrule
    PSCCNet\cite{liu2021pscc} & 76.9$\pm$1.3 & 62.2$\pm$1.2 & \bf{78.8$\pm$2.2} & \bf{58.6$\pm$2.1} & \bf{76.7$\pm$1.3} & \bf{54.1$\pm$1.1} & \bf{62.3$\pm$1.4} & \bf{41.7$\pm$2.1} & 57.4$\pm$1.5 & 40.3$\pm$1.5 \\
    \midrule
    ObjectFormer\cite{wang2022objectformer} & 66.3$\pm$1.2 & 56.7$\pm$1.1 & 73.1$\pm$1.2 & 56.4$\pm$1.3 & 72.7$\pm$1.7 & 52.6$\pm$1.7 & 59.1$\pm$1.2 & 37.6$\pm$1.2 & 51.9$\pm$1.4 & 38.3$\pm$1.1 \\    
    
    \bottomrule
    \end{tabular}
    }
    \vspace{-5px}
    \caption{Manipulation Localization AUC(\%) and F1(\%) scores of \textbf{Fine-tuned models}, when trained with author-specified backbone networks and EfficientNetV2 \cite{tan2021efficientnetv2} backbone network respectively: Upper and Lower limits over 6 runs}
    \label{tab.finetuned_localization}
% \end{table}
\end{table*}

\begin{figure*}%[!h]
    \centering
    \settoheight{\tempdima}{\includegraphics[width=.15\linewidth]    {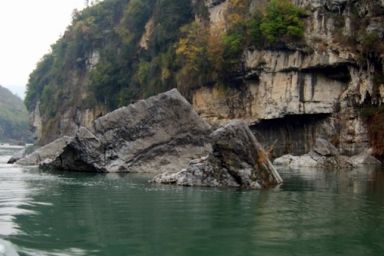}}
    
    \centering\begin{tabular}{@{ }c@{ }c@{ }c@{ }c@{ }c@{ }c@{ }c@{ }c@{ }}
    
    &\textbf{Pristine} & \textbf{Manipulated} & \textbf{Groundtruth} & \textbf{MVSS-Net\cite{chen2021image}} & \textbf{Cat-Net\cite{kwon2021cat}} & \textbf{PSCCNet\cite{liu2021pscc}} & \textbf{ObjectFormer\cite{wang2022objectformer}} \\
    
    \rowname{Casia\cite{dong2013casia}}&
    \includegraphics[width=.13\linewidth]{Images/FigVis/casia_orig1.jpg}& 
    \includegraphics[width=.13\linewidth]{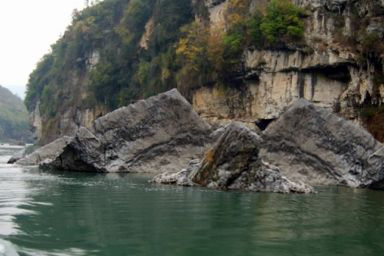}& 
    \includegraphics[width=.13\linewidth]{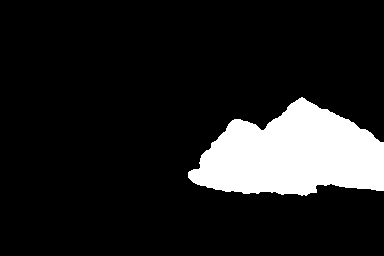}&
    \includegraphics[width=.13\linewidth]{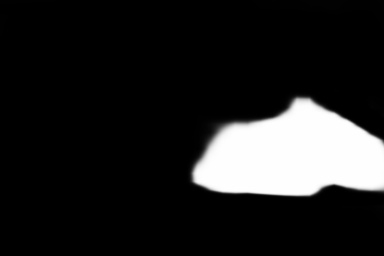}&
    \includegraphics[width=.13\linewidth]{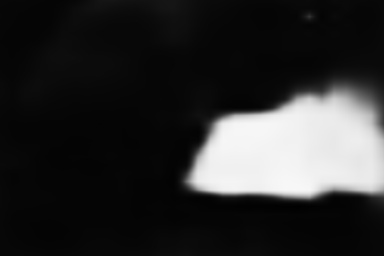}&
    \includegraphics[width=.13\linewidth]{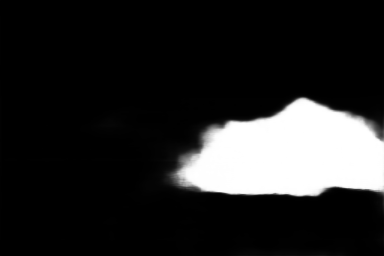}&
    \includegraphics[width=.13\linewidth]{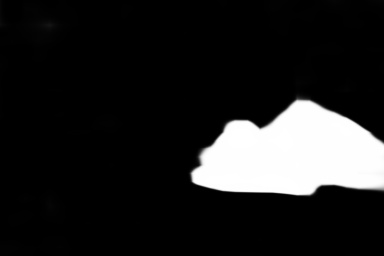}\\[-1ex] 
    
    \rowname{Columbia\cite{ng2009columbia}}&
    \includegraphics[width=.13\linewidth]{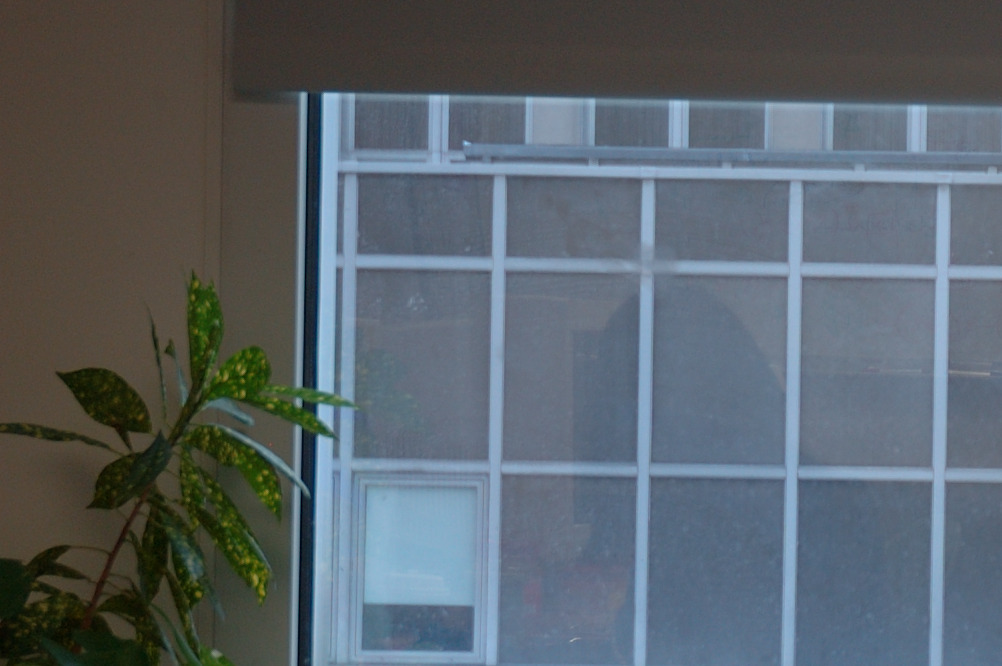}& 
    \includegraphics[width=.13\linewidth]{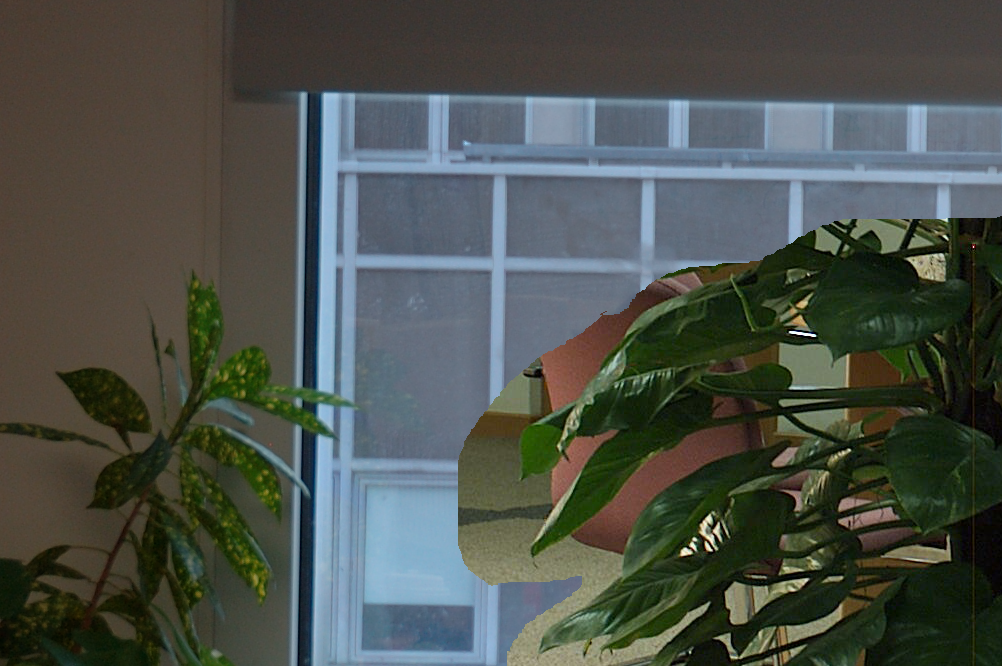}& 
    \includegraphics[width=.13\linewidth]{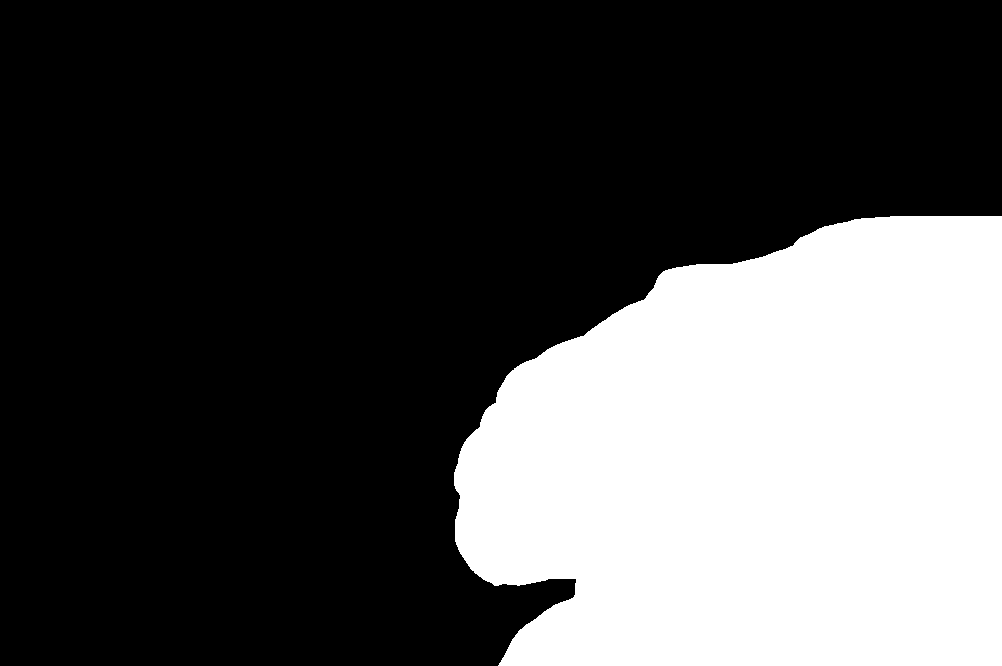}&
    \includegraphics[width=.13\linewidth]{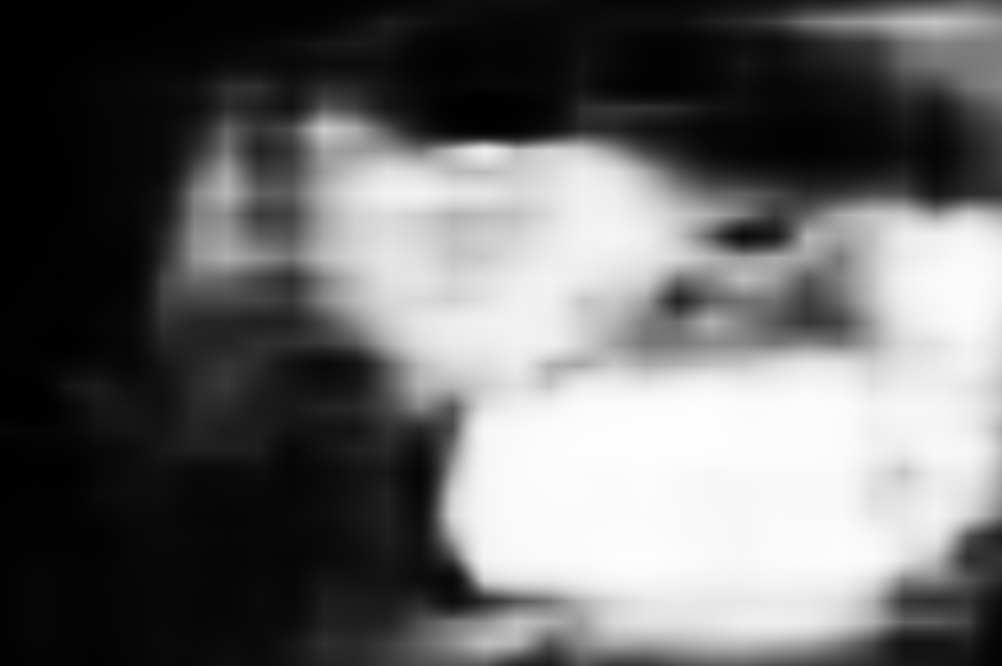}&
    \includegraphics[width=.13\linewidth]{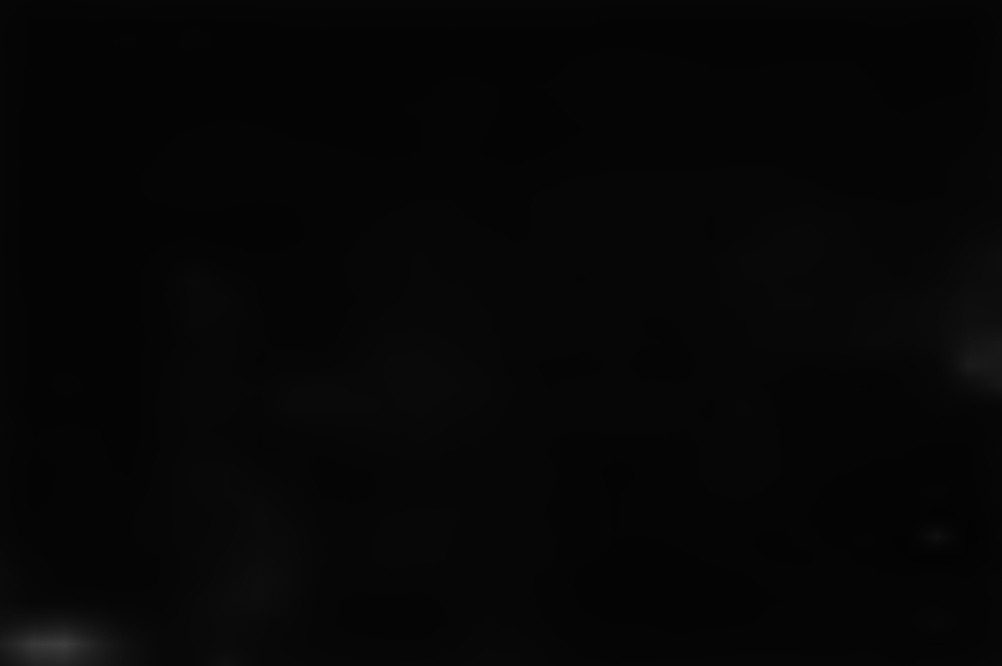}&
    \includegraphics[width=.13\linewidth]{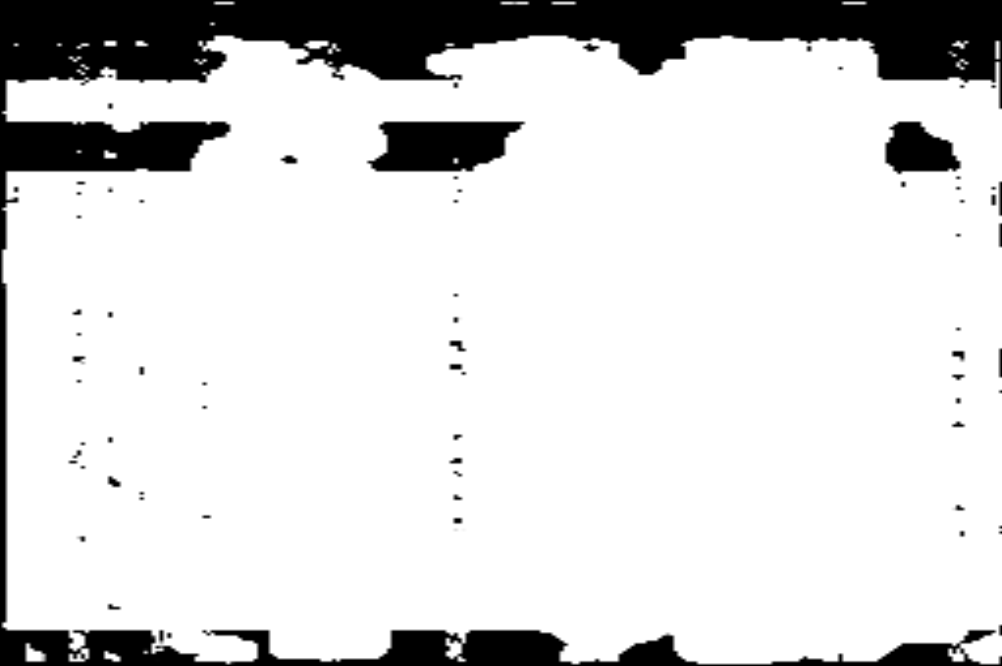}&
    \includegraphics[width=.13\linewidth]{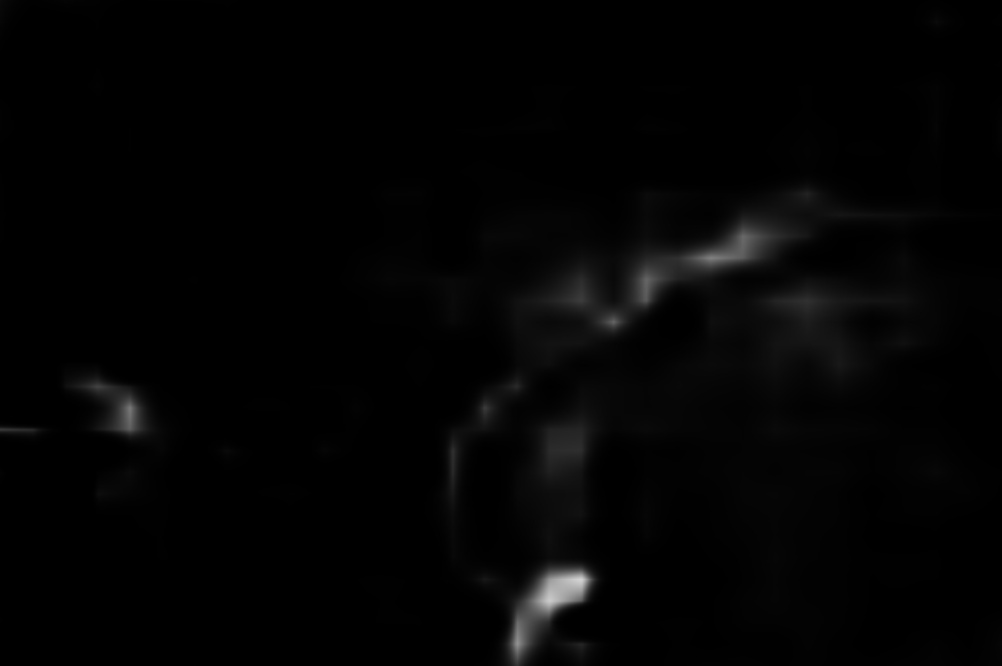}\\ 

    \rowname{Coverage\cite{wen2016coverage}}&
    \includegraphics[width=.13\linewidth]{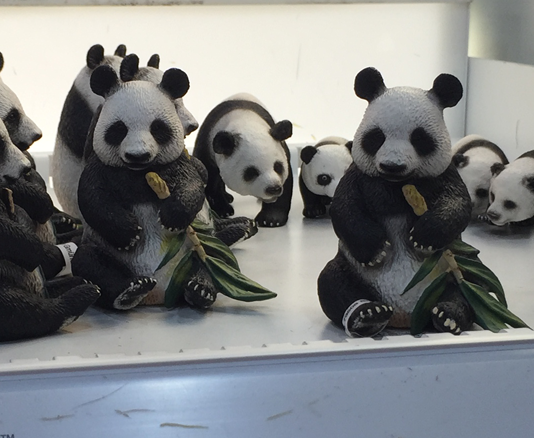}& 
    \includegraphics[width=.13\linewidth]{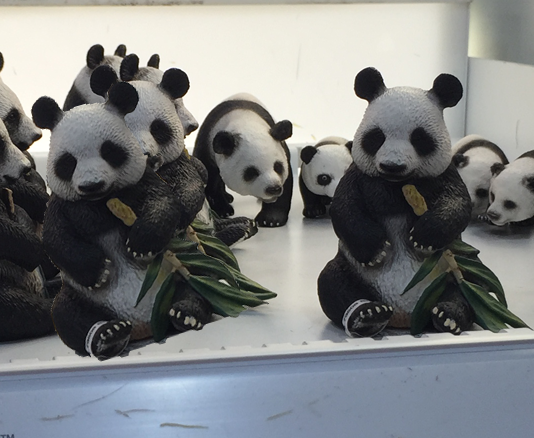}& 
    \includegraphics[width=.13\linewidth]{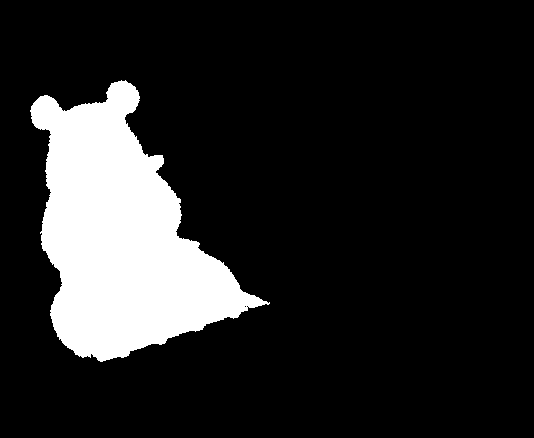}&
    \includegraphics[width=.13\linewidth]{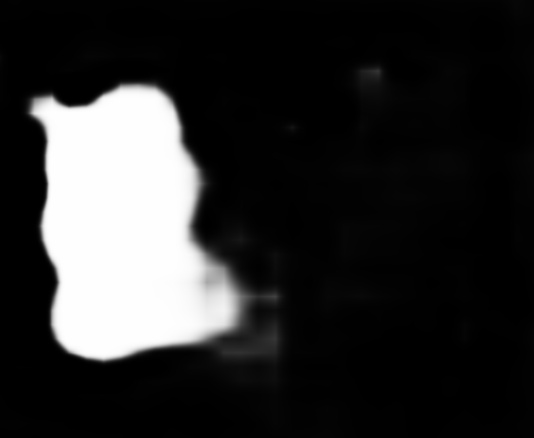}&
    \includegraphics[width=.13\linewidth]{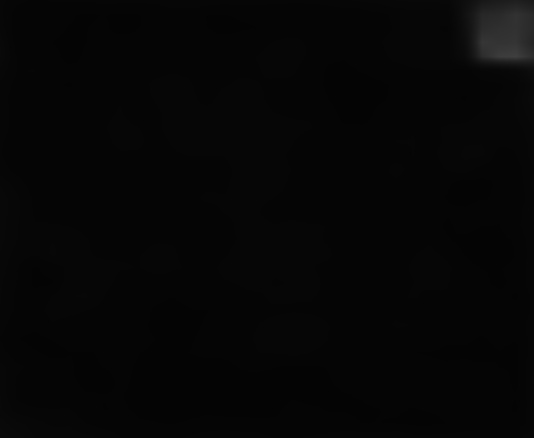}&
    \includegraphics[width=.13\linewidth]{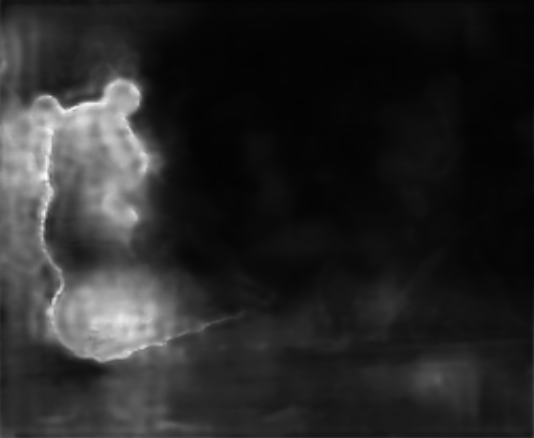}&
    \includegraphics[width=.13\linewidth]{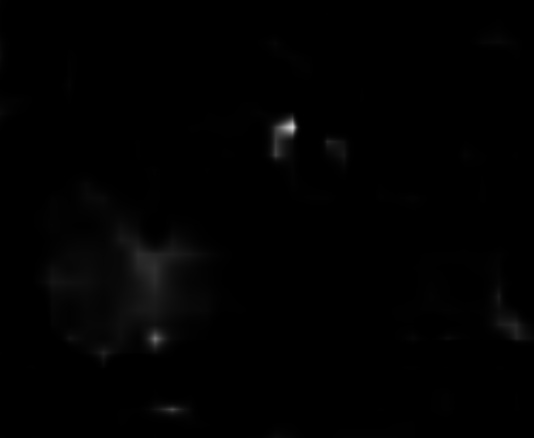}\\[-1ex] 

    \rowname{NIST16\cite{NimbleCh33:online}}&
    \includegraphics[width=.13\linewidth]{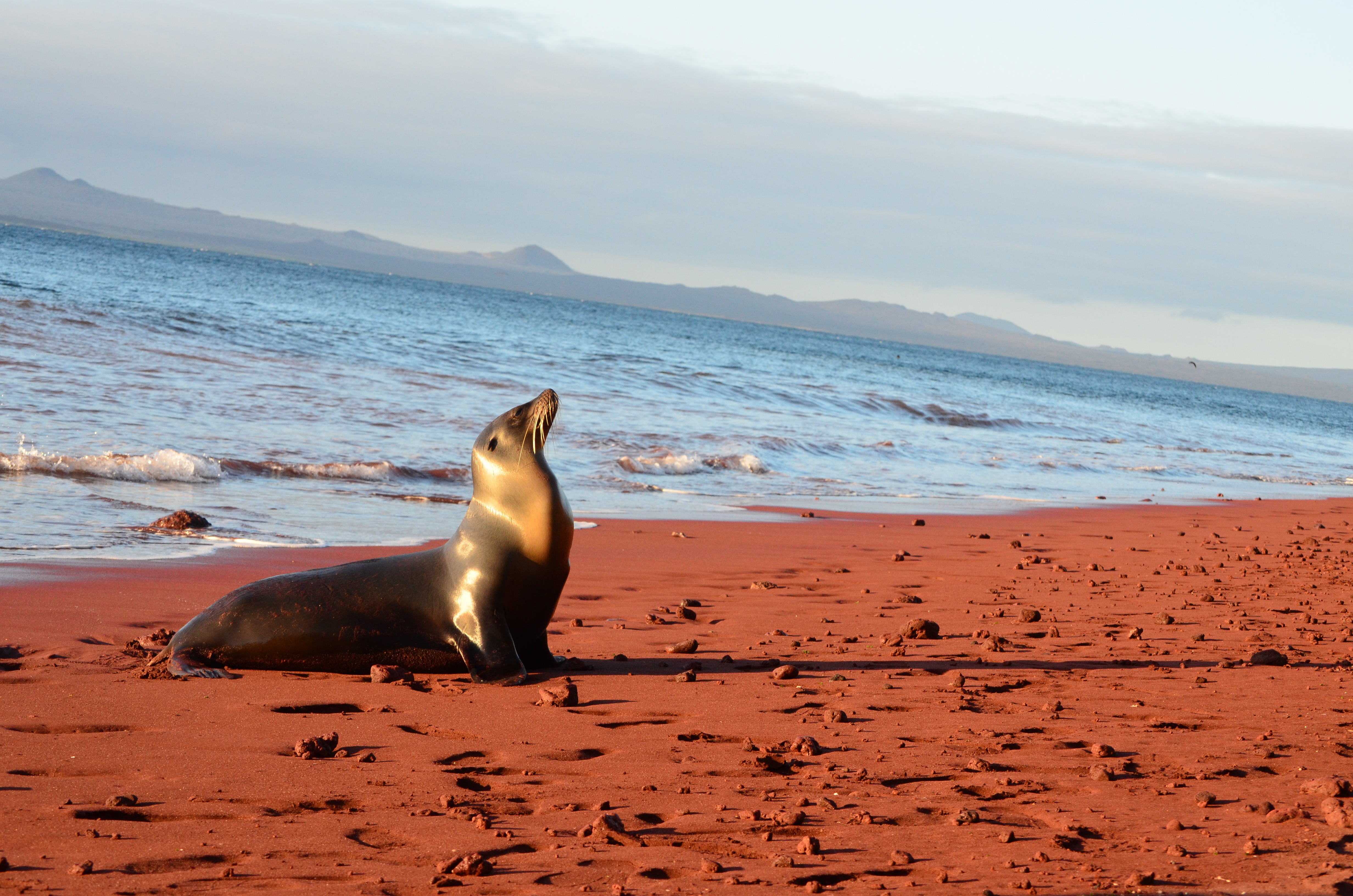}& 
    \includegraphics[width=.13\linewidth]{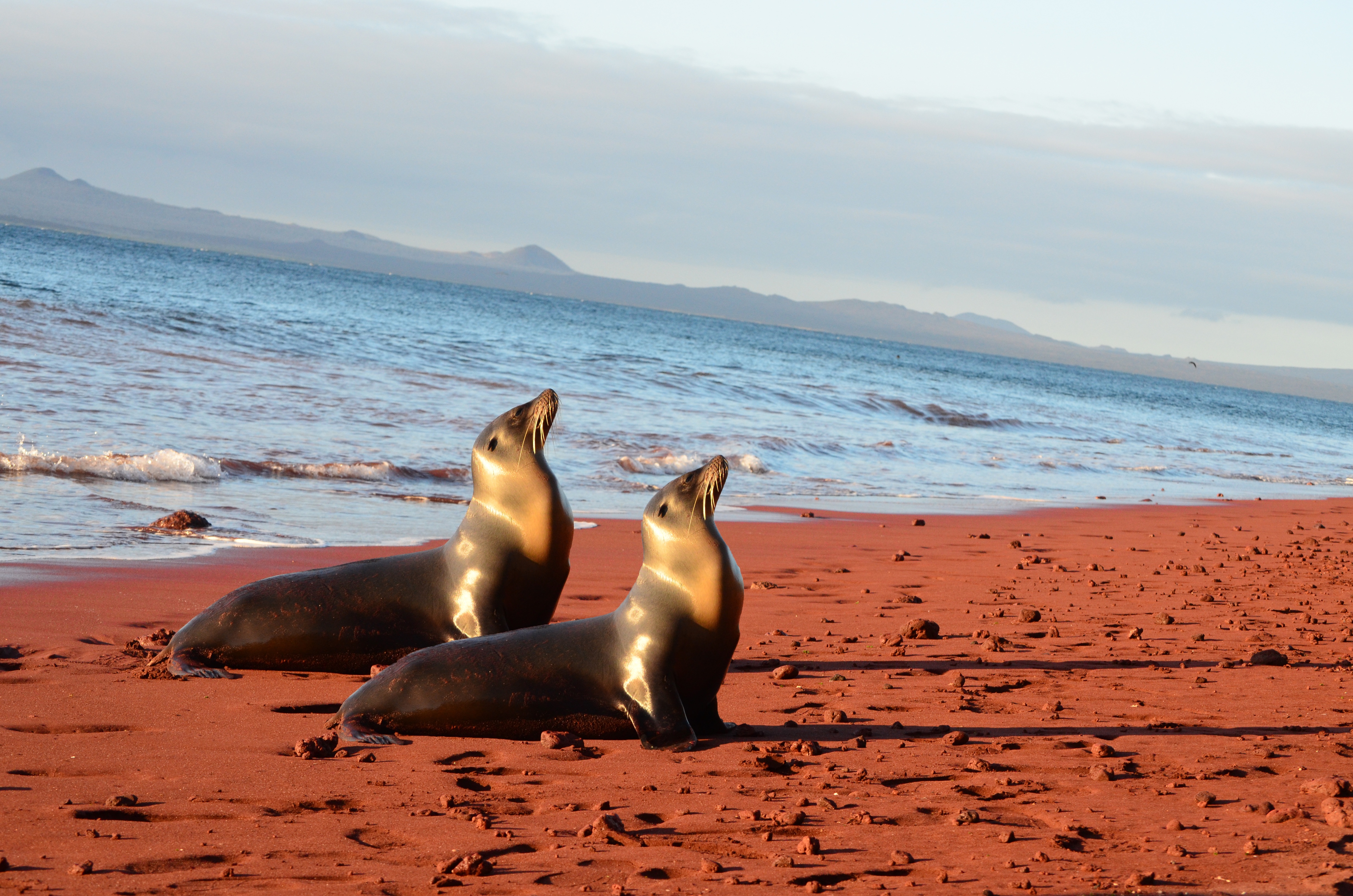}& 
    \includegraphics[width=.13\linewidth]{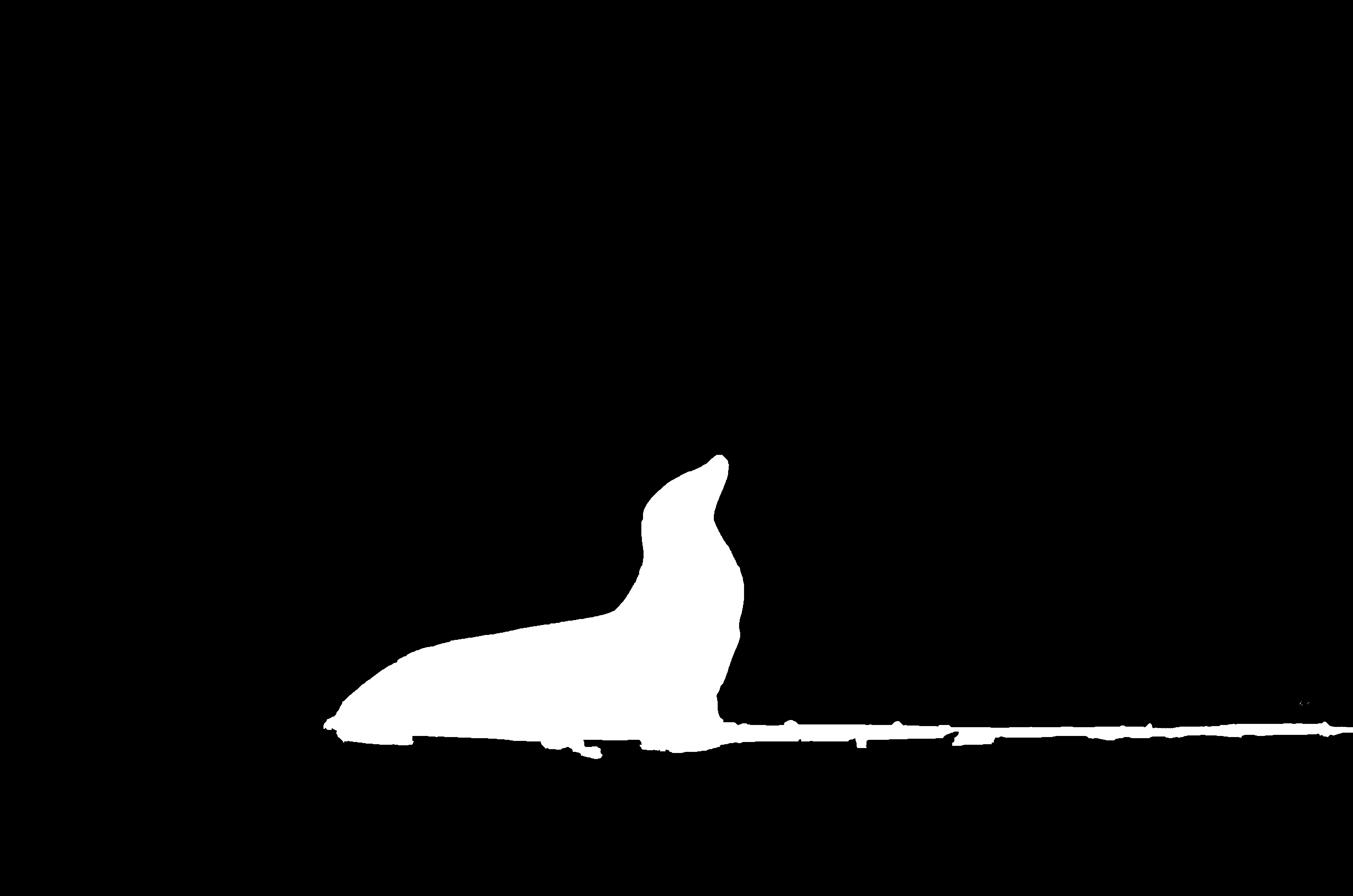}&
    \includegraphics[width=.13\linewidth]{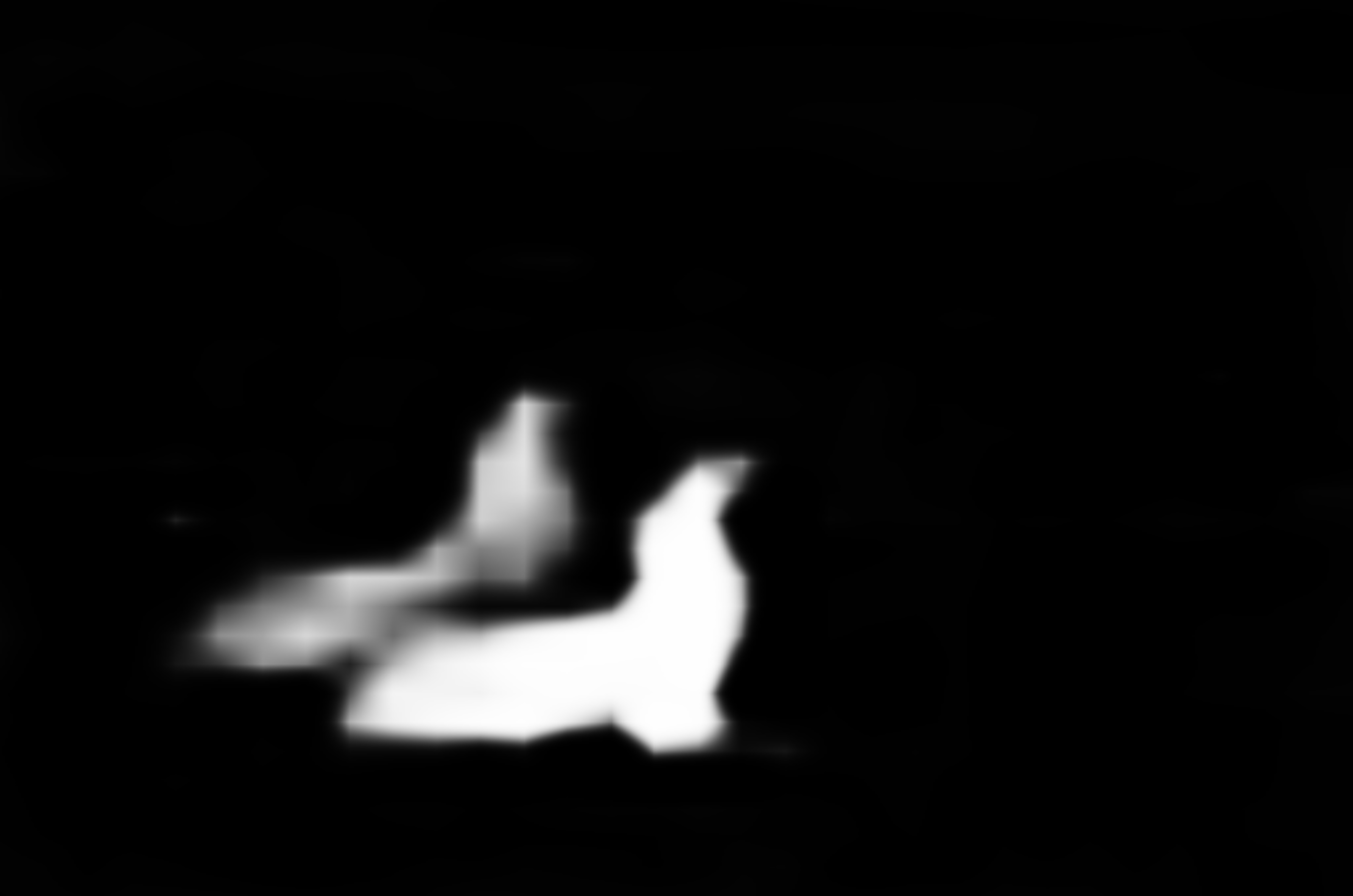}&
    \includegraphics[width=.13\linewidth]{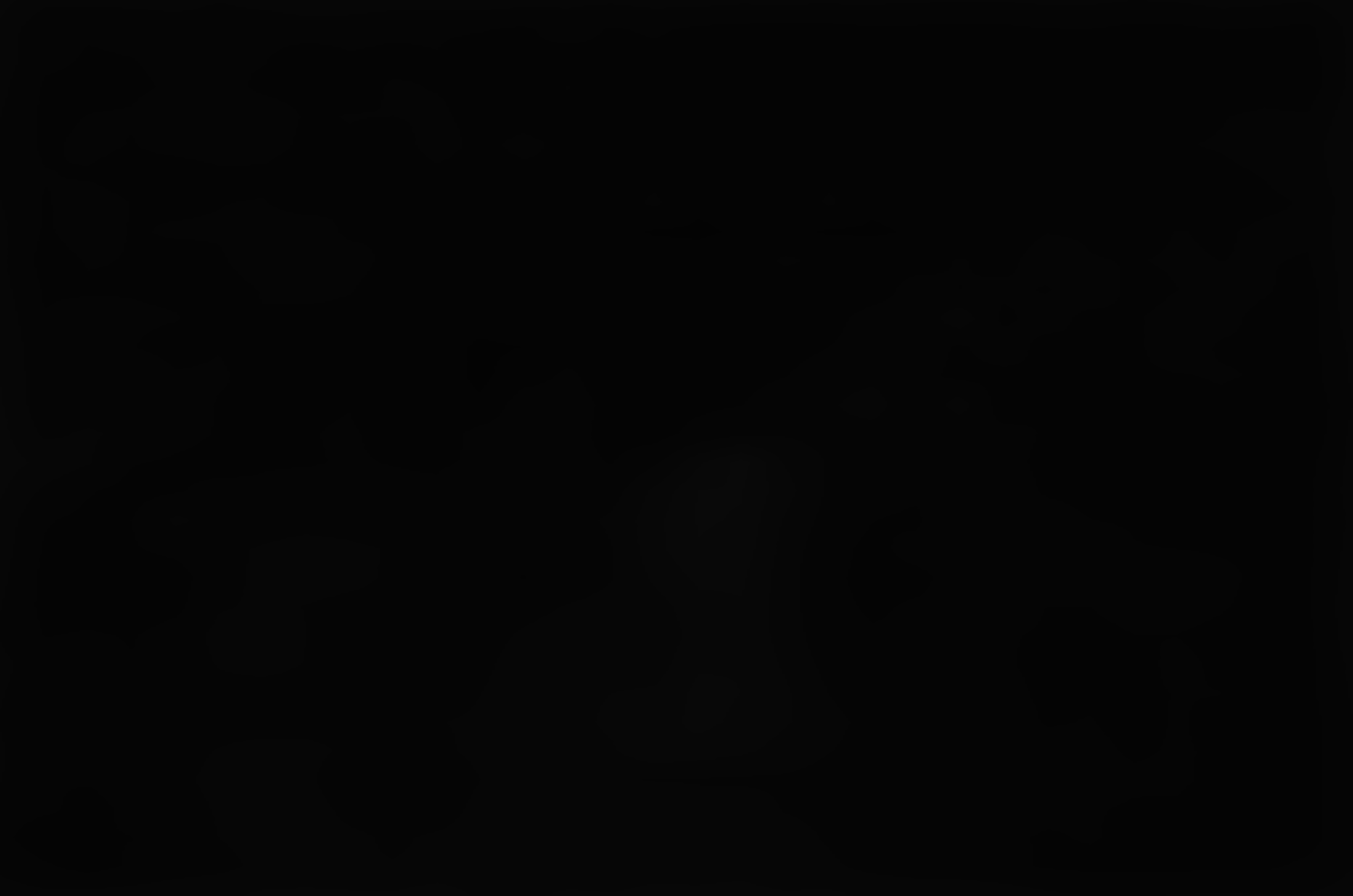}&
    \includegraphics[width=.13\linewidth]{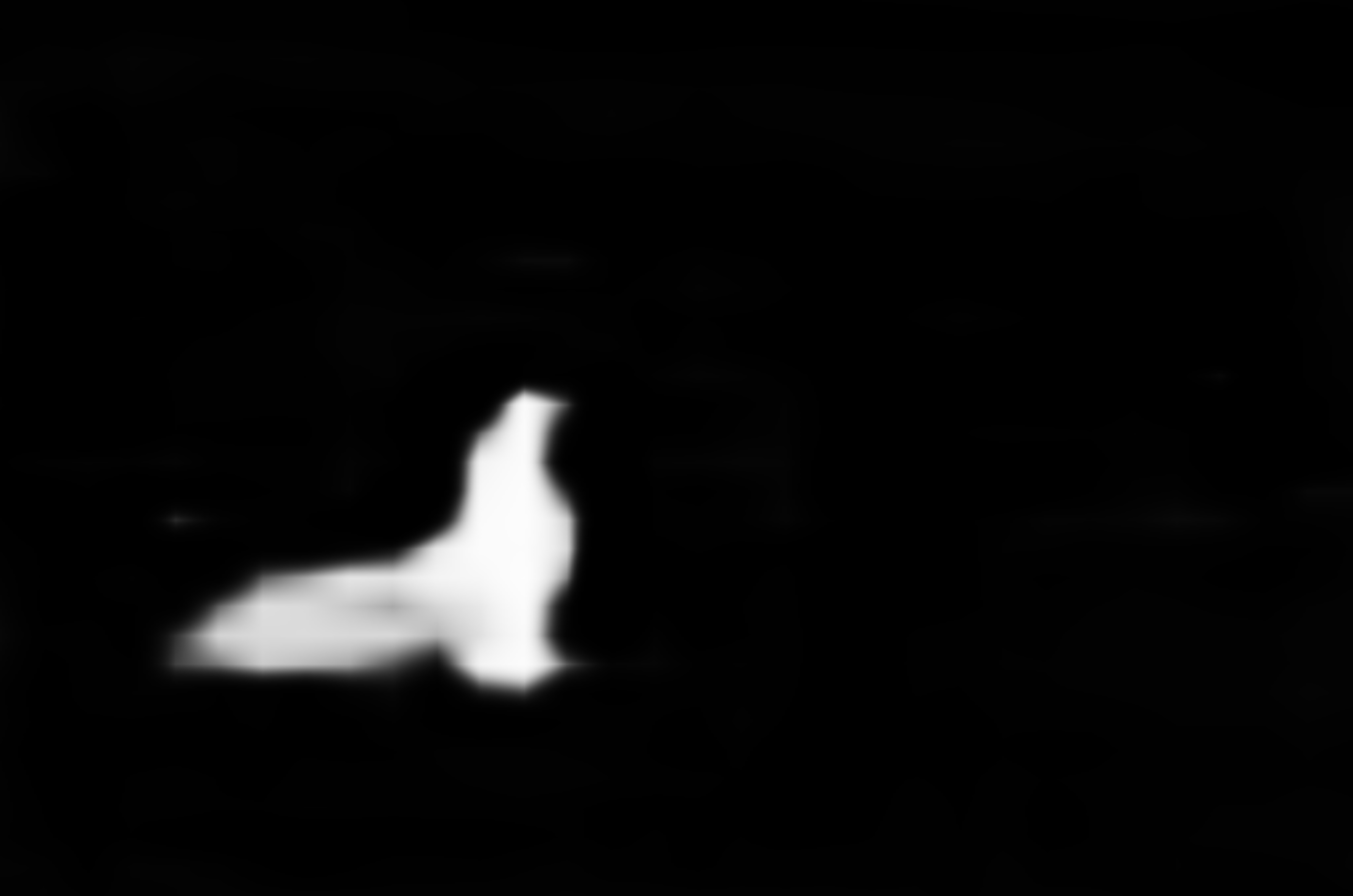}&
    \includegraphics[width=.13\linewidth]{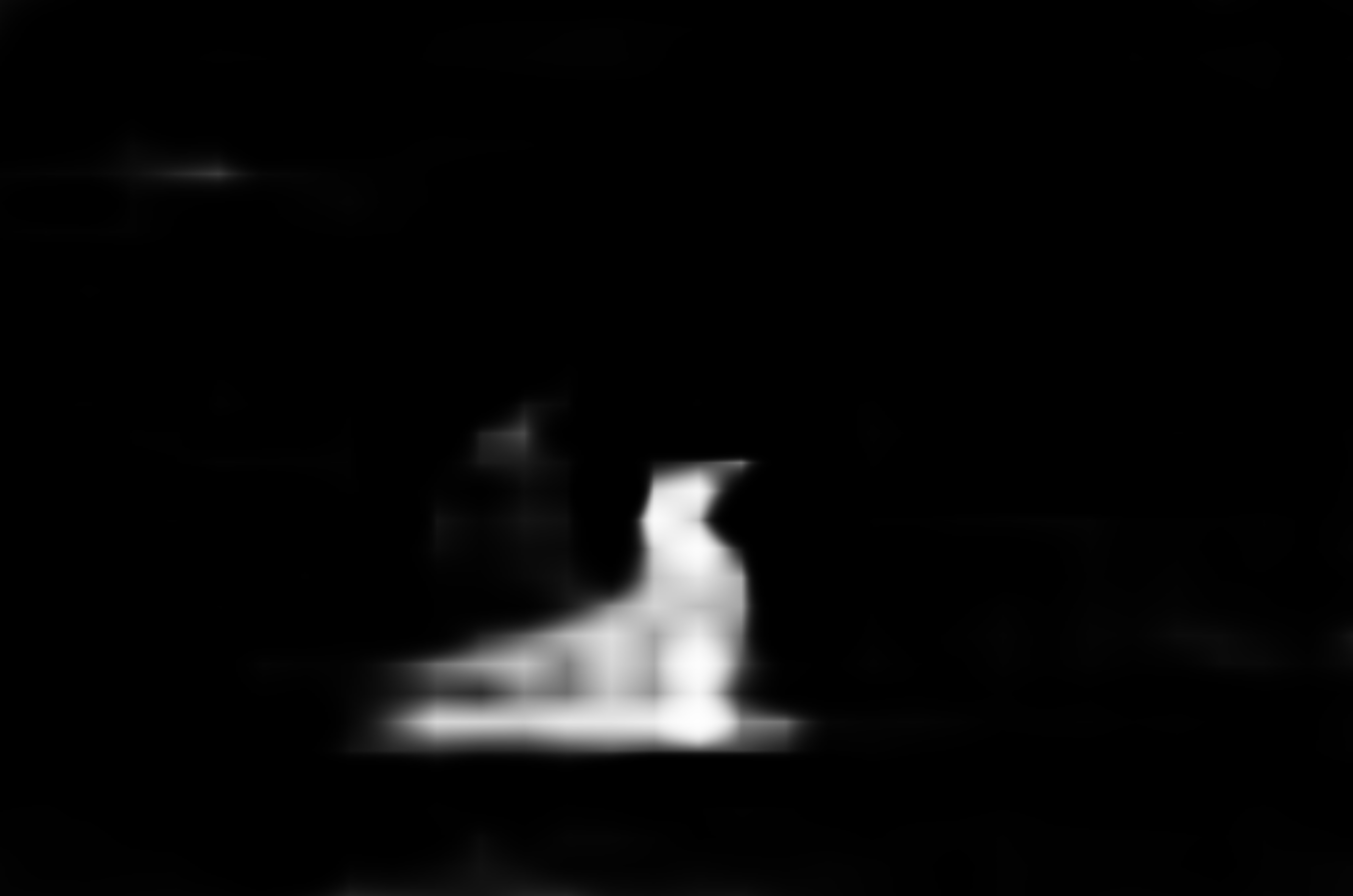}\\

    \rowname{IMD20\cite{novozamsky2020imd2020}}&
    \includegraphics[width=.13\linewidth]{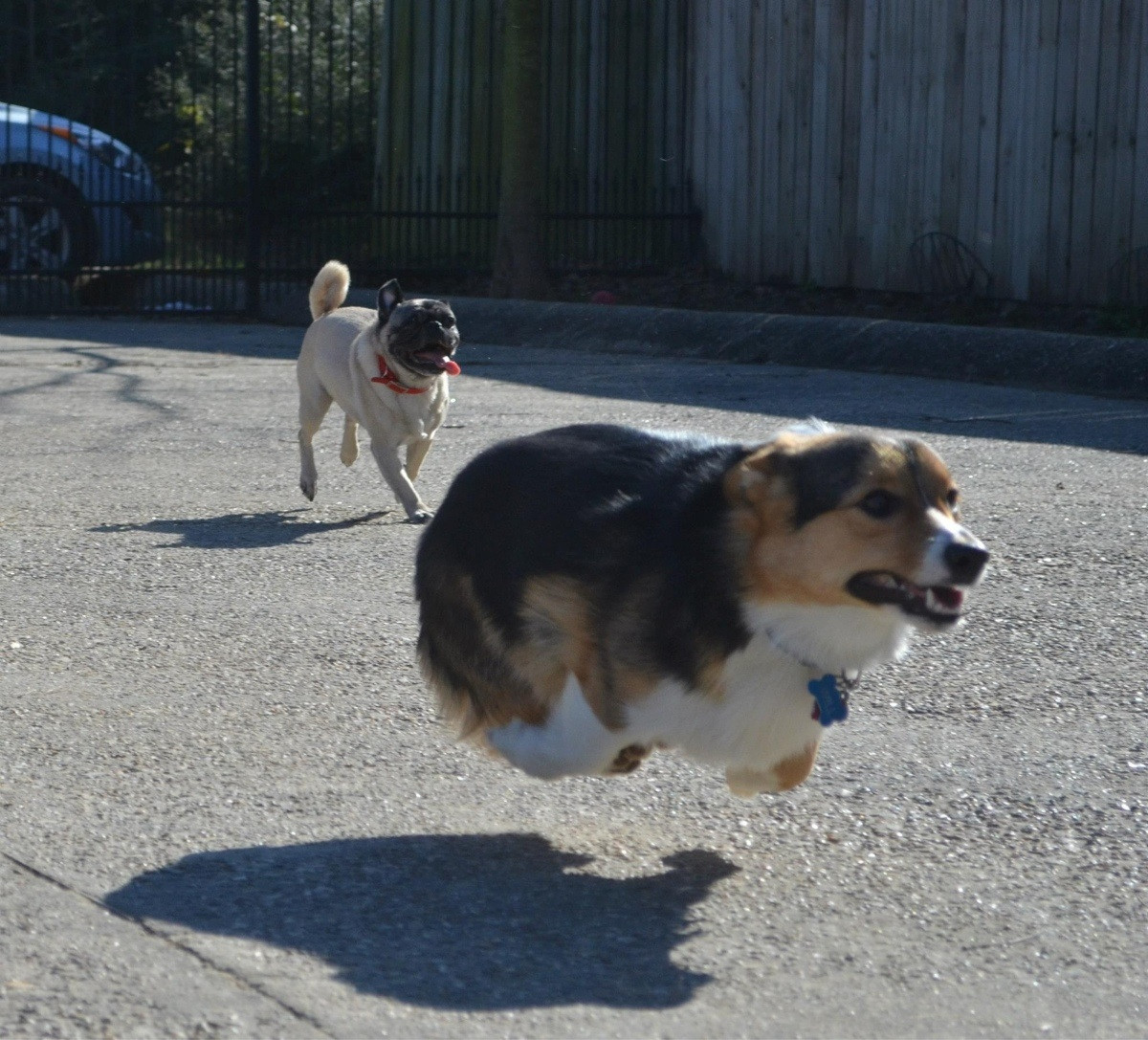}& 
    \includegraphics[width=.13\linewidth]{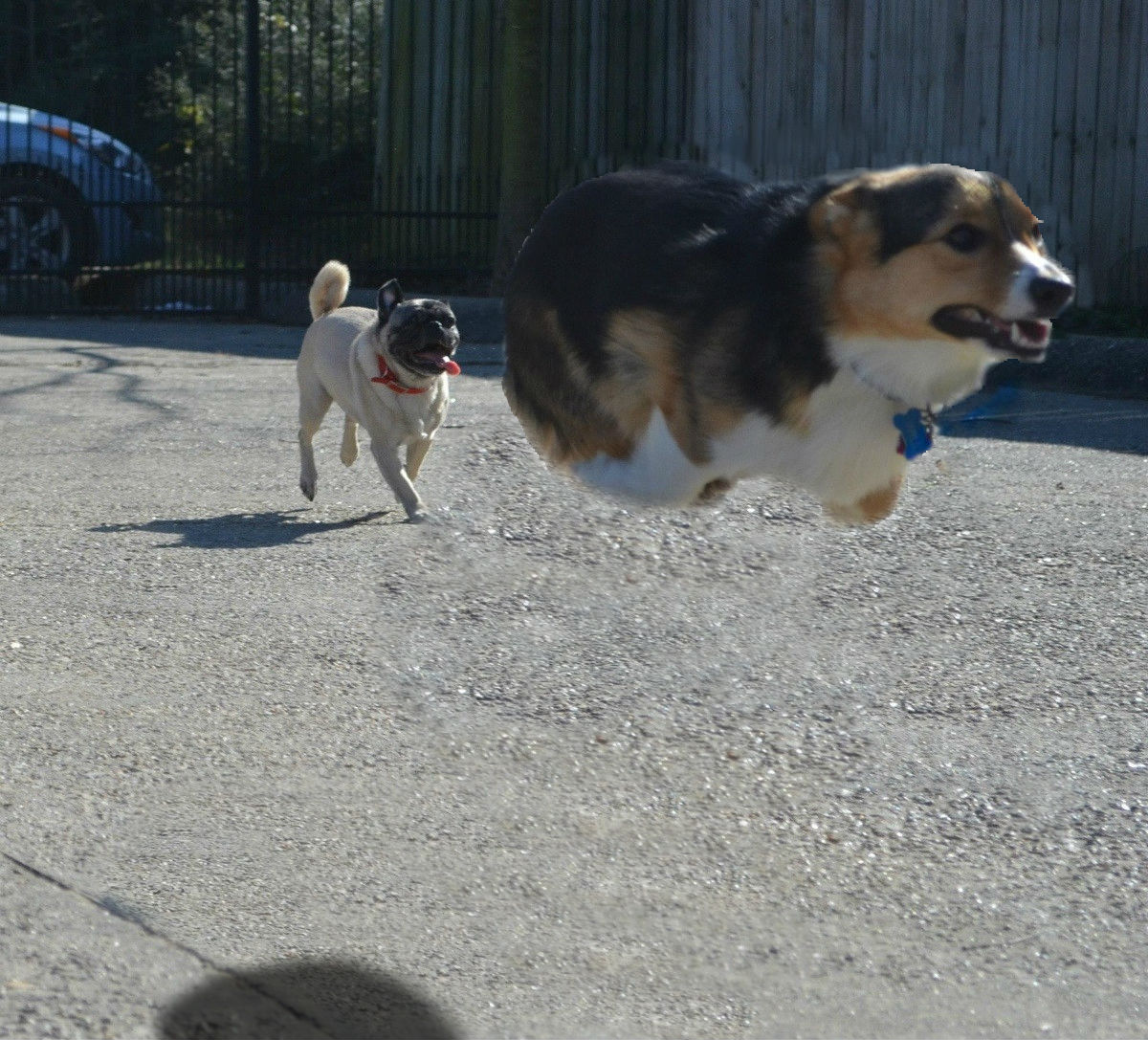}& 
    \includegraphics[width=.13\linewidth]{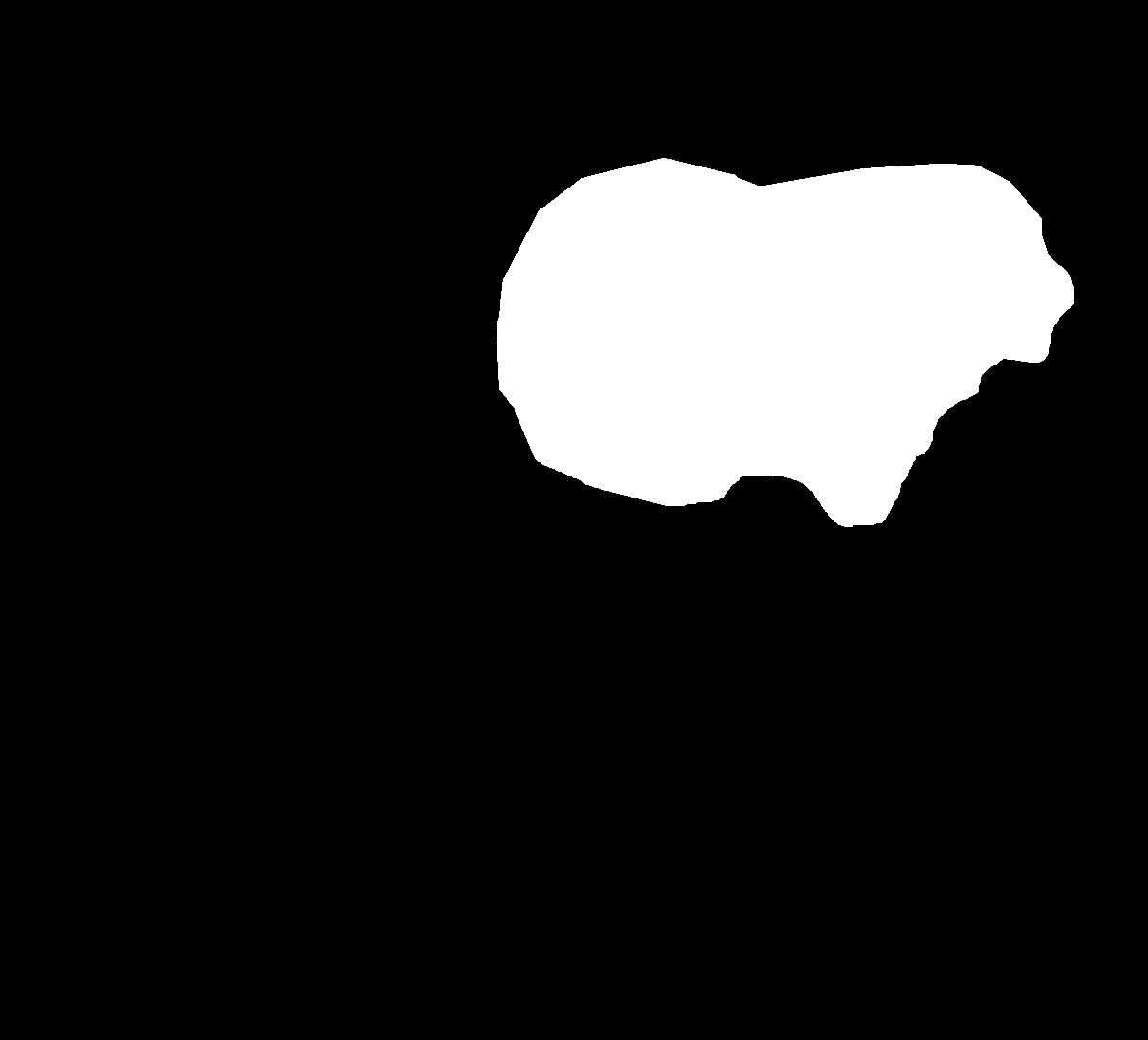}&
    \includegraphics[width=.13\linewidth]{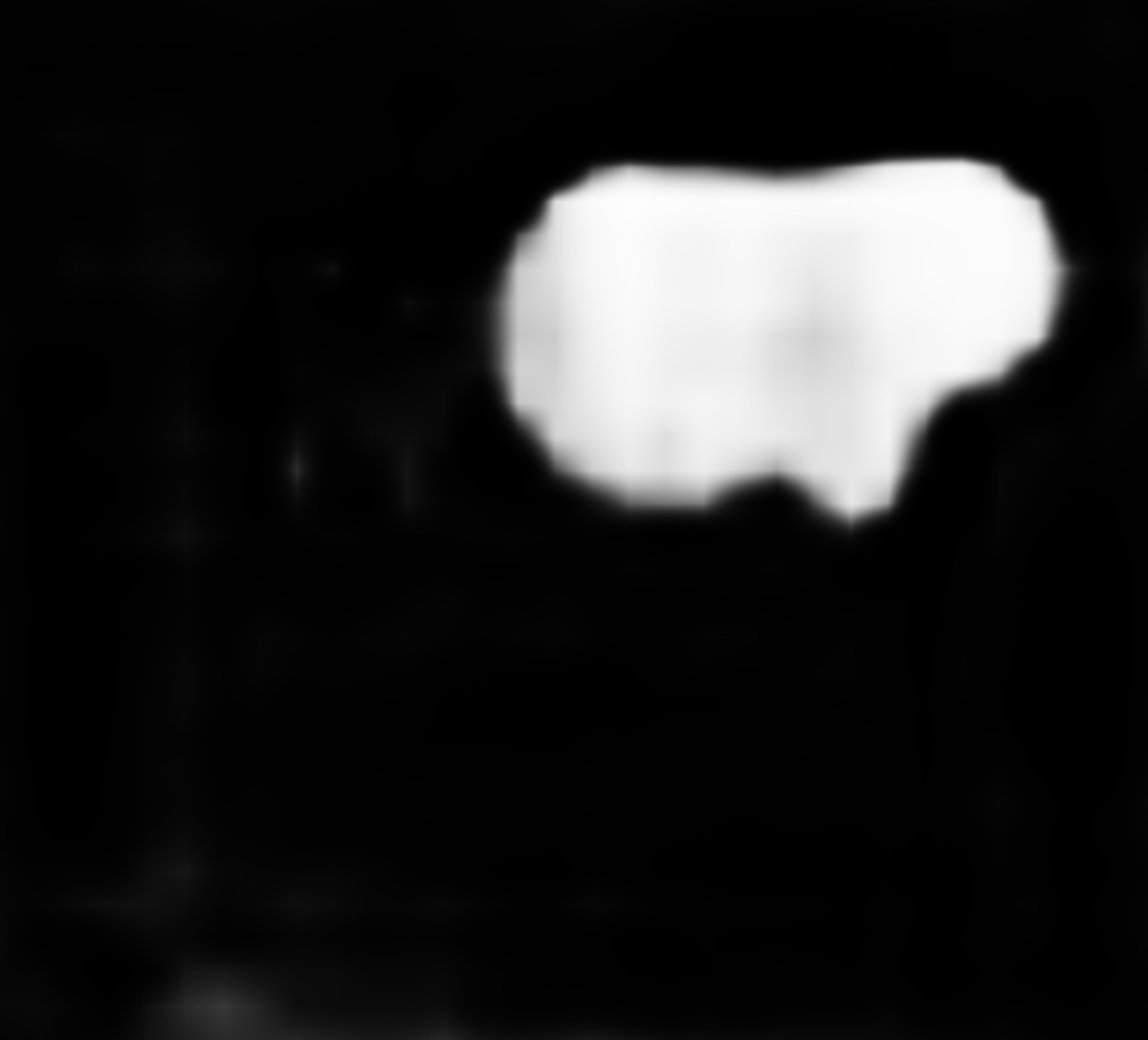}&
    \includegraphics[width=.13\linewidth]{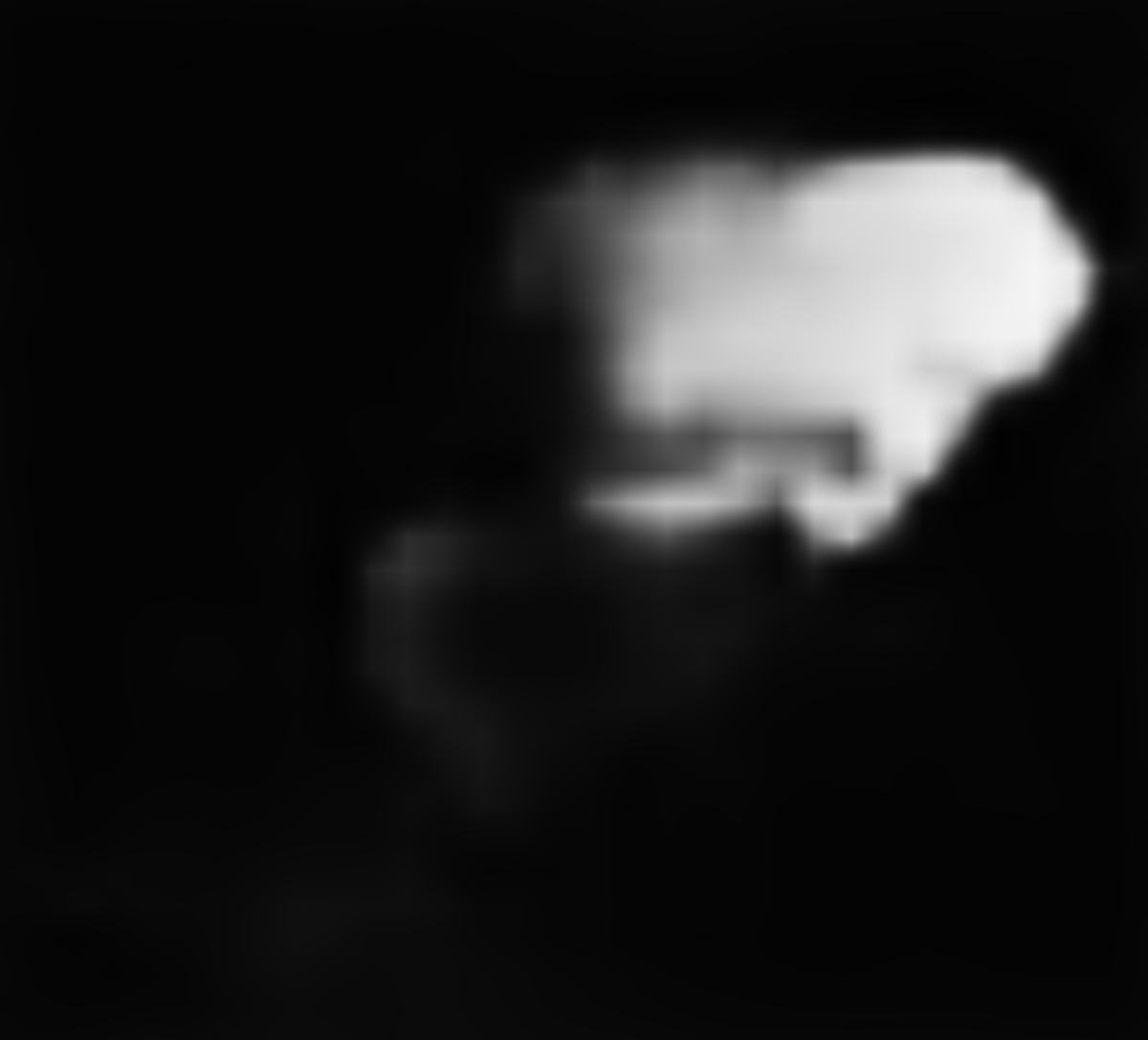}&
    \includegraphics[width=.13\linewidth]{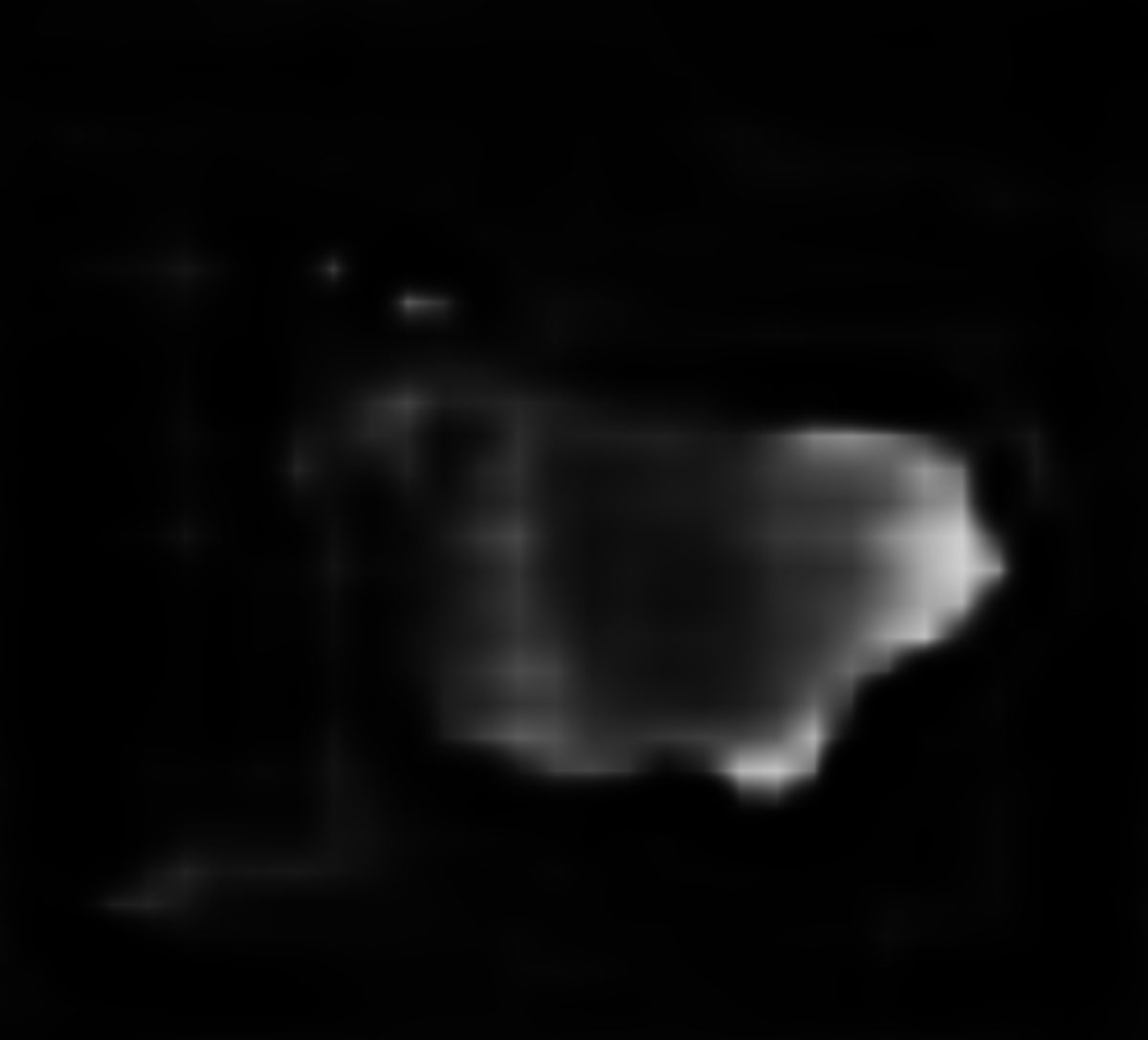}&
    \includegraphics[width=.13\linewidth]{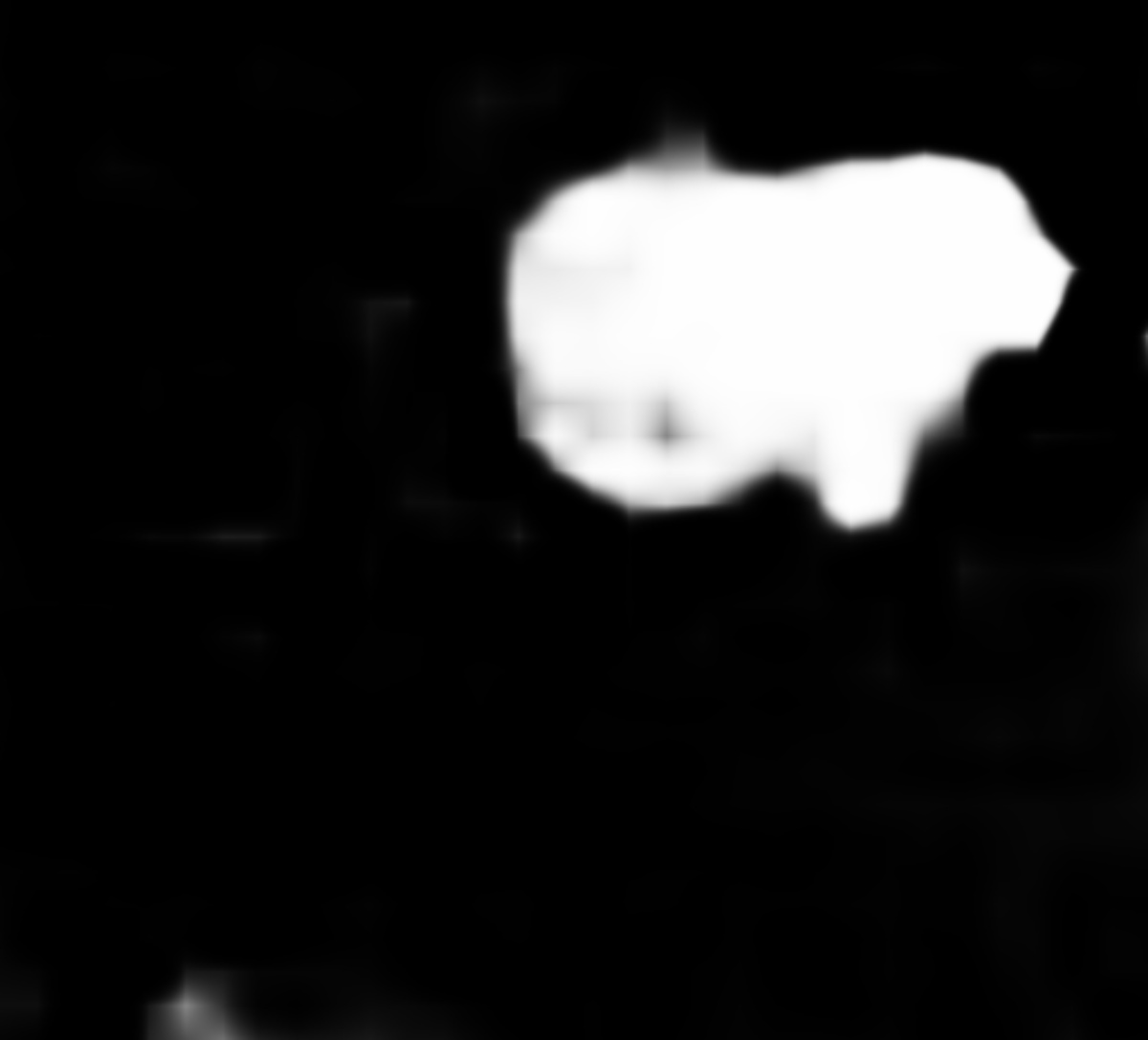}\\[-1ex] 
    
    \end{tabular}
    \caption{Image Manipulation Localization Prediction Visualization - From top to bottom, we show examples from 5 benchmark evaluation datasets: Casia, Columbia, Coverage Nist16, and IMD20 by 4 baseline IMDL models: MVSS-Net, Cat-Net, PSCC-Net, and ObjectFormer, trained on our proposed TrainFors dataset.}
     % , (d) Enhancement Manipulation}
    \label{fig:IMDL_Predictions}
\end{figure*}

\begin{table}
    \vspace{-5px}
    \centering
    \small
    \setlength{\tabcolsep}{3.5pt}

    \scalebox{0.7}{
    \begin{tabular}{l cc cc cc cc}
    \toprule

     & \multicolumn{2}{c}{\bf{Columbia\cite{ng2009columbia}}} & \multicolumn{2}{c}{\bf{Coverage\cite{wen2016coverage}}} &
     \multicolumn{2}{c}{\bf{CASIAv1\cite{dong2013casia}}} & \multicolumn{2}{c}{\bf{IMD20\cite{novozamsky2020imd2020}}} \\

    % \midrule
    \cmidrule(lr){2-3} \cmidrule(lr){4-5} \cmidrule(lr){6-7} \cmidrule(lr){8-9}
    
    \bf{Method} & \bf{AUC} & \bf{F1} & \bf{AUC} & \bf{F1} & \bf{AUC} & \bf{F1} & \bf{AUC} & \bf{F1} \\
    
    \midrule
    \textbf{Author-Specified Backbone} \\
    \cmidrule(lr){1-1}
    MVSS-Net\cite{chen2021image} & 82.1	& 62.3 & 42.6 &	24.3 & 78.6 & 59.8 & 65.7 & 35.1 \\
    Cat-Net\cite{kwon2021cat} & 81.6 & 61.7 & 37.4 & 29.8 & 64.7 & 38.0 & 59.1 & 35.4 \\
    PSCCNet\cite{liu2021pscc} & 83.4 & 63.8 & 81.5 & 62.8 & 84.6 & 76.9 & 78.6 & 57.3 \\
    ObjectFormer\cite{wang2022objectformer} & 84.8 & 64.5 & 82.7 & 63.9 & 86.1 & 77.4 & 79.3 & 58.2 \\   

    \midrule
    \textbf{EfficientNetV2 \cite{tan2021efficientnetv2} Backbone} \\
    \cmidrule(lr){1-1}
    MVSS-Net\cite{chen2021image} & \bf{85.6} & \bf{65.9} & 57.6 & 41.5 & 83.6 & 64.3 & \bf{69.8} & \bf{40.2} \\
    Cat-Net\cite{kwon2021cat} & 83.5 & 63.9 & 38.6 & 29.9 & 68.9 & 43.2 & 64.3 & 38.6 \\
    PSCCNet\cite{liu2021pscc} & 85.4 & 65.6 & \bf{83.6} & \bf{64.7} & \bf{87.2} & \bf{79.6} & 67.4 & 39.6 \\
    ObjectFormer\cite{wang2022objectformer} & 84.8 & 64.5 & 82.7 & 63.9 & 86.1 & 77.4 & 79.3 & 58.2 \\ 
    
    \bottomrule
    \end{tabular}}
    \vspace{-5px}
    \caption{Manipulation Detection AUC(\%) and F1(\%) scores on CASIA-D dataset\cite{dong2013casia}, when trained with author-specified backbone networks and EfficientNetV2 \cite{tan2021efficientnetv2} backbone network respectively}
    \label{tab.manipulation_detection}
\end{table}

We trained four state-of-the-art IMDL models on our proposed TrainFors dataset and evaluated the IMDL performance on five benchmark evaluation datasets. The evaluation mainly comprises two tasks - detecting manipulated images and in the latter task, we generate a manipulated pixel map of the positively detected manipulated images. 

% \vspace{-5px}
\subsection{\noindent\textbf{{Benchmark Evaluation Datasets}}}\label{eval_datasets}
\vspace{-5px}
We have evaluated the manipulation localization task on five benchmark datasets: \noindent\textbf{Columbia}\cite{ng2009columbia}, \noindent\textbf{Coverage}\cite{wen2016coverage}, \noindent\textbf{CASIAv1}\cite{dong2013casia}, \noindent\textbf{NIST16}\cite{NimbleCh33:online} and \noindent\textbf{IMD20}\cite{novozamsky2020imd2020}. Columbia\cite{ng2009columbia} is an image-splicing dataset, consisting of 180 images. Coverage\cite{wen2016coverage} is an image copy-move detection dataset composed of 100 images. CASIA\cite{dong2013casia} has both copy-move and splicing images: 5123 images in v2.0 and 921 images in v1.0. NIST16\cite{NimbleCh33:online} dataset includes splicing, copy-move, and image enhancement/reduction manipulations with 611 images. IMD20\cite{novozamsky2020imd2020} includes 2,010 manipulated images scraped from the internet. Refer to \cref{tab.trainfor_desc} for the detailed summary of the number of images of each type of manipulation for all the evaluation datasets.

% \vspace{-5px}
\subsection{\noindent\textbf{{Evaluation Metrics}}} 
\vspace{-5px}
We have evaluated the model performance for both image manipulation localization and detection tasks. For manipulation localization, pixel-level Area Under Curve (AUC) and F1 scores are reported. While for manipulation detection, image-level AUC and F1 scores are reported.

% \vspace{-5px}
\subsection{\noindent\textbf{{Baseline Models}}} \label{baseline_models}
\vspace{-5px}
Most of the previous works followed a two-step training method in generating the IMDL models. Firstly they pretrained a backbone network with a synthetic training dataset and then fine-tuned the pretrained model with the train-split of the evaluation datasets. \noindent\textbf{PSCCNet\cite{liu2021pscc}} \footnote{\href{https://github.com/proteus1991/PSCC-Net}{\textcolor{red}{https://github.com/proteus1991/PSCC-Net}}} used HRNetV2p-W18 \cite{wang2020deep} as a backbone network for pretraining their own synthetic data. \noindent\textbf{{ObjectFormer\cite{wang2022objectformer}}} used EfficientNetb4 \cite{tan2019efficientnet} as the backbone network. \noindent\textbf{MVSS-Net}\cite{chen2021image} \footnote{\href{https://github.com/dong03/MVSS-Net}{\textcolor{red}{https://github.com/dong03/MVSS-Net}}} used ResNet-50 \cite{he2016deep} as the backbone network. \noindent\textbf{CAT-Net} \cite{kwon2021cat} \footnote{\href{https://github.com/mjkwon2021/CAT-Net}{\textcolor{red}{https://github.com/mjkwon2021/CAT-Net}}} also used HRNet \cite{wang2020deep} as the backbone network. All the state-of-the-art models used varying input image sizes for pretraining the network. We ran two sets of experiments - firstly pretraining all the models on TrainFors using the backbone network specified by each of them respectively and then fine-tuned using the models from the inference code, if released by the authors, else we wrote the model code referencing their respective papers. In the second set of experiments, we fixed the backbone network as EfficientNetV2 \cite{tan2021efficientnetv2} (pretrained on ImageNet \cite{deng2009imagenet} weights) for all the models and fine-tuned it in a process similar to the first set of experiments. We were not able to report the performance of the early DNN-based models from \cref{early_DNN} and some models from \cref{recent_DNN} because either the code is not released or is no longer publicly accessible.

% \vspace{-5px}
\subsection{\noindent\textbf{{Implementation Details}}}
\vspace{-5px}
All the images in TrainFors are resized to 256X256 before feeding them to the backbone network. We implemented all the models in PyTorch and trained them on NVIDIA GeForce RTX 2080 Ti GPU. We used Adam \cite{kingma2014adam} optimizer with a batch size of 24 and a learning rate periodically varying between \num{1e-3} to \num{1e-6}. We trained all the models for 100 epochs. As pointed out by \cite{liu2021pscc}, it is inefficient to pre-train the models with the entire 1 Million images in TrainFors. We sampled 0.1 Million images randomly for training in each epoch.

% \vspace{-5px}
\subsection{Image Manipulation Localization}\label{sec.img_mani_loc}
\vspace{-5px}
We used the evaluation protocol defined in \cite{hu2020span} to evaluate the localization performance using two modules: Firstly, the pretrained model is trained on the TrainFors dataset and evaluated on the entire test set. Secondly, the pre-trained model is fine-tuned on the train-split of the evaluation datasets and evaluated on the test-splits of the datasets. We reported the upper and lower limits of the metrics on 6 runs to evaluate the performance variance. 

\noindent\textbf{Pre-trained model:} \cref{tab.pretrained_localization} reports the localization performance of the pre-trained models from \cref{baseline_models} on the five benchmark evaluation datasets from \cref{eval_datasets}, reporting the pixel-level AUC and F1 scores.  
The Cat-Net performance is best on the Columbia dataset and showed very poor performance on the Coverage dataset. This could be attributed to the fact that the Cat-Net model is designed for a Splicing dataset and Columbia and Coverage datasets have all-splicing and no-splicing images respectively (see \cref{tab.trainfor_desc}).
The pre-trained PSCCNet model achieves the best localization performance on Coverage, CASIAv1, and NIST16 datasets, when all the models were pre-trained with the same backbone network. The pre-trained MVSS-Net model achieves the best performance on the IMD20 dataset because it has real-world images and PSCCNet was pretrained on synthetic datasets. The PSCCNet has the best generalization ability compared to the other models as it is pre-trained on a large amount of synthetic training data.

\noindent\textbf{Fine-tuned model:} According to the second protocol, we further fine-tuned the pre-trained model on the train-split of the evaluation datasets. The training strategy of fine-tuned models is similar to that of the pre-trained models. We compare the pixel-level AUC and F1 scores of fine-tuned models on \cref{tab.finetuned_localization}. All the fine-tuned models behaved similarly to the results portrayed by the pre-trained models. The fine-tuned PSCCNet model achieves the best localization performance on Coverage, CASIAv1, and NIST16 datasets. The fine-tuned Cat-Net and MVSS-Net models achieve the best localization performance on Columbia and IMD20 datasets respectively. It should be noted that the fine-tuned model performance improves when we train them with the EfficientNetV2 backbone network. 

% \begin{table*}%[!h]
\begin{table}
    \vspace{-5px}
    \centering
    \small
    \setlength{\tabcolsep}{0.1pt}
    
    \scalebox{0.55}{
    % \begin{tabular}{l| c| c| c| c| c| c| c| c| c| c}
    \begin{tabular}{l c c c c c c c c c c}
    \toprule

    & \bf{No Dis-} & \bf{Resize} & \bf{Resize} & \bf{Gau-Blur} & \bf{Gau-Blur} & \bf{Gau-N} & \bf{Gau-N} & \bf{JPG-Comp} & \bf{JPG-Comp} &  \\

    & \bf{tortion} & \bf{(0.78X)} & \bf{(0.25X)} & \bf{(k=3)} & \bf{(k=15)} & \bf{($\sigma$=3)} & \bf{($\sigma$=15)} & \bf{(q=100)} & \bf{(q=50)} & \bf{Mixed} \\
    
     \cmidrule(lr){2-2} \cmidrule(lr){3-3} \cmidrule(lr){4-4} \cmidrule(lr){5-5} \cmidrule(lr){6-6} \cmidrule(lr){7-7}  \cmidrule(lr){8-8} \cmidrule(lr){9-9} \cmidrule(lr){10-10} \cmidrule(lr){11-11}
    
    & & & & & \bf{Columbia} \\
    \midrule
    \textbf{Author-Specified Backbone} \\
    \cmidrule(lr){1-1}
    MVSS-Net\cite{chen2021image} & 61.1 & 60.9 & 60.9 & 59.8 & 58.7 & 57.8 & 56.4 & 59.3 & 59.5 & 56.3 \\
    Cat-Net\cite{kwon2021cat} & 56.3 & 56.2 & 56.2 & 55.4 & 54.8 & 53.2 & 52.7 & 55.6 & 55.7 & 52.6 \\
    PSCCNet\cite{liu2021pscc} & 58.7 & 58.6 & 58.6 & 57.9 & 57.2 & 56.5 & 55.6 & 57.8 & 57.9 & 56.4 \\
    ObjectFormer\cite{wang2022objectformer} & 55.8 & 55.8 & 55.7 & 53.2 & 52.8 & 48.4 & 47.8 & 54.6 & 54.7 & 47.5 \\   

    \midrule
    \textbf{EfficientNetV2 \cite{tan2021efficientnetv2} Backbone} \\
    \cmidrule(lr){1-1}
    MVSS-Net\cite{chen2021image} & 68.3 & 67.9 & 67.8 & 66.7 & 65.6 & 64.7 & 63.5 & 66.8 & 66.7 & 63.2 \\
    Cat-Net\cite{kwon2021cat} & \bf{70.4} & \bf{70.3} & \bf{70.2} & \bf{68.5} & \bf{67.7} & \bf{67.2} & \bf{66.6} & \bf{69.2} & \bf{69.4} & \bf{66.2} \\
    PSCCNet\cite{liu2021pscc} & 62.7 & 62.5 & 62.4 & 60.7 & 59.8 & 57.7 & 57.1 & 60.3 & 60.4 & 56.6 \\
    ObjectFormer\cite{wang2022objectformer} &  55.8 & 55.8 & 55.7 & 53.2 & 52.8 & 48.4 & 47.8 & 54.6 & 54.7 & 47.5 \\ 

    \midrule
    & & & & & \bf{NIST16} \\
    \midrule
    \textbf{Author-Specified Backbone} \\
    \cmidrule(lr){1-1}
    MVSS-Net\cite{chen2021image} & 38.7 & 38.7 & 38.7 & 38.6 & 36.4	& 35.6 & 34.7 & 37.6 & 37.7 & 34.2 \\
    Cat-Net\cite{kwon2021cat} & 29.6 & 29.6 & 29.6 & 29.5 & 28.6 & 27.9 & 27.2 & 29.1 & 29.2 & 26.8 \\
    PSCCNet\cite{liu2021pscc} & 48.6 & 48.6 & 48.6 & 48.5 & 47.1 & 46.2 & 45.3 & 47.6 & 47.5 & 45.1 \\
    ObjectFormer\cite{wang2022objectformer} & 48.7 & 48.7 & 48.6 & 48.5 & 46.4 & 45.7 & 44.9 & 48.6 & 48.6 & 44.7 \\   

    \midrule
    \textbf{EfficientNetV2 \cite{tan2021efficientnetv2} Backbone} \\
    \cmidrule(lr){1-1}
    MVSS-Net\cite{chen2021image} & 42.1 & 42.1 & 42.0 & 41.9 & 40.2 & 39.8 & 39.3 & 41.9 & 41.9 & 39.1 \\
    Cat-Net\cite{kwon2021cat} & 37.2 & 37.2 & 37.1 & 36.9 & 35.1 & 34.7 & 33.8 & 37.1 & 37.2 & 33.6 \\
    PSCCNet\cite{liu2021pscc} & \bf{52.6} & \bf{52.5} & \bf{52.4} & \bf{51.7} & \bf{50.9} & \bf{50.1} & \bf{49.6} & \bf{51.8} & \bf{51.8} & \bf{49.4} \\
    ObjectFormer\cite{wang2022objectformer} & 48.7 & 48.7 & 48.6 & 48.5 & 46.4 & 45.7 & 44.9 & 48.6 & 48.6 & 44.7 \\ 
    
    \bottomrule
    \end{tabular}
    }
    \vspace{-5px}
    \caption{Robustness Comparison of Pixel-level Manipulation Localization AUC(\%) with various distortions evaluated on Columbia\cite{ng2009columbia} and NIST16\cite{NimbleCh33:online}  datasets, when pretrained with author-specified backbone networks and EfficientNetV2 \cite{tan2021efficientnetv2} backbone network respectively}
     \label{tab.robustness_localization}
% \end{table*}
\end{table}

It is clearly evident from \cref{tab.pretrained_localization} and \cref{tab.finetuned_localization} that the performance of pre-trained and fine-tuned models vary if the backbone network is altered. The Objectformer performance was on par with the other baseline models for using a superior backbone pre-training model, but it was outperformed, when the backbone network was fixed. The evaluated results reported in the baseline papers deteriorate when pre-trained with a more generalized dataset, showcasing the importance of a generalized TrainFors dataset. 

\begin{figure*}%[!h]
    \centering
    \settoheight{\tempdima}{\includegraphics[width=.15\linewidth]    {Images/FigVis/casia_orig1.jpg}}
    
    \centering\begin{tabular}{@{ }c@{ }c@{ }c@{ }c@{ }c@{ }c@{ }c@{ }c@{ }}
    
    \textbf{Pristine} & \textbf{Manipulated} & \textbf{Groundtruth} & \textbf{MVSS-Net\cite{chen2021image}} & \textbf{Cat-Net\cite{kwon2021cat}} & \textbf{PSCCNet\cite{liu2021pscc}} & \textbf{ObjectFormer\cite{wang2022objectformer}} \\
    
    % \rowname{NIST16}&
    \includegraphics[width=.13\linewidth]{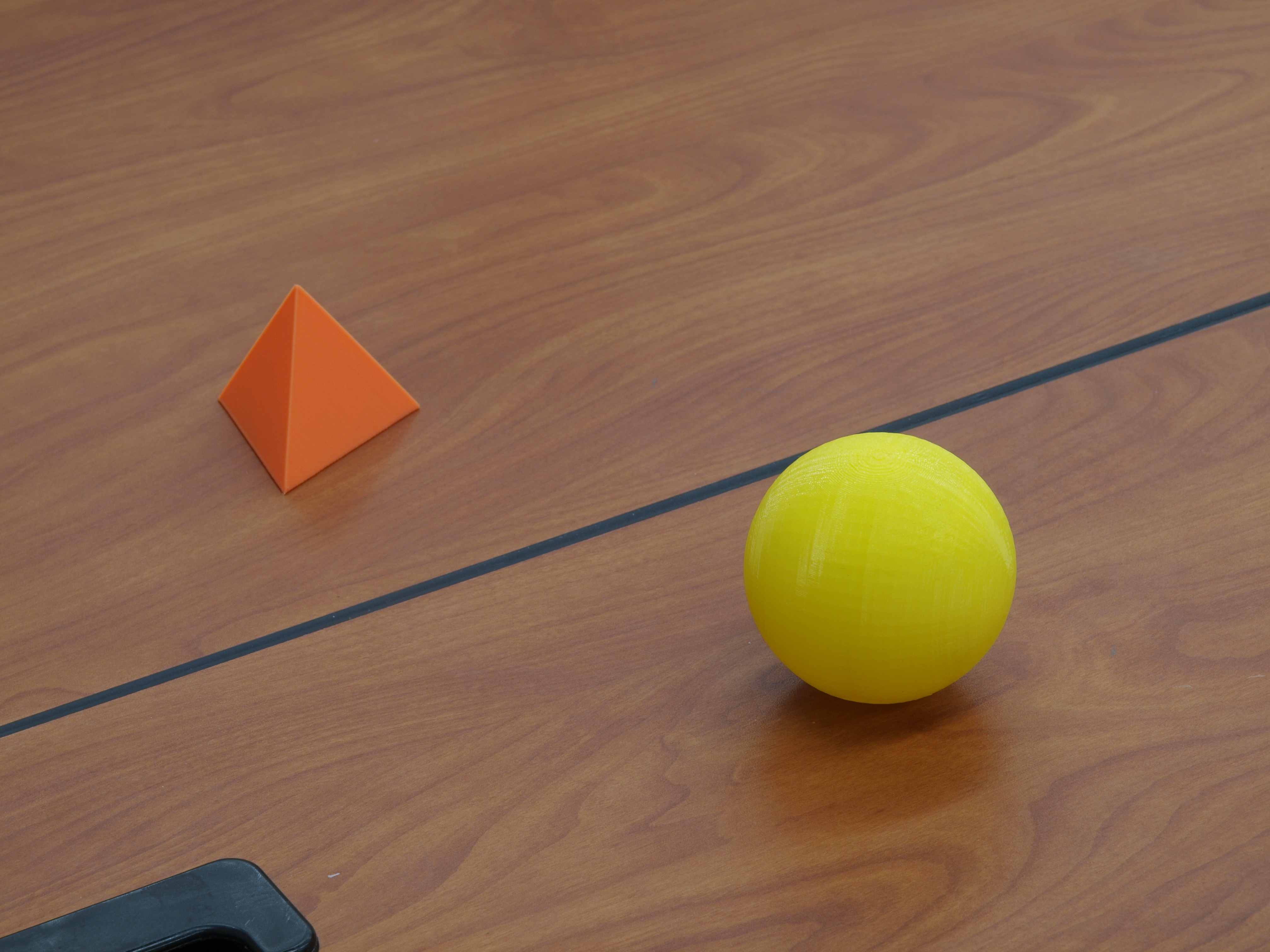}& 
    \includegraphics[width=.13\linewidth]{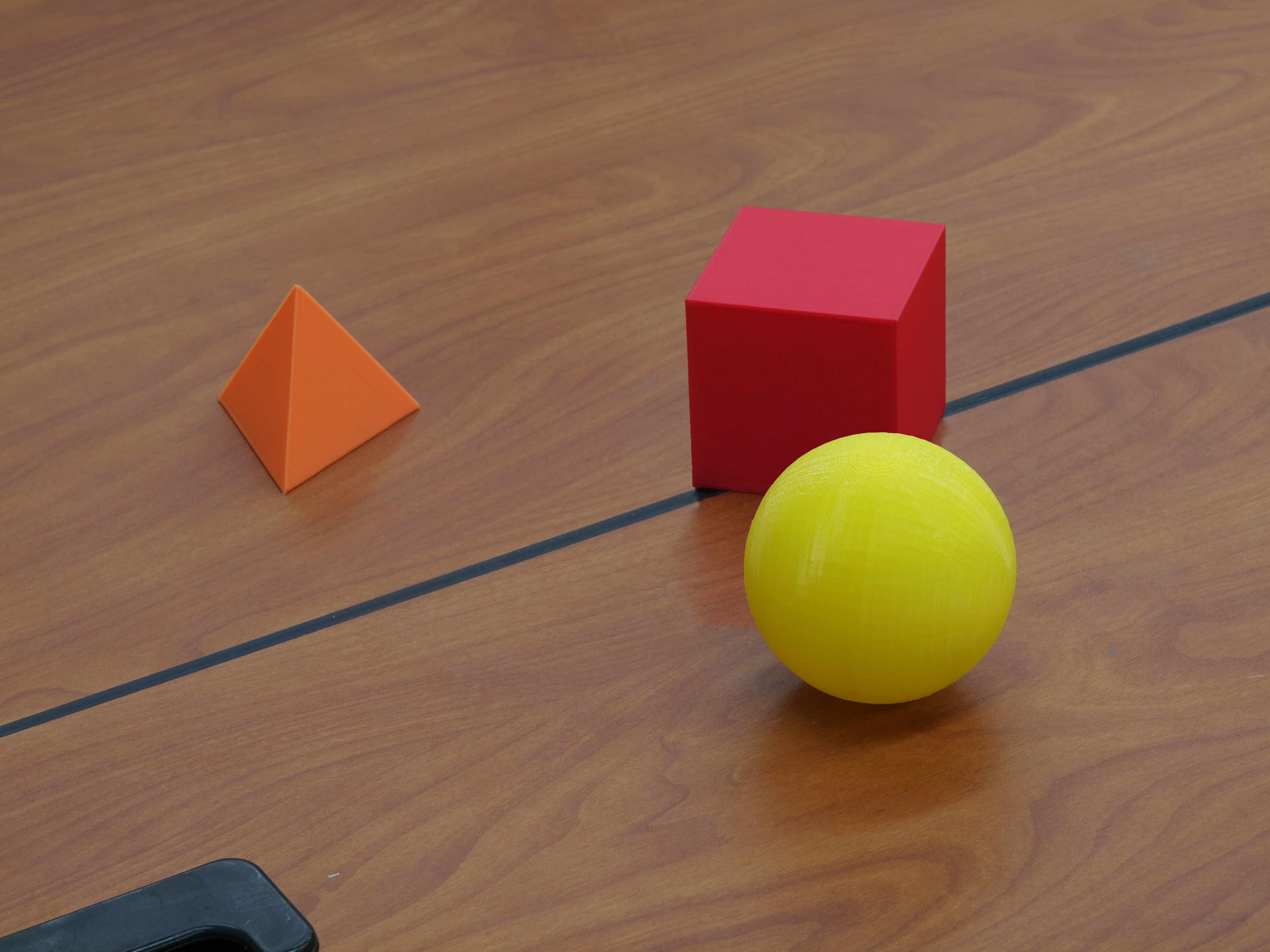}& 
    \includegraphics[width=.13\linewidth]{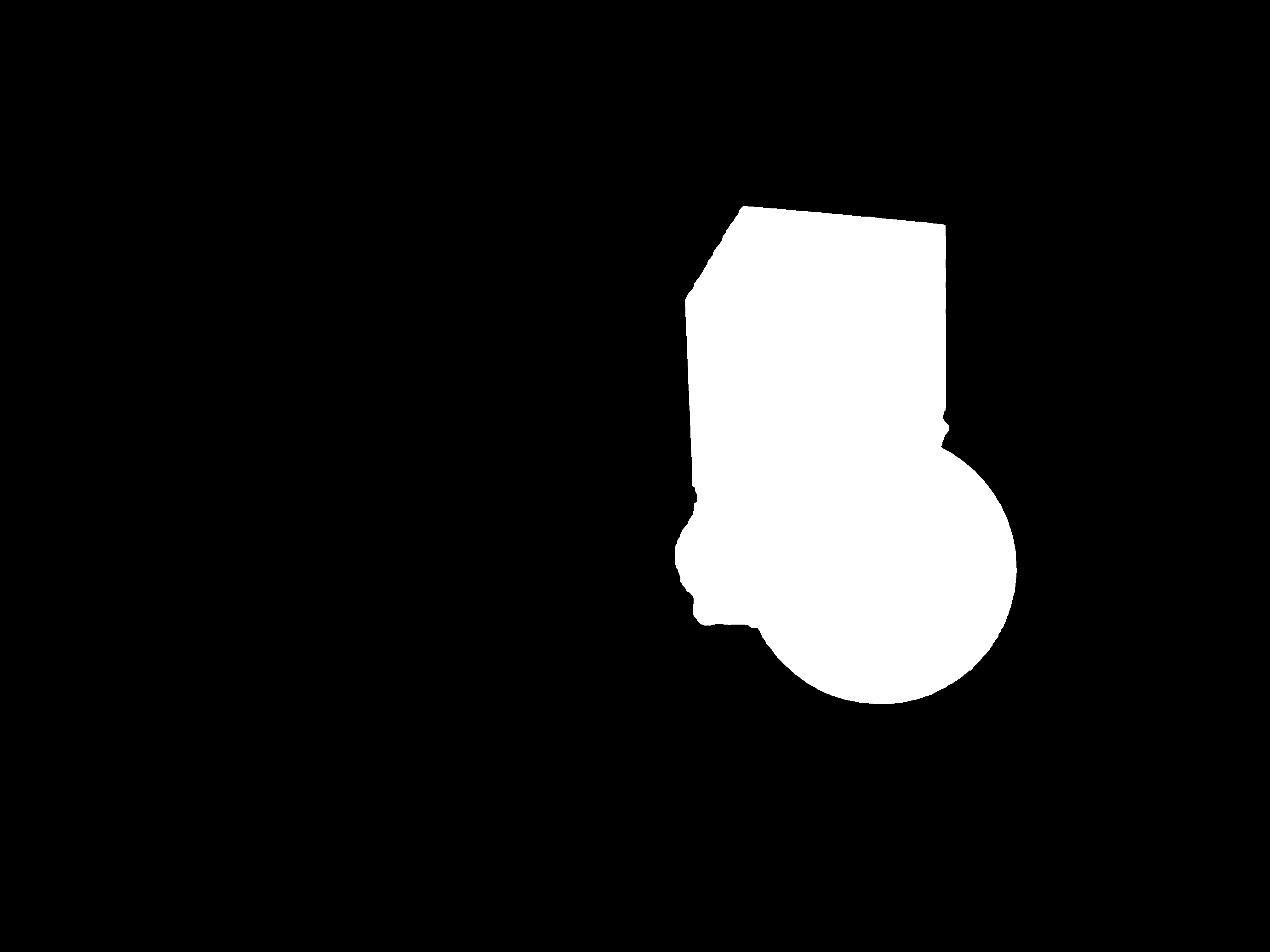}&
    \includegraphics[width=.13\linewidth]{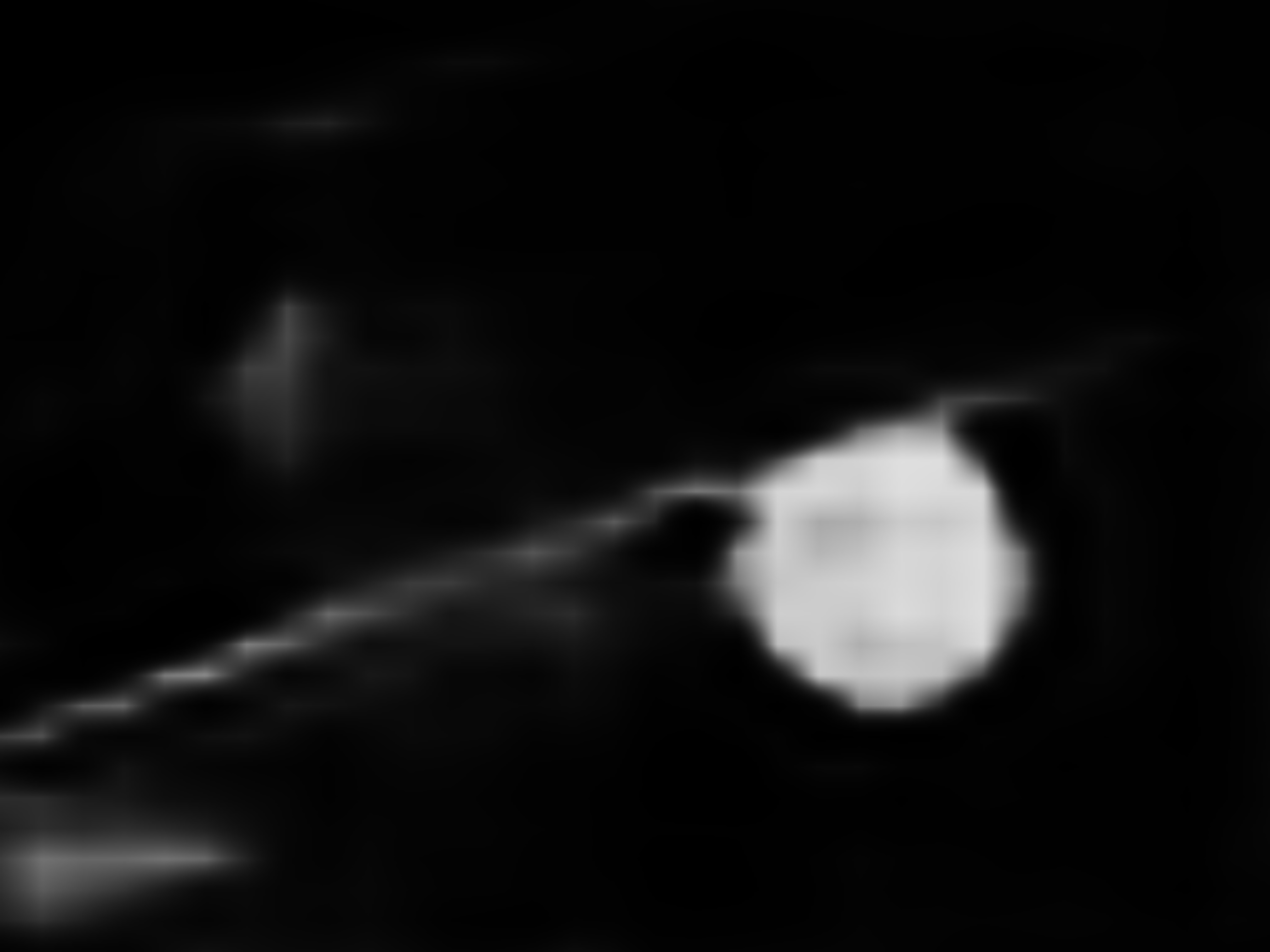}&
    \includegraphics[width=.13\linewidth]{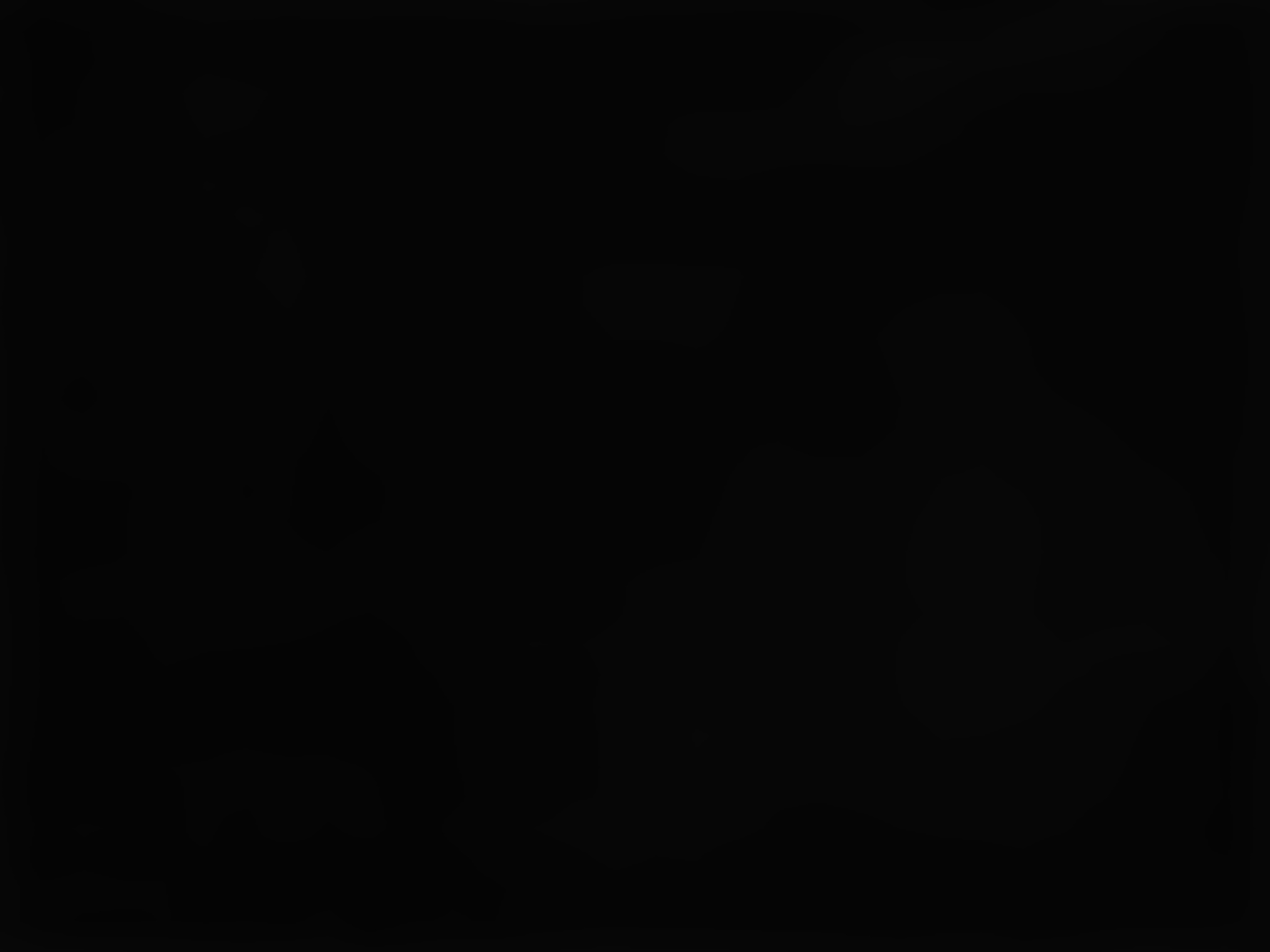}&
    \includegraphics[width=.13\linewidth]{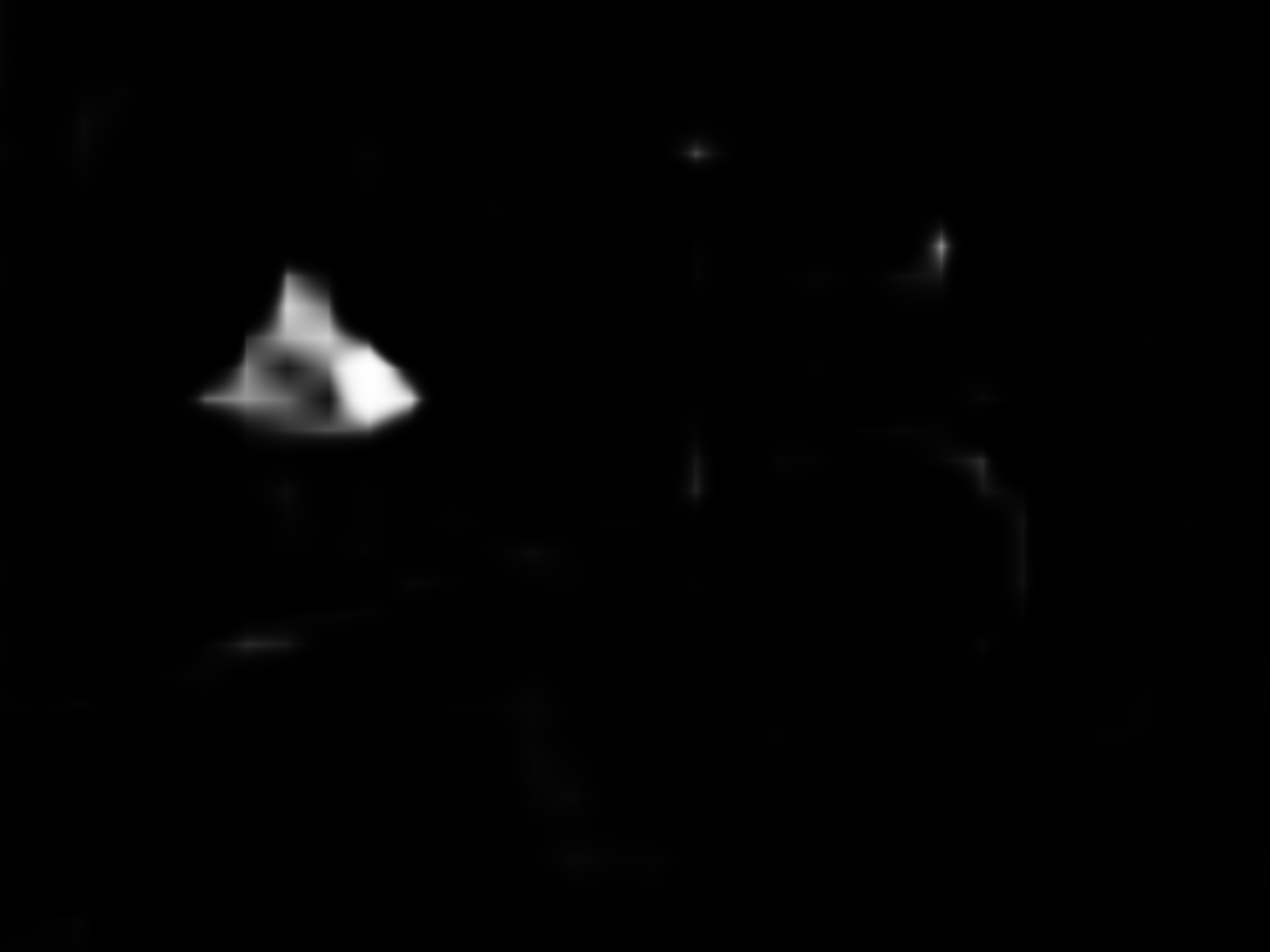}&
    \includegraphics[width=.13\linewidth]{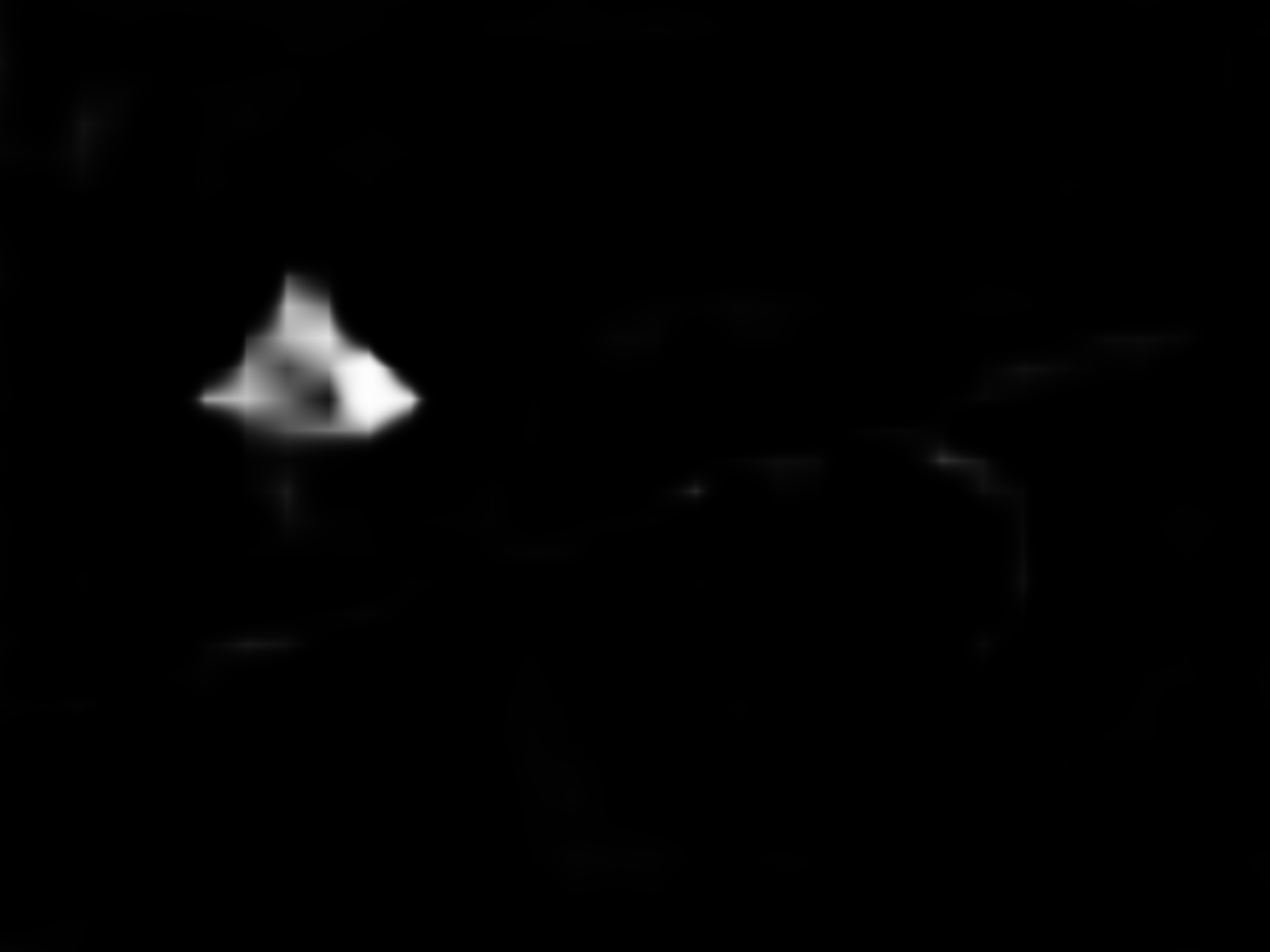}

    \end{tabular}
    \caption{Image Manipulation False Alarm Predictions by 4 baseline models: MVSS-Net, Cat-Net, PSCC-Net, and ObjectFormer, trained on our proposed TrainFors dataset.}
    \label{fig:IMDL_False_Alarm}
\end{figure*}

\subsection{Image Manipulation Detection}\label{sec.img_mani_det}
\vspace{-5px}
Similar to \cite{liu2021pscc} and \cite{wang2022objectformer}, we used the pre-trained models for the image manipulation detection task and reported the image-level AUC and F1 scores in \cref{tab.manipulation_detection} on four benchmark evaluation datasets. We cannot evaluate the manipulation detection results on the NIST16 dataset as it does not have any pristine (negative) images (see \cref{tab.trainfor_desc}). All the previous research work reported the detection results (AUC score only) on the Casia dataset only, but for a fair comparison, we reported the results (both AUC and F1 scores) on all four datasets. With the author-specified backbone network pre-training, Objectformer achieved the best performance, but with the EfficientNetV2 backbone network, all the other baseline models outperformed Objectformer. PSCCNet gave the best manipulation detection performance on the Coverage and Casia datasets and MVSS-Net on Columbia and IMD20 datasets. For a fair evaluation of the baseline models, we reported the image manipulation localization and detection tasks separately. But for an efficient IMDL task, manipulation detection should be performed before manipulation localization and only the detected images should be checked for the manipulated pixels.

% \vspace{-5px}
\subsection{Qualitative Analysis}\label{sec.qualitative_analysis}
\vspace{-5px}

We reported the manipulation predictions of the baseline models, after training them with TrainFors dataset in \cref{fig:IMDL_Predictions}. The models can easily predict manipulated pixels if large objects are tampered with in an image (eg, Casia and IMD20). For copymove examples, if multiple similar objects are present in an image (eg, more than one panda in Coverage), model prediction is not very accurate. MVSS-Net showed promising predictions in this case. However, if a single object is copymoved, almost all the baseline models can predict the tampered pixels, except Cat-Net (eg, sea-lion copymoved in Nist16). The reason for Cat-Net's poor performance may be attributed to the fact that it was designed for image splicing and may fail for other types of manipulations. In removal images, a good blending method can result in model failure. It can be concluded that the baseline IMDL models' performance changes when pretrained with the same training dataset and the same backbone pre-training network. Further discussion in the supplementary material.

% \vspace{-5px}
\subsection{Robustness Evaluation}\label{sec.img_mani_robust}
\vspace{-5px}
All the previous methods \cite{wu2019mantra}, \cite{hu2020span}, \cite{liu2021pscc}, \cite{wang2022objectformer} used different image distortion methods on raw images from the NIST16\cite{NimbleCh33:online} and/or Columbia\cite{ng2009columbia} datasets, and
evaluated the robustness of the models. Similarly, we added the following distortions to the manipulated images: \noindent\textbf{Image scaling} with scales=0.78X, 0.25X, \noindent\textbf{Gaussian Blurring} with a kernel size k=3, 15, added \noindent\textbf{Gaussian Noise} with a standard deviation $\sigma$= 3, 15, and \noindent\textbf{JPEG Compression} with a quality factor q=50, 100. We also added a mix of these distortions in the mixed column and the No Distortion column. We compared the manipulation localization performance (AUC scores) of the pre-trained models with all the baseline methods on these distorted images and report the results on Columbia and NIST16 datasets in \cref{tab.robustness_localization}. Cat-Net and PSCCNet demonstrated the best robustness against various image distortions, on the Columbia and NIST16 datasets respectively, when the EfficientNetV2 backbone network was used for pre-training. Image manipulation detection robustness analysis is discussed in the supplementary material.

% \vspace{-5px}
\subsection{Runtime Analysis}\label{sec.runtime}
\vspace{-5px}
We measured the runtime in terms of frames per second (FPS) on the inference models and evaluated them on NVIDIA GeForce RTX 2080 Ti GPU. Cat-Net is quite inefficient with 4.2 FPS, when compared to 18.6 FPS in MVSS-Net and 51.3 FPS in PSCCNet. Objectformer is much slower than the other counterparts at 1.9 FPS. 

% \vspace{-5px}
\subsection{Limitations}
\vspace{-5px}

The biggest challenge for the IMDL researchers is failed predictions and false alarms, see \cref{fig:IMDL_False_Alarm}. The presence of multiple objects can sometimes misguide the IMDL models to easily predict a pristine object as manipulated. Other cases where an image is enhanced, without content modification, may also lead to manipulation detection failures.

\vspace{-5px}
\section{Conclusion}
\vspace{-5px}

We introduced TrainFors, a large-scale training dataset for the image manipulation detection and localization task. We hope that TrainFors will be used as a benchmark training dataset for the IMDL task, similar to the benchmark evaluation datasets. We performed extensive experiments with the state-of-the-art baseline IMDL models and showcased fair comparisons under a similar training setup that was not done by the IMDL community previously. The experimental results authenticate that the IMDL model performance is dependent on the training data and the backbone networks used for pre-training and fine-tuning. Along with standardized training and evaluation datasets, standardized evaluation metrics (both pixel-level and image-level AUC and F1 scores) were used to fairly compare the IMDL models. We will release the dataset, which will be available to the research community for future research on detecting and localizing manipulated images.

% \vspace{-5px}
\section{Acknowledgement}
\vspace{-5px}

This research was funded by the US Government Tailored INCAS EcoSystem (TIES) project HR0011-21-C-0164. The views, opinions, and/or findings expressed are those of the author(s) and should not be interpreted as representing the official views or policies of the Department of Defense or the U.S. Government.

%-------------------------------------------------------------------------
%%%%%%%%% REFERENCES

{\small
\bibliographystyle{ieee_fullname}
\bibliography{egbib}

\begin{thebibliography}{10}\itemsep=-1pt

\bibitem{kcmi:online}
Ieee’s signal processing society - camera model identification, 2018.
\newblock
  \url{https://www.kaggle.com/competitions/sp-society-camera-model-identification}.

\bibitem{NimbleCh33:online}
Nimble challenge 2017 evaluation | nist.
\newblock
  \url{https://www.nist.gov/itl/iad/mig/nimble-challenge-2017-evaluation}.
\newblock (Accessed on 11/14/2020).

\bibitem{bammey2022non}
Quentin Bammey, Tina Nikoukhah, Marina Gardella, Rafael~Grompone von Gioi,
  Miguel Colom, and Jean-Michel Morel.
\newblock Non-semantic evaluation of image forensics tools: Methodology and
  database.
\newblock In {\em Proceedings of the IEEE/CVF Winter Conference on Applications
  of Computer Vision}, pages 3751--3760, 2022.

\bibitem{bappy2017exploiting}
Jawadul~H Bappy, Amit~K Roy-Chowdhury, Jason Bunk, Lakshmanan Nataraj, and BS
  Manjunath.
\newblock Exploiting spatial structure for localizing manipulated image
  regions.
\newblock In {\em Proceedings of the IEEE international conference on computer
  vision}, pages 4970--4979, 2017.

\bibitem{bappy2019hybrid}
Jawadul~H Bappy, Cody Simons, Lakshmanan Nataraj, BS Manjunath, and Amit~K
  Roy-Chowdhury.
\newblock Hybrid lstm and encoder--decoder architecture for detection of image
  forgeries.
\newblock {\em IEEE Transactions on Image Processing}, 28(7):3286--3300, 2019.

\bibitem{bayar2016deep}
Belhassen Bayar and Matthew~C Stamm.
\newblock A deep learning approach to universal image manipulation detection
  using a new convolutional layer.
\newblock In {\em Proceedings of the 4th ACM workshop on information hiding and
  multimedia security}, pages 5--10, 2016.

\bibitem{bayar2018constrained}
Belhassen Bayar and Matthew~C Stamm.
\newblock Constrained convolutional neural networks: A new approach towards
  general purpose image manipulation detection.
\newblock {\em IEEE Transactions on Information Forensics and Security},
  13(11):2691--2706, 2018.

\bibitem{bi2021reality}
Xiuli Bi, Zhipeng Zhang, and Bin Xiao.
\newblock Reality transform adversarial generators for image splicing forgery
  detection and localization.
\newblock In {\em Proceedings of the IEEE/CVF International Conference on
  Computer Vision}, pages 14294--14303, 2021.

\bibitem{bondi2016first}
Luca Bondi, Luca Baroffio, David G{\"u}era, Paolo Bestagini, Edward~J Delp, and
  Stefano Tubaro.
\newblock First steps toward camera model identification with convolutional
  neural networks.
\newblock {\em IEEE Signal Processing Letters}, 24(3):259--263, 2016.

\bibitem{bondi2017tampering}
Luca Bondi, Silvia Lameri, David Guera, Paolo Bestagini, Edward~J Delp, Stefano
  Tubaro, et~al.
\newblock Tampering detection and localization through clustering of
  camera-based cnn features.
\newblock In {\em CVPR Workshops}, volume~2, 2017.

\bibitem{chen2021image}
Xinru Chen, Chengbo Dong, Jiaqi Ji, Juan Cao, and Xirong Li.
\newblock Image manipulation detection by multi-view multi-scale supervision.
\newblock In {\em Proceedings of the IEEE/CVF International Conference on
  Computer Vision}, pages 14185--14193, 2021.

\bibitem{choi2017detecting}
Hak-Yeol Choi, Han-Ul Jang, Dongkyu Kim, Jeongho Son, Seung-Min Mun, Sunghee
  Choi, and Heung-Kyu Lee.
\newblock Detecting composite image manipulation based on deep neural networks.
\newblock In {\em 2017 International Conference on Systems, Signals and Image
  Processing (IWSSIP)}, pages 1--5. IEEE, 2017.

\bibitem{cozzolino2015efficient}
Davide Cozzolino, Giovanni Poggi, and Luisa Verdoliva.
\newblock Efficient dense-field copy--move forgery detection.
\newblock {\em IEEE Transactions on Information Forensics and Security},
  10(11):2284--2297, 2015.

\bibitem{cozzolino2015splicebuster}
Davide Cozzolino, Giovanni Poggi, and Luisa Verdoliva.
\newblock Splicebuster: A new blind image splicing detector.
\newblock In {\em 2015 IEEE International Workshop on Information Forensics and
  Security (WIFS)}, pages 1--6. IEEE, 2015.

\bibitem{cozzolino2019noiseprint}
Davide Cozzolino and Luisa Verdoliva.
\newblock Noiseprint: a cnn-based camera model fingerprint.
\newblock {\em IEEE Transactions on Information Forensics and Security},
  15:144--159, 2019.

\bibitem{deng2009imagenet}
Jia Deng, Wei Dong, Richard Socher, Li-Jia Li, Kai Li, and Li Fei-Fei.
\newblock Imagenet: A large-scale hierarchical image database.
\newblock In {\em 2009 IEEE conference on computer vision and pattern
  recognition}, pages 248--255. Ieee, 2009.

\bibitem{dhamo2020semantic}
Helisa Dhamo, Azade Farshad, Iro Laina, Nassir Navab, Gregory~D Hager, Federico
  Tombari, and Christian Rupprecht.
\newblock Semantic image manipulation using scene graphs.
\newblock In {\em Proceedings of the IEEE/CVF conference on computer vision and
  pattern recognition}, pages 5213--5222, 2020.

\bibitem{dong2013casia}
Jing Dong, Wei Wang, and Tieniu Tan.
\newblock Casia image tampering detection evaluation database.
\newblock In {\em 2013 IEEE China Summit and International Conference on Signal
  and Information Processing}, pages 422--426. IEEE, 2013.

\bibitem{dosovitskiy2020image}
Alexey Dosovitskiy, Lucas Beyer, Alexander Kolesnikov, Dirk Weissenborn,
  Xiaohua Zhai, Thomas Unterthiner, Mostafa Dehghani, Matthias Minderer, Georg
  Heigold, Sylvain Gelly, et~al.
\newblock An image is worth 16x16 words: Transformers for image recognition at
  scale.
\newblock {\em arXiv preprint arXiv:2010.11929}, 2020.

\bibitem{fan2015image}
Yu Fan, Philippe Carr{\'e}, and Christine Fernandez-Maloigne.
\newblock Image splicing detection with local illumination estimation.
\newblock In {\em 2015 IEEE international conference on Image processing
  (ICIP)}, pages 2940--2944. IEEE, 2015.

\bibitem{ferrara2012image}
Pasquale Ferrara, Tiziano Bianchi, Alessia De~Rosa, and Alessandro Piva.
\newblock Image forgery localization via fine-grained analysis of cfa
  artifacts.
\newblock {\em IEEE Transactions on Information Forensics and Security},
  7(5):1566--1577, 2012.

\bibitem{fridrich2012rich}
Jessica Fridrich and Jan Kodovsky.
\newblock Rich models for steganalysis of digital images.
\newblock {\em IEEE Transactions on information Forensics and Security},
  7(3):868--882, 2012.

\bibitem{galdi2019socrates}
Chiara Galdi, Frank Hartung, and Jean-Luc Dugelay.
\newblock Socrates: A database of realistic data for source camera recognition
  on smartphones.
\newblock In {\em ICPRAM}, pages 648--655, 2019.

\bibitem{gloe2010dresden}
Thomas Gloe and Rainer B{\"o}hme.
\newblock The'dresden image database'for benchmarking digital image forensics.
\newblock In {\em Proceedings of the 2010 ACM symposium on applied computing},
  pages 1584--1590, 2010.

\bibitem{goodfellow2014generative}
Ian Goodfellow, Jean Pouget-Abadie, Mehdi Mirza, Bing Xu, David Warde-Farley,
  Sherjil Ozair, Aaron Courville, and Yoshua Bengio.
\newblock Generative adversarial nets.
\newblock {\em Advances in neural information processing systems}, 27, 2014.

\bibitem{hadwiger2021forchheim}
Benjamin Hadwiger and Christian Riess.
\newblock The forchheim image database for camera identification in the wild.
\newblock In {\em Pattern Recognition. ICPR International Workshops and
  Challenges: Virtual Event, January 10--15, 2021, Proceedings, Part VI}, pages
  500--515. Springer, 2021.

\bibitem{han2016efficient}
Jong~Goo Han, Tae~Hee Park, Yong~Ho Moon, and Il~Kyu Eom.
\newblock Efficient markov feature extraction method for image splicing
  detection using maximization and threshold expansion.
\newblock {\em Journal of Electronic Imaging}, 25(2):023031, 2016.

\bibitem{hao2021transforensics}
Jing Hao, Zhixin Zhang, Shicai Yang, Di Xie, and Shiliang Pu.
\newblock Transforensics: image forgery localization with dense self-attention.
\newblock In {\em Proceedings of the IEEE/CVF International Conference on
  Computer Vision}, pages 15055--15064, 2021.

\bibitem{he2016deep}
Kaiming He, Xiangyu Zhang, Shaoqing Ren, and Jian Sun.
\newblock Deep residual learning for image recognition.
\newblock In {\em Proceedings of the IEEE conference on computer vision and
  pattern recognition}, pages 770--778, 2016.

\bibitem{horvath2021manipulation}
J{\'a}nos Horv{\'a}th, Sriram Baireddy, Hanxiang Hao, Daniel~Mas Montserrat,
  and Edward~J Delp.
\newblock Manipulation detection in satellite images using vision transformer.
\newblock In {\em Proceedings of the IEEE/CVF Conference on Computer Vision and
  Pattern Recognition}, pages 1032--1041, 2021.

\bibitem{hu2020span}
Xuefeng Hu, Zhihan Zhang, Zhenye Jiang, Syomantak Chaudhuri, Zhenheng Yang, and
  Ram Nevatia.
\newblock Span: spatial pyramid attention network for image manipulation
  localization.
\newblock In {\em European Conference on Computer Vision}, pages 312--328.
  Springer, 2020.

\bibitem{huang2017densely}
Gao Huang, Zhuang Liu, Laurens Van Der~Maaten, and Kilian~Q Weinberger.
\newblock Densely connected convolutional networks.
\newblock In {\em Proceedings of the IEEE conference on computer vision and
  pattern recognition}, pages 4700--4708, 2017.

\bibitem{huh2018fighting}
Minyoung Huh, Andrew Liu, Andrew Owens, and Alexei~A Efros.
\newblock Fighting fake news: Image splice detection via learned
  self-consistency.
\newblock In {\em Proceedings of the European conference on computer vision
  (ECCV)}, pages 101--117, 2018.

\bibitem{islam2020doa}
Ashraful Islam, Chengjiang Long, Arslan Basharat, and Anthony Hoogs.
\newblock Doa-gan: Dual-order attentive generative adversarial network for
  image copy-move forgery detection and localization.
\newblock In {\em Proceedings of the IEEE/CVF Conference on Computer Vision and
  Pattern Recognition}, pages 4676--4685, 2020.

\bibitem{kingma2014adam}
Diederik~P Kingma and Jimmy Ba.
\newblock Adam: A method for stochastic optimization.
\newblock {\em arXiv preprint arXiv:1412.6980}, 2014.

\bibitem{kingma2013auto}
Diederik~P Kingma and Max Welling.
\newblock Auto-encoding variational bayes.
\newblock {\em arXiv preprint arXiv:1312.6114}, 2013.

\bibitem{kniaz2019point}
Vladimir~V Kniaz, Vladimir Knyaz, and Fabio Remondino.
\newblock The point where reality meets fantasy: Mixed adversarial generators
  for image splice detection.
\newblock {\em Advances in Neural Information Processing Systems}, 32, 2019.

\bibitem{krawetz2007picture}
Neal Krawetz and Hacker~Factor Solutions.
\newblock A picture’s worth.
\newblock {\em Hacker Factor Solutions}, 6(2):2, 2007.

\bibitem{kwon2021cat}
Myung-Joon Kwon, In-Jae Yu, Seung-Hun Nam, and Heung-Kyu Lee.
\newblock Cat-net: Compression artifact tracing network for detection and
  localization of image splicing.
\newblock In {\em Proceedings of the IEEE/CVF Winter Conference on Applications
  of Computer Vision}, pages 375--384, 2021.

\bibitem{li2020manigan}
Bowen Li, Xiaojuan Qi, Thomas Lukasiewicz, and Philip~HS Torr.
\newblock Manigan: Text-guided image manipulation.
\newblock In {\em Proceedings of the IEEE/CVF Conference on Computer Vision and
  Pattern Recognition}, pages 7880--7889, 2020.

\bibitem{li2017image}
Ce Li, Qiang Ma, Limei Xiao, Ming Li, and Aihua Zhang.
\newblock Image splicing detection based on markov features in qdct domain.
\newblock {\em Neurocomputing}, 228:29--36, 2017.

\bibitem{lin2014microsoft}
Tsung-Yi Lin, Michael Maire, Serge Belongie, James Hays, Pietro Perona, Deva
  Ramanan, Piotr Doll{\'a}r, and C~Lawrence Zitnick.
\newblock Microsoft coco: Common objects in context.
\newblock In {\em Computer Vision--ECCV 2014: 13th European Conference, Zurich,
  Switzerland, September 6-12, 2014, Proceedings, Part V 13}, pages 740--755.
  Springer, 2014.

\bibitem{liu2021pscc}
Xiaohong Liu, Yaojie Liu, Jun Chen, and Xiaoming Liu.
\newblock Pscc-net: Progressive spatio-channel correlation network for image
  manipulation detection and localization.
\newblock {\em arXiv preprint arXiv:2103.10596}, 2021.

\bibitem{lyu2014exposing}
Siwei Lyu, Xunyu Pan, and Xing Zhang.
\newblock Exposing region splicing forgeries with blind local noise estimation.
\newblock {\em International journal of computer vision}, 110(2):202--221,
  2014.

\bibitem{mahdian2009using}
Babak Mahdian and Stanislav Saic.
\newblock Using noise inconsistencies for blind image forensics.
\newblock {\em Image and Vision Computing}, 27(10):1497--1503, 2009.

\bibitem{mahfoudi2019defacto}
Ga{\"e}l Mahfoudi, Badr Tajini, Florent Retraint, Frederic Morain-Nicolier,
  Jean~Luc Dugelay, and PIC Marc.
\newblock Defacto: image and face manipulation dataset.
\newblock In {\em 2019 27Th european signal processing conference (EUSIPCO)},
  pages 1--5. IEEE, 2019.

\bibitem{mirza2014conditional}
Mehdi Mirza and Simon Osindero.
\newblock Conditional generative adversarial nets.
\newblock {\em arXiv preprint arXiv:1411.1784}, 2014.

\bibitem{ng2009columbia}
Tian-Tsong Ng, Jessie Hsu, and Shih-Fu Chang.
\newblock Columbia image splicing detection evaluation dataset.
\newblock {\em DVMM lab. Columbia Univ CalPhotos Digit Libr}, 2009.

\bibitem{novozamsky2020imd2020}
Adam Novozamsky, Babak Mahdian, and Stanislav Saic.
\newblock Imd2020: A large-scale annotated dataset tailored for detecting
  manipulated images.
\newblock In {\em Proceedings of the IEEE/CVF Winter Conference on Applications
  of Computer Vision Workshops}, pages 71--80, 2020.

\bibitem{park2020swapping}
Taesung Park, Jun-Yan Zhu, Oliver Wang, Jingwan Lu, Eli Shechtman, Alexei
  Efros, and Richard Zhang.
\newblock Swapping autoencoder for deep image manipulation.
\newblock {\em Advances in Neural Information Processing Systems},
  33:7198--7211, 2020.

\bibitem{pathak2016context}
Deepak Pathak, Philipp Krahenbuhl, Jeff Donahue, Trevor Darrell, and Alexei~A
  Efros.
\newblock Context encoders: Feature learning by inpainting.
\newblock In {\em Proceedings of the IEEE conference on computer vision and
  pattern recognition}, pages 2536--2544, 2016.

\bibitem{pu2016variational}
Yunchen Pu, Zhe Gan, Ricardo Henao, Xin Yuan, Chunyuan Li, Andrew Stevens, and
  Lawrence Carin.
\newblock Variational autoencoder for deep learning of images, labels and
  captions.
\newblock {\em Advances in neural information processing systems}, 29, 2016.

\bibitem{rao2016deep}
Yuan Rao and Jiangqun Ni.
\newblock A deep learning approach to detection of splicing and copy-move
  forgeries in images.
\newblock In {\em 2016 IEEE international workshop on information forensics and
  security (WIFS)}, pages 1--6. IEEE, 2016.

\bibitem{ren2015faster}
Shaoqing Ren, Kaiming He, Ross Girshick, and Jian Sun.
\newblock Faster r-cnn: Towards real-time object detection with region proposal
  networks.
\newblock {\em Advances in neural information processing systems}, 28, 2015.

\bibitem{reshniak2020nonlocal}
Viktor Reshniak, Jeremy Trageser, and Clayton~G Webster.
\newblock A nonlocal feature-driven exemplar-based approach for image
  inpainting.
\newblock {\em SIAM Journal on Imaging Sciences}, 13(4):2140--2168, 2020.

\bibitem{sabir2019recurrent}
Ekraam Sabir, Jiaxin Cheng, Ayush Jaiswal, Wael AbdAlmageed, Iacopo Masi, and
  Prem Natarajan.
\newblock Recurrent convolutional strategies for face manipulation detection in
  videos.
\newblock {\em Interfaces (GUI)}, 3(1):80--87, 2019.

\bibitem{sabir2021biofors}
Ekraam Sabir, Soumyaroop Nandi, Wael Abd-Almageed, and Prem Natarajan.
\newblock Biofors: A large biomedical image forensics dataset.
\newblock In {\em Proceedings of the IEEE/CVF International Conference on
  Computer Vision}, pages 10963--10973, 2021.

\bibitem{salloum2018image}
Ronald Salloum, Yuzhuo Ren, and C-C~Jay Kuo.
\newblock Image splicing localization using a multi-task fully convolutional
  network (mfcn).
\newblock {\em Journal of Visual Communication and Image Representation},
  51:201--209, 2018.

\bibitem{shullani2017vision}
Dasara Shullani, Marco Fontani, Massimo Iuliani, Omar~Al Shaya, and Alessandro
  Piva.
\newblock Vision: a video and image dataset for source identification.
\newblock {\em EURASIP Journal on Information Security}, 2017(1):1--16, 2017.

\bibitem{tan2019efficientnet}
Mingxing Tan and Quoc Le.
\newblock Efficientnet: Rethinking model scaling for convolutional neural
  networks.
\newblock In {\em International conference on machine learning}, pages
  6105--6114. PMLR, 2019.

\bibitem{tan2021efficientnetv2}
Mingxing Tan and Quoc Le.
\newblock Efficientnetv2: Smaller models and faster training.
\newblock In {\em International conference on machine learning}, pages
  10096--10106. PMLR, 2021.

\bibitem{uricchio2017localization}
Tiberio Uricchio, Lamberto Ballan, Irene Roberto~Caldelli, et~al.
\newblock Localization of jpeg double compression through multi-domain
  convolutional neural networks.
\newblock In {\em Proceedings of the IEEE Conference on Computer Vision and
  Pattern Recognition Workshops}, pages 53--59, 2017.

\bibitem{vinker2020deep}
Yael Vinker, Eliahu Horwitz, Nir Zabari, and Yedid Hoshen.
\newblock Deep single image manipulation.
\newblock {\em arXiv preprint arXiv:2007.01289}, 2020.

\bibitem{bbcfakenews2022}
Jane Wakefield.
\newblock Deepfake presidents used in russia-ukraine war.
\newblock {\em BBC News}.

\bibitem{wang2020deep}
Jingdong Wang, Ke Sun, Tianheng Cheng, Borui Jiang, Chaorui Deng, Yang Zhao,
  Dong Liu, Yadong Mu, Mingkui Tan, Xinggang Wang, et~al.
\newblock Deep high-resolution representation learning for visual recognition.
\newblock {\em IEEE transactions on pattern analysis and machine intelligence},
  43(10):3349--3364, 2020.

\bibitem{wang2022objectformer}
Junke Wang, Zuxuan Wu, Jingjing Chen, Xintong Han, Abhinav Shrivastava, Ser-Nam
  Lim, and Yu-Gang Jiang.
\newblock Objectformer for image manipulation detection and localization.
\newblock {\em arXiv preprint arXiv:2203.14681}, 2022.

\bibitem{wen2016coverage}
Bihan Wen, Ye Zhu, Ramanathan Subramanian, Tian-Tsong Ng, Xuanjing Shen, and
  Stefan Winkler.
\newblock Coverage—a novel database for copy-move forgery detection.
\newblock In {\em 2016 IEEE International Conference on Image Processing
  (ICIP)}, pages 161--165. IEEE, 2016.

\bibitem{wu2022robust}
Haiwei Wu, Jiantao Zhou, Jinyu Tian, and Jun Liu.
\newblock Robust image forgery detection over online social network shared
  images.
\newblock In {\em Proceedings of the IEEE/CVF Conference on Computer Vision and
  Pattern Recognition}, pages 13440--13449, 2022.

\bibitem{wu2017deep}
Yue Wu, Wael Abd-Almageed, and Prem Natarajan.
\newblock Deep matching and validation network: An end-to-end solution to
  constrained image splicing localization and detection.
\newblock In {\em Proceedings of the 25th ACM international conference on
  Multimedia}, pages 1480--1502, 2017.

\bibitem{wu2018busternet}
Yue Wu, Wael Abd-Almageed, and Prem Natarajan.
\newblock Busternet: Detecting copy-move image forgery with source/target
  localization.
\newblock In {\em Proceedings of the European conference on computer vision
  (ECCV)}, pages 168--184, 2018.

\bibitem{wu2018image}
Yue Wu, Wael Abd-Almageed, and Prem Natarajan.
\newblock Image copy-move forgery detection via an end-to-end deep neural
  network.
\newblock In {\em 2018 IEEE Winter Conference on Applications of Computer
  Vision (WACV)}, pages 1907--1915. IEEE, 2018.

\bibitem{wu2019mantra}
Yue Wu, Wael AbdAlmageed, and Premkumar Natarajan.
\newblock Mantra-net: Manipulation tracing network for detection and
  localization of image forgeries with anomalous features.
\newblock In {\em Proceedings of the IEEE/CVF Conference on Computer Vision and
  Pattern Recognition}, pages 9543--9552, 2019.

\bibitem{yang2016image}
Qiuwei Yang, Fei Peng, Jiao-Ting Li, and Min Long.
\newblock Image tamper detection based on noise estimation and lacunarity
  texture.
\newblock {\em Multimedia Tools and Applications}, 75(17):10201--10211, 2016.

\bibitem{yu2021mask}
Qihang Yu, Jianming Zhang, He Zhang, Yilin Wang, Zhe Lin, Ning Xu, Yutong Bai,
  and Alan Yuille.
\newblock Mask guided matting via progressive refinement network.
\newblock In {\em Proceedings of the IEEE/CVF Conference on Computer Vision and
  Pattern Recognition}, pages 1154--1163, 2021.

\bibitem{zhang2018boundary}
Zhongping Zhang, Yixuan Zhang, Zheng Zhou, and Jiebo Luo.
\newblock Boundary-based image forgery detection by fast shallow cnn.
\newblock In {\em 2018 24th International Conference on Pattern Recognition
  (ICPR)}, pages 2658--2663. IEEE, 2018.

\bibitem{zhou2020generate}
Peng Zhou, Bor-Chun Chen, Xintong Han, Mahyar Najibi, Abhinav Shrivastava,
  Ser-Nam Lim, and Larry Davis.
\newblock Generate, segment, and refine: Towards generic manipulation
  segmentation.
\newblock In {\em Proceedings of the AAAI conference on artificial
  intelligence}, volume~34, pages 13058--13065, 2020.

\bibitem{zhou2018learning}
Peng Zhou, Xintong Han, Vlad~I Morariu, and Larry~S Davis.
\newblock Learning rich features for image manipulation detection.
\newblock In {\em Proceedings of the IEEE conference on computer vision and
  pattern recognition}, pages 1053--1061, 2018.

\bibitem{zhu2017unpaired}
Jun-Yan Zhu, Taesung Park, Phillip Isola, and Alexei~A Efros.
\newblock Unpaired image-to-image translation using cycle-consistent
  adversarial networks.
\newblock In {\em Proceedings of the IEEE international conference on computer
  vision}, pages 2223--2232, 2017.

\bibitem{zhu2018deep}
Xinshan Zhu, Yongjun Qian, Xianfeng Zhao, Biao Sun, and Ya Sun.
\newblock A deep learning approach to patch-based image inpainting forensics.
\newblock {\em Signal Processing: Image Communication}, 67:90--99, 2018.

\bibitem{zhu2020deformable}
Xizhou Zhu, Weijie Su, Lewei Lu, Bin Li, Xiaogang Wang, and Jifeng Dai.
\newblock Deformable detr: Deformable transformers for end-to-end object
  detection.
\newblock {\em arXiv preprint arXiv:2010.04159}, 2020.

\end{thebibliography}
}

\end{document}